\documentclass[sigconf, nonacm]{acmart}

\usepackage{algorithm}
\usepackage{algpseudocode}
\usepackage{newtxmath}
\usepackage{boldline}
\usepackage{subcaption}
\usepackage{textcomp}
\usepackage{makecell}
\usepackage{multirow}
\usepackage{color}
\usepackage{soul}
\usepackage{pifont}
\usepackage{caption}
\usepackage{siunitx}
\usepackage{setspace} 
\usepackage{placeins}
\usepackage{stfloats}

\AtBeginDocument{%
  }
    
\setcopyright{acmlicensed}
\copyrightyear{2026}
\acmYear{2026}
\acmDOI{XXXXXXX.XXXXXXX}

\acmConference[KDD '26]{the 32st ACM SIGKDD Conference on Knowledge Discovery and Data Mining}{August 09--13,
  2026}{Jeju, Korea}
  
\acmISBN{978-1-4503-XXXX-X/2018/06}

\begin{document}

\newcommand{\jz}[1]{{\color{blue}{\bf{[JZ:]}} #1}}
\newcommand{\zq}[1]{{\color{green}{\bf{[ZQ:]}} #1}}
\newcommand*{\red}{\textcolor{red}}
\newcommand*{\blue}{\textcolor{blue}}
\newcommand*{\orange}{\textcolor{orange}}
\newcommand*{\green}{\textcolor{green}}
\newcommand{\pmstd}{$\,\pm\,$}

\newcommand{\model}{UrbanVerse}
\newcommand{\RWM}{CELearning}
\newcommand{\diffM}{HCondDiffCT}
\newcommand{\priorGen}{RegCondP}
\newcommand{\denoisNet}{TaskCondD}

\title{\model: Learning Urban Region Representation Across Cities and Tasks}

\author{Fengze Sun}
\affiliation{%
  \institution{School of Computing and Information System, University of Melbourne}
  \city{Melbourne}
  \country{Australia}}
\email{fengzes@student.unimelb.edu.au}

\author{Egemen Tanin}
\affiliation{%
  \institution{School of Computing and Information System, University of Melbourne}
  \city{Melbourne}
  \country{Australia}}
\email{etanin@.unimelb.edu.au}

\author{Shanika Karunasekera}
\affiliation{%
  \institution{School of Computing and Information System, University of Melbourne}
  \city{Melbourne}
  \country{Australia}}
\email{karus@.unimelb.edu.au}

\author{Zuqing Li}
\affiliation{%
  \institution{School of Computing and Information System, University of Melbourne}
  \city{Melbourne}
  \country{Australia}}
\email{zuqing.li@student.unimelb.edu.au}

\author{Flora D. Salim}
\affiliation{%
  \institution{School of Computer Science and Engineering, University of New South Wales }
  \city{Sydney}
  \country{Australia}}
\email{flora.salim@unsw.edu.au}

\author{Jianzhong Qi}
\authornote{Corresponding author.}
\affiliation{%
  \institution{School of Computing and Information System, University of Melbourne}
  \city{Melbourne}
  \country{Australia}}
\email{jianzhong.qi@unimelb.edu.au}

\begin{abstract}
Recent advances in urban region representation learning have enabled a wide range of applications in urban analytics, yet existing methods remain limited in their capabilities to generalize across cities and analytic tasks. We aim to generalize urban representation learning beyond city- and task-specific settings, towards a foundation-style model for urban analytics. To this end, we propose \model, a model for cross-city urban representation learning and cross-task urban analytics.
For cross-city generalization, \model\ focuses on  features local to the target regions and structural features of the nearby regions rather than the entire city. We model regions as nodes on a graph, which enables a random walk-based  procedure to form  ``sequences of regions'' that reflect both local and neighborhood structural features for urban region representation learning.   
For cross-task generalization, we propose a cross-task learning module named \diffM. This module integrates region-conditioned prior knowledge and task-conditioned semantics into the diffusion process to jointly model multiple downstream urban prediction tasks. \diffM\ is generic. It can also be integrated with existing urban representation learning  models to enhance their downstream task effectiveness. 
Experiments on real-world datasets show that \model\ consistently outperforms state-of-the-art methods across six tasks under cross-city settings, achieving up to 35.89\% improvements in prediction accuracy.

\end{abstract}

\ccsdesc[500]{Information systems~Location based services}
\ccsdesc[500]{Information systems~Data mining}

\keywords{Urban region representation, data mining}

\maketitle

\section{Introduction}
\label{sec:introduction}

Urban region representation learning has attracted increasing attention in the urban computing community~\cite{zhang2017urban, zheng2014urban, kdd1, kdd2, su2025generalising, ijcai25su, chang}, aiming to encode urban regions as low-dimensional vector representations, i.e., embeddings. The  embeddings support a wide range of predictive urban analytics tasks, such as crime prediction~\cite{MGFN, HAFusion, HREP, ReCP} and population estimation~\cite{urbanclip, FlexiReg, GeoHG}. 

Recently, the growing emphasis on generalization in foundation models, e.g., for natural language processing~\cite{nlp, nlp2} and computer vision~\cite{cv, cv2},  reflects a paradigm shift in machine learning. 
Motivated by this shift, we aim to achieve a foundation-style model for urban region representation learning. 
We propose to design a model that, once trained, can be used to generate embeddings for regions \emph{across cities}, while the resulting embeddings can be used \emph{across multiple downstream task} with a single model trained at once.



\begin{figure}[htbp]
  \centering
  \includegraphics[width=\columnwidth,
  trim=0cm 0cm 0cm 0cm,
  clip]{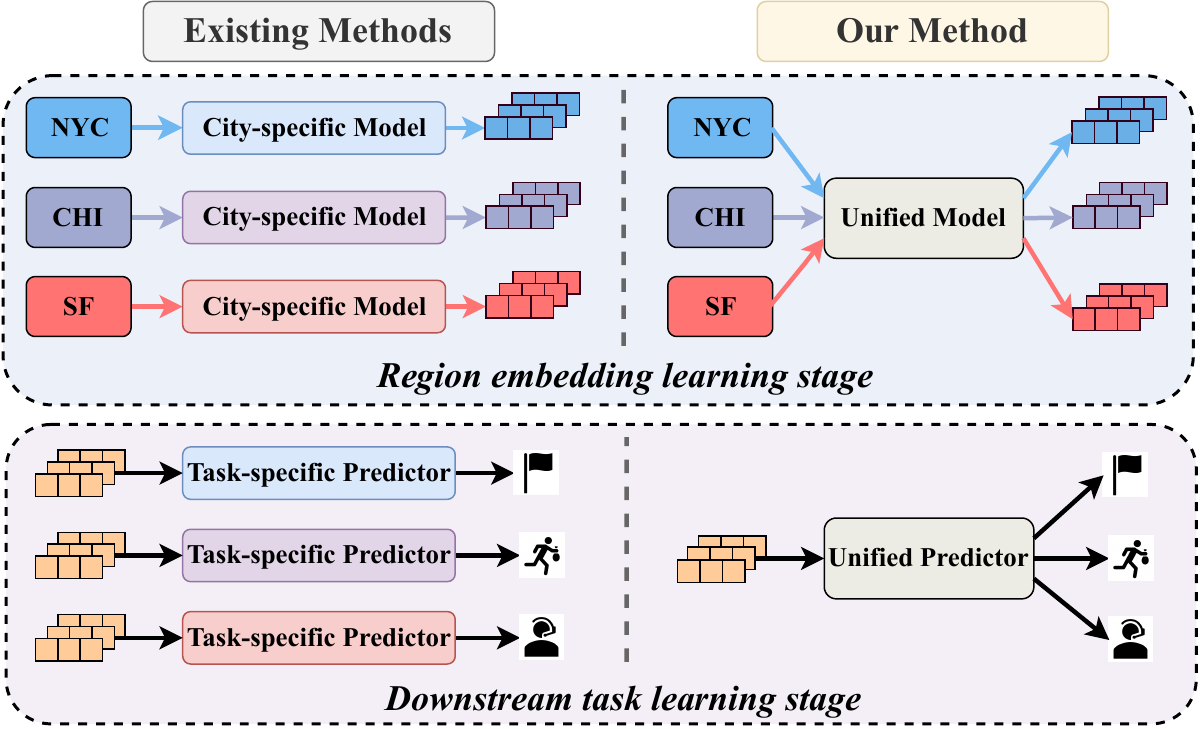}
  \vspace{-4mm}
  \caption{Region representation learning frameworks.}
  \label{fig:intro}
\end{figure}

\paragraph{Related Work}
Existing methods typically require retraining for regions in each different city and for each different downstream task. It is difficult to adopt their learning procedures for cross-city and cross-task settings. Achieving a foundation-style urban representation learning model calls for new model architectures. 

As Fig.~\ref{fig:intro} shows, 
urban representation learning studies often take two stages:  \emph{region embedding learning} and \emph{downstream task learning}. Existing methods fall short in generalizability for both stages. 

\textbf{Limitation 1. Existing methods have limited cross-city generalizability.}
For region embedding learning, existing methods are trained for embedding generation on data from individual cities (cf.~top left of Fig.~\ref{fig:intro}). These methods learn region embeddings in a self-supervised manner from a variety of region features, typically following one of three paradigms. The first paradigm reconstructs predefined region-level correlations, such as POI distribution similarity or taxi flow statistics between regions~\cite{DLCL, MP-VN, CDAE, CGAL, MVURE, CGAP, MGRL4RE, MGFN, RAW, Region2Vec}. 
The second paradigm adopts contrastive learning by constructing similar (i.e., positive) and dissimilar (i.e., negative) region pairs, e.g., based on region similarity derived from input features such as geographic proximity~\cite{urban2vec, ReMVC, MVGCL, cityFM, m3g, MMGR, MoRA, MuseCL, ReCP, RegionDCL, urbanclip, hyperregion, GeoHG}. 
The third paradigm combines reconstruction and contrastive objectives to further enhance representation learning~\cite{HAFusion,HREP, FlexiReg, GURPP, BPURF}. 
Most existing methods follow a city-centric design that optimizes city-specific objectives, leading to embeddings that overfit city-dependent patterns and generalize poorly across cities. 
Moreover, many existing solutions model an entire city as a  graph for model learning. Such city-centric graph modeling couples region representations with city-specific structures, limiting their generalization across cities. It also introduces scalability issues, as cities vary substantially in size and number of regions. 

\textbf{Limitation 2. Existing methods have limited cross-task generalizability.}
Existing methods apply learned region embeddings to downstream tasks using individual task-specific predictors, as illustrated in the bottom left of Fig.~\ref{fig:intro}. A few recent works~\cite{HREP, FlexiReg, GURPP} further inject task-relevant information to tailor embeddings for individual tasks. 
This paradigm fails to meet the cross-task generalization requirements and introduces a mismatch between representation learning and task modeling: while region embeddings are designed to be task-agnostic, downstream predictors are optimized individually for each task. Such task-specific training prevents the predictor models from exploiting shared patterns across urban tasks, such as regions with high population density often have high carbon emissions as well. 

More details about existing works can be found in Appendix~\ref{sec:appendic_of_related_work}.




\paragraph{Proposed Solution}

Towards our goal of a foundation-style urban region representation learning model, we propose \model\ for cross-city region representation learning and cross-task predictions.


Strong cross-city generalizability requires an embedding model that captures urban patterns that are transferable across cities. 
Following this intuition, we propose a \underline{C}ross-city \underline{E}mbedding \underline{Learning} module named \RWM\ that takes a region-centric approach.
Instead of modeling each city as a graph and regions as nodes, \RWM\ partitions a city into fine-grained grid cells, treats the cells as basic units, learns embeddings for them, and aggregates the embeddings of the cells overlapped by a region to form the embedding of the region. 

To achieve cross-city generalizability, we form a graph over the grid cells and perform random walks over the graph to form cell sequences. Such sequences focus on the local urban patterns of each target cell and its nearby neighbors, rather than global patterns at the city level, hence enabling cross-city generalization. Meanwhile, such sequences are not restricted to any city. Cell sequences from different cities can be used for cell embedding model training together. \RWM\ uses a transformer backbone~\cite{attention} to learn cell embeddings over the cell sequences. Once trained, this module can be used to generate embeddings for cells (and hence regions) from different cities, including those unseen at model training.

Given embeddings generated by \RWM, 
for cross-task generalization, 
we propose a \emph{\underline{H}eterogeneous \underline{C}onditional \underline{Diff}usion-based \underline{C}ross-\underline{T}ask Learning} (\diffM) module that jointly models multiple downstream tasks.
\diffM\ formulates urban prediction as a conditional diffusion-based regression problem, enabling a single model to capture complex, task-dependent value  distributions as well as the inherent uncertainty across different regions and tasks. 
To effectively distinguish among different regions and tasks, \diffM\ incorporates heterogeneous conditioning mechanisms that integrate both region-level priors and task semantics into the denoising process. A region-conditioned prior guidance (\priorGen) module injects region-specific prior knowledge to guide the diffusion process towards plausible outcomes, while a task-conditioned denoiser (\denoisNet) module conditions the denoising network on task semantics, enabling flexible adaptation across multiple tasks. Importantly, our \diffM\ module is generic -- it can be integrated with existing models to enhance their downstream task learning process, as shown in Section~\ref{subsec:app_of_diffusion}. 

To summarize, this paper makes the following contributions:

(1)~We propose \model, a model for cross-
city urban representation learning and cross-task urban analytics, addressing the generalization requirements of urban region embedding learning.

(2)~We introduce a cross-city embedding learning module to learn region embeddings across multiple cities, which captures transferable local urban patterns.

(3)~We develop a cross-task learning module that jointly models multiple downstream tasks within a unified diffusion process, capturing task-dependent value distributions and uncertainty.

(4)~We conduct extensive experiments on three real-world datasets. The results demonstrate that \model\ consistently outperforms all competing methods across six downstream tasks, achieving improvements of up to 35.89\% in accuracy. 



\begin{figure*}[htbp]
  \centering
  \includegraphics[width=\textwidth, trim=1cm 0cm 0cm 0cm, clip]{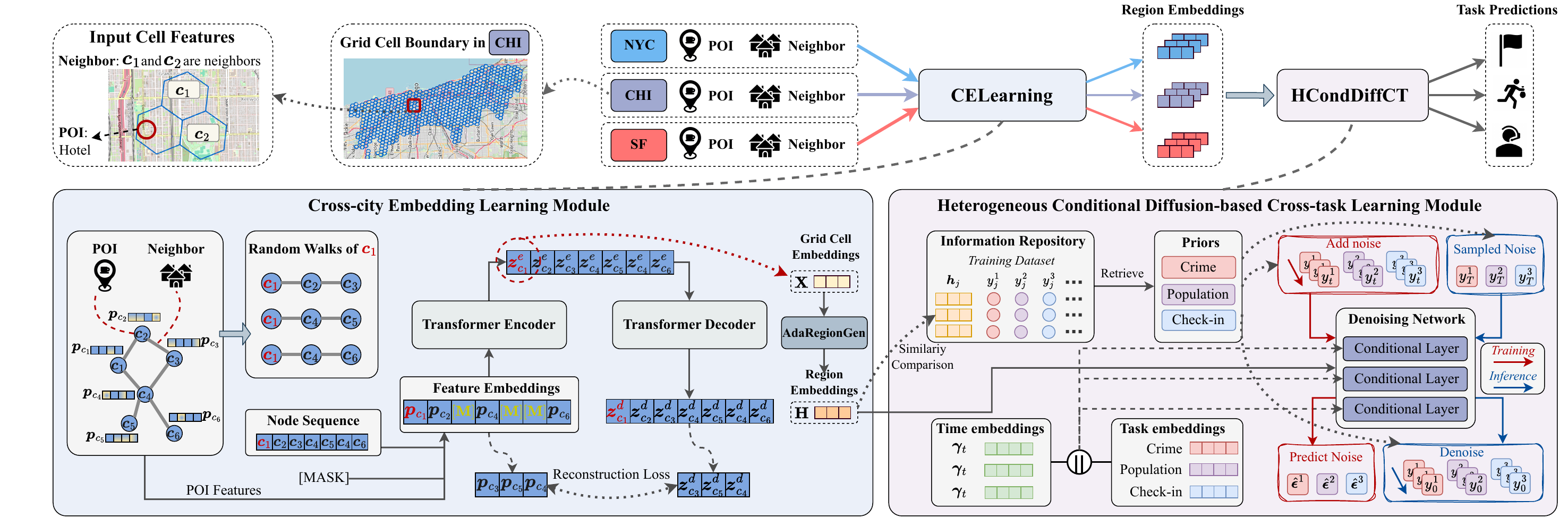}
  \vspace{-6mm}
  \caption{\model\ model overview. The model supports cross-city urban representation learning and cross-task urban analytics through two components: (1)~\RWM\ takes a set $C$ of grid cells and first learns cell embeddings $\mathbf{E}$ via cell sequences formed by random walks over a graph of cells. It then aggregates cell embeddings to generate region embeddings $\mathbf{H}$ across multiple cities, facilitating cross-city generalization. (2)~\diffM\ jointly models multiple tasks within a diffusion process to achieve cross-task generalization.}
  \label{fig:model_overview}
\end{figure*}

\section{Preliminaries}
\label{sec:preliminary}


We start with a few basic concepts and a problem statement. A notation table summarizing the symbols is included in Appendix~\ref{subsec:appendix_of_symbols}.

\textbf{Regions.} We aim to learn an embedding function $f$ to map an urban region to a low-dimensional vector. Each region, denoted as $r_i$, is a non-overlapping space partition from some area with given feature data (detailed below). We do not concern the exact partitioning method to obtain the regions (e.g., by census tracts). 

Instead of learning $f$ directly from regions, we learn a \emph{cell embedding function} $f_c$ from cells following FlexiReg~\cite{FlexiReg}, which are finer-grained space partitions with local feature patterns that can generalize to cells across cities. Once learned, we can use $f_c$ to generate cell embeddings, which are then aggregated to form region embeddings, via an aggregate function $f_a$ detailed in Section~\ref{subsubsec:region_emb_learning}. Function $f$ is formed by combining $f_c$ and $f_a$, i.e., $f = f_c \circ f_a$.


\textbf{Grid Cells.}  Grid cells $C$ refer to a set of fine-grained spatial partitions, where $c_i \in C$ denotes the $i$-th cell. We employ a hexagonal grid with an 
edge length of 150 meters to partition an area  with given feature data, as
illustrated by the top left of Fig.~\ref{fig:model_overview} and in  Appendix~\ref{subsec:appendix_of_cell_construction}. 
We use two general features for each cell as follows.




\textbf{POI features.} For each cell $c_i$, we count the number of POIs that belong to one of 15 POI categories (cf.~Appendix~\ref{subsec:appendix_of_dataset_statistics}) from OpenStreetMap~\cite{osm} as the POI feature, denoted as 
$\boldsymbol{p}_i (\boldsymbol{p}_i \in \mathbb{R}^{15})$.

\textbf{Geographic neighbor features.}
We also track the neighboring cells of each cell $c_i$, which reflect the adjacency relationships between cells. We use $\boldsymbol{gn}_{i} = \{ {gn}_{i,1}, {gn}_{i,2}, \cdots \}$ (${gn}_{i,j}$ is a cell ID) to denote the set of direct neighboring cells of $c_i$. Note that cells at the boundary of the area of interest may have fewer neighbors.

\textbf{Urban prediction task.} After the region embeddings are obtained, they are used make region-based predictions, such as crime counts, service calls, and population of a region.

\textbf{Problem statement.} 
Given a number of spatial areas (e.g., cities) with feature data, we aim to learn a region embedding function $f : r_i \rightarrow \bold h_i$ that maps a region $r_i$ from these areas to a $d$-dimensional vector $\boldsymbol{h}_{i}$. The function is expected to generalize to regions across different spatial areas, including those unseen during training. We achieve such a function via learning a generalizable cell embedding function $f_c$ as described above. 

Given $f$ and a set of regions of interest, we further aim to learn a prediction function $g: (\boldsymbol{h}_i, u) \rightarrow {y}_{i}^{u}$ that jointly models a set of urban prediction tasks based on the region embeddings $\boldsymbol{h}_i$ generated by $f$, and can flexibly predict for a task conditioned on the task indicator $u$. Here, ${y}_{i}^u$ denotes the target value of region $i$ for task $u$.

\section{Proposed Model}
\label{sec:model}




The overall structure of our model \model\ is illustrated with Fig.~\ref{fig:model_overview}.  \model\ takes grid cells from multiple cities as input and performs cross-city urban representation learning and cross-task urban analytics through two key modules. 

The \underline{C}ross-city \underline{E}mbedding \underline{Learning} (\RWM) module learns grid cell embeddings via random walks over graphs formed by the cells, which capture transferable local contexts.
The stochasticity of random walks serves as implicit data augmentation, enhancing robustness and generalization~\cite{perozzi2014deepwalk}.
The learned embedding function can compute cell embeddings across cities, which are aggregated to form region embeddings across cities (Section~\ref{subsec:RW_emb_learning}).

Given embeddings for regions of interest, the \underline{H}eterogeneous \underline{C}onditional \underline{Diff}usion-based \underline{C}ross-\underline{T}ask Learning (\diffM) module produces predictions for multiple downstream tasks together. It jointly models multiple  tasks within a diffusion process conditioned on region-specific prior knowledge and task semantics, thereby achieving cross-task generalization (Section~\ref{subsec:diff_multiTask_learning}).


\subsection{Cross-city Embedding Learning}
\label{subsec:RW_emb_learning}

The \RWM\ module consists of two components: \emph{cell embedding learning} (CellLearning, which is our target function $f_c$) and \emph{adaptive region embedding generation} (function $f_a$). CellLearning can handle cells from different cities and is less sensitive to city-level patterns, which is essential for generalization across multiple cities. AdapRegionGen further captures  variations across cells in the same  regions, enabling more expressive region representations.

\subsubsection{Cell Embedding Learning} 
CellLearning learns grid cell embeddings with a random-walk-based transformer architecture.

\textbf{Random walk construction for grid cells.}
We construct a grid graph based on the geographic neighbor features of grid cells, denoted  as $\mathcal{G} = (\mathcal{V}, \mathcal{E}, \mathbf{A})$. Here, $\mathcal{V} = \{c_1, c2, $ $ \cdots,$ $c_n\}$ represents the set of $n$ vertices, corresponding to the  $n$ grid cells in the given area with cell feature data, and  $\mathcal{E}$ denotes the set of edges between vertices.
The adjacency matrix $\mathbf{A}$ is defined by geospatial proximity, where an entry equals 1 if two cells are connected (i.e., sharing an edge) and 0 otherwise. Recall that we assume multiple input areas, which lead to multiple grid graphs. We consider one grid graph here to simplify the discussion. The steps below apply to all grid graphs constructed.

We generate a random-walk-based representation for each grid cell using $\mathcal{G}$. 
For each cell $c_i$, we treat it as a  root node and denote it as $c_{i,0}$. 
Starting from $c_{i,0}$, we perform $k$ (a system parameter) independent random walks, each with a length of $l$ ($l$ = 4 in our experiment). 
We concatenate the nodes from all random walks in order, walk by walk, to form a single sequence as follows, where $c_{i,j}^{x}$ represents the node reached at the $j$-th step of the $x$-th walk:
{\small
\begin{equation}
    seq(c_i) = [c_{i,0}, c_{i,1}^{1}, \cdots ,c_{i,l}^{1}, c_{i,1}^{2}, \cdots ,c_{i,l}^{2}, \cdots ,c_{i,1}^{k}, \cdots ,c_{i,l}^{k}], 
\end{equation}}
We follow the walk implementation of Node2vec~\cite{grover2016node2vec} to  balance depth-first and breadth-first exploration via parameters $p$ and $q$.



Given $seq(c_i)$, we construct  embedding $\mathbf{S}_i \in \mathbb{R}^{(k\times l + 1) \times 15}$  by extracting the corresponding POI feature vectors $\boldsymbol{p} \in \mathbb{R}^{15}$ associated with each node in the sequence. For conciseness, we use position indices to denote the elements in $\mathbf{S}_i$, where  $\boldsymbol{p}_{i,j}$ corresponds to the $j$-th node in $seq(c_i)$:
{\small
\begin{equation}
   \mathbf{S}_i = [\boldsymbol{p}_{i,0}, \boldsymbol{p}_{i,1}, \cdots, \boldsymbol{p}_{i, k  \times l}]. 
\end{equation}}

\textbf{Self-supervised masking strategy.}
Masking-based reconstruction is a widely used self-supervised objective in pretrained language models~\cite{devlin2019bert}.
Following this paradigm, we apply node masking within each sequence $seq(c_i)$ (i.e., the embedding $\mathbf{S}_i$) and train a model to reconstruct the features of the masked nodes. 
We adopt a random masking strategy with masking ratio $\rho$ (empirically set to 0.3). 
For an input sequence $\mathbf{S}_i$, we uniformly sample a fraction $\rho$ of positions, $\mathcal{M} \sim \mathrm{Uniform}({1,2,\ldots,k\times l})$, and replace them with special `\text{[MASK]}' tokens.
{\small
\begin{equation}
     \mathbf{S}_i = [\boldsymbol{p}_{i,0}, \boldsymbol{p}_{i,1}, \cdots ,\text{[MASK]}_{m \in \mathcal{M}}, \cdots ,\boldsymbol{p}_{i,k \times l}],
\end{equation}
}where $\rho = |\mathcal{M}|/(k \times l)$, the `\text{[MASK]}' tokens are initialized as learnable vectors representing masked positions. 
Random masking encourages the modeling of diverse relationships while serving as a regularizer that improves generalization.

\textbf{Transformer backbone.}
After constructing a sequential cell  representation, we employ an encoder–decoder transformer model to capture correlations among grid cells.
A masked input sequence $\mathbf{S}_i$ 
is first fed into the encoder, which consists of $L_e$ (a system parameter) stacked transformer blocks. 
The input $\mathbf{S}_i$ is first processed by a multi-head self-attention~\cite{attention} module to generate hidden representations $\mathbf{Z}_i$:
{\small
\begin{equation}
    \mathbf{Z}_i = \text{Attention} (\mathbf{Q}, \mathbf{K}, \mathbf{V}) = \mathrm{Softmax}\big(\frac{\mathbf{Q} \cdot {\mathbf{K}}^\intercal}{\sqrt{d}}\big) \cdot \mathbf{V},
\end{equation}
}where $\mathbf{Q}, \mathbf{K}$, and $\mathbf{V}$ denote query, key, and value matrices, respectively, obtained via linear transformations of $\mathbf{S}_i$ into latent spaces.

The output of self-attention, $\mathbf{Z}_i$, is combined with the masked input $\mathbf{S}_i$ through a residual connection, followed by layer normalization and dropout. A feed-forward neural network~(FFN) is then applied, after which another residual connection with layer normalization is used to further enhance the model’s learning capacity:
{\small
\begin{gather}
    \mathbf{Z}_i' = \mathrm{LayerNorm}(\mathbf{S}_i + \mathrm{Dropout}(\mathbf{Z}_i)), \label{eq:attn_postprocess1} \\
    {\mathbf{Z}_i^e} = \mathrm{LayerNorm}(\mathbf{Z}_i' + \mathrm{Dropout}(\mathrm{FFN}(\mathbf{Z}_i'))). \label{eq:attn_postprocess2}
\end{gather}
}Here,  $\mathbf{Z}_i'$ is the output of the first layer normalization, and $\mathbf{Z}_i^e = \{\boldsymbol{z}_{i,0}^e, \boldsymbol{z}_{i,1}^e, \cdots, \boldsymbol{z}_{i,(k \times l)}^e \} $ denotes the output of the encoder. 
We use $\boldsymbol{z}_{i,0}^e$ as the embedding of cell $c_i$, denotes as $\boldsymbol{x}_i$.

The decoder then reconstructs the masked node based on $\mathbf{Z}_i^e$. The decoder consists of $L^d$ transformer blocks, each sharing the same structure as the encoder blocks. Accordingly, the decoder produces $\mathbf{Z}_i^d$, i.e., $\mathbf{Z}_i^d$ = decoder$(\mathbf{Z}_i^e)$.

\textbf{Model training.}
For each input sequence, the model is trained to reconstruct the POI feature vectors of the masked nodes. We apply a  multi-layer perceptron (MLP) to project $\mathbf{Z}_i^d$ back to the POI feature space, i.e., $\mathbf{Z}_i^p = \text{MLP} (\mathbf{Z}_i^d)$. We then compute the mean squared error between the reconstructed POI features $\mathbf{Z}_i^p$ and the original POI features $\mathbf{S}_i$ for the masked positions:
{\small
\begin{equation}
    \mathcal{L} = \frac{1}{n}\frac{1}{|\mathcal{M}|}\sum_{i=1}^{n}\sum_{j=1}^{|\mathcal{M}|} \big( \mathbf{Z}_{i,j}^p - \mathbf{S}_{i,j} \big)^2
\end{equation}
}where $n$ is the number of cells and $\mathcal{M}$ is the set of masked positions.

\subsubsection{Adaptive Region Embedding learning.}
\label{subsubsec:region_emb_learning}

We adopt the adaptive region embedding generation module (AdaRegionGen) from FlexiReg~\cite{FlexiReg} to generate region embeddings $\mathbf{H} = \{ \boldsymbol{h}_j \}_{j=1}^{N}$ by aggregating the grid cell embeddings $\mathbf{X} = \{ \boldsymbol{x}_{i}\}_{i=1}^{n}$ corresponding to the cells $c_i \in \mathcal{C}_{r_j}$ that overlap with region $r_j$, weighted by their overlapping areas. We elaborate this module in Appendix~\ref{subse:appendix_of_adaRegionGen}.


\subsection{Heterogeneous Conditional  Diffusion-based Cross-task Learning}
\label{subsec:diff_multiTask_learning}




Given region embeddings $\mathbf{H}=\{{\boldsymbol{h}}\}_{j=1}^N$, we perform cross-task prediction within a unified framework by formulating the problem as a conditional diffusion-based regression task. This formulation enables the model to capture complex, task-dependent output distributions and the inherent uncertainty across regions and tasks.

We propose \diffM, a diffusion-based module with heterogeneous conditioning, built upon the Denoising Diffusion Probabilistic Model (DDPM, detailed in Appendix~\ref{subsec:appendix_of_ddpm})~\cite{ho2020denoising}. \diffM\ explicitly integrates region-specific prior knowledge and task semantics as conditions into the diffusion process to estimate the conditional distribution $p(\boldsymbol{y} \mid \boldsymbol{h}, u)$.
It contains two sub-modules: a \emph{Region-conditioned Prior Guidance}  (\priorGen) module, which injects region-specific prior knowledge into both the forward and reverse processes, and a \emph{Task-conditioned Denoiser}~(\denoisNet) module, which incorporates task semantics.

\subsubsection{Region-conditioned Prior Guidance}
\label{subsubsec:RLPrior}

\priorGen\ consists of two main steps: prior knowledge generation and injection.

\begin{figure}[htbp]
  \centering
  \includegraphics[width=\columnwidth,
  trim=1.3cm 0cm 1.2cm 0cm,
  clip]{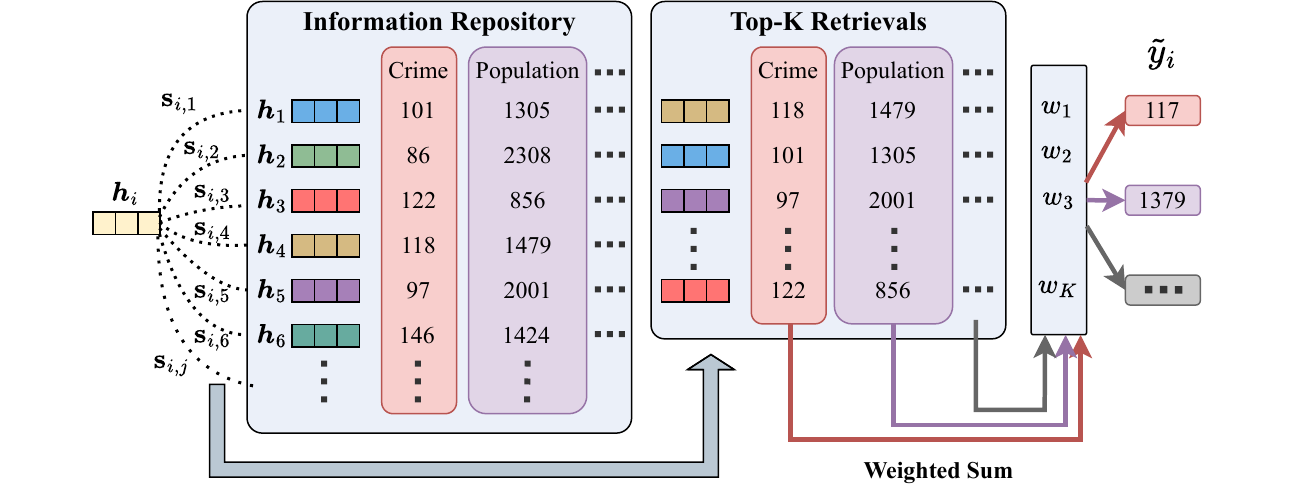}
  \caption{Prior knowledge generation.}
  \label{fig:prior_generation}
\end{figure}

\textbf{Prior knowledge generation.} 
To generate informative prior knowledge that captures the relationship between region embeddings and downstream task predictions without requiring additional training, we propose a retrieval-based prior knowledge generation method, as illustrated in Fig.~\ref{fig:prior_generation}.

Given a collection of region embeddings together with their corresponding ground-truth values from multiple downstream tasks, we aim to estimate task predictions for target regions in the test set by retrieving relevant information from the training set.

We construct an information repository $\mathbb{IR}$ from the training set as the retrieval pool, where each entry contains a region embedding and its corresponding ground-truth values across all tasks:
{\small
\begin{equation}
\mathbb{IR} = \left\{ (\boldsymbol{h}_i, y_i^1, \ldots, y_i^u) \mid i = 1, \ldots, |\mathbb{IR}| \right\}.
\end{equation}
}where $\boldsymbol{h}_i$ denotes the embedding of region $i$, ${y}_i^u$ represents the ground-truth value of region $i$ for the $u$-th downstream task, and $|\mathbb{IR}|$ denotes the size of the information repository.

We retrieve task-related information for target regions based on embedding similarity. Intuitively, regions with similar embeddings are expected to exhibit similar patterns in their downstream task values. 
For a target region $i$, we compute its similarity scores with all regions in the training set and select the top-$K$ most similar regions (with $K=5$). The corresponding similarity scores $\mathbf{s}_i$ are: 
{\small
\begin{equation}
\mathbf{s}_i = \{ \mathrm{sim}(\boldsymbol{h}_i, \boldsymbol{h}_j) \mid j \in \mathcal{N}_i^K \},
\end{equation}
}where $\mathrm{sim}(\cdot,\cdot)$ denotes the cosine similarity function, $\boldsymbol{h}_i$ and $\boldsymbol{h}_j$ are the embeddings of the target region and a training region, respectively, and $\mathcal{N}_i^K$ represents the index set of the top-$K$ most similar training regions to region $i$.
To obtain normalized retrieval weights $w_{i}$ for region $i$, we apply a softmax function over $\mathbf{s}_i$:
{\small
\begin{equation}
    w_{i,j} = \mathrm{softmax}(s_{i,j}), \quad j \in \mathcal{N}_i^K.
\end{equation}
}
For a downstream task $u$ over region $i$ with values $\{{y}_{i,j}^u\}_{j \in \mathcal{N}_i^K}$ retrieved from the top-K similar training regions, the estimated value $\widetilde{{y}}_i^u$, serving as prior knowledge, is computed as:
{\small
\begin{equation}
\widetilde{{y}}_i^u
=
\sum_{j \in \mathcal{N}_i^K}
w_{i,j} \, {y}_{i,j}^u .
\end{equation}}




\textbf{Prior knowledge injection.}
In our setting, a single region embedding is used to perform multiple downstream task predictions. As a result, each task is associated with its own task-specific prior knowledge $\widetilde{\boldsymbol{y}}^u$ (we drop the subscript $i$ from now on as the discussion does not concern a particular region $i$). 
To incorporate task-specific prior knowledge $\widetilde{\boldsymbol{y}}^u$ from multiple downstream tasks $u \in \{1,2,\cdots,$ $U\}$ as conditional information, we explicitly integrate it into both the forward and reverse processes of the diffusion model.
Inspired by Han et al.~\cite{han2022card}, and different from vanilla diffusion models that assume the endpoint of the diffusion process follows a standard normal distribution $\mathcal{N}(\mathbf{0}, \boldsymbol{I})$, we model the endpoint $\boldsymbol{y}_T^u$ of the diffusion process as:
{\small
\begin{equation}
p\!\left(\boldsymbol{y}_T^u \mid \boldsymbol{h}, u \right)
= \mathcal{N}\!\left(\widetilde{\boldsymbol{y}}^u, \boldsymbol{I}\right),
\end{equation}
}where $\widetilde{\boldsymbol{y}}^u$ represents the prior knowledge between region embedding $\boldsymbol{h}$ and downstream task ground-truth values  $\boldsymbol{y}_0^u$. This prior can be interpreted as an approximation to the conditional expectation
$\mathbb{E}\!\left[ \boldsymbol{y}_0^u \mid \boldsymbol{h}, u \right]$.
With a diffusion schedule $\{\beta_t\}_{t=1}^{T}$, the forward process at intermediate timesteps is defined as:
{\small
\begin{equation}\label{eq:forward_1}
q\!\left(
\boldsymbol{y}_t^{u} \mid \boldsymbol{y}_{t-1}^{u}, \widetilde{\boldsymbol{y}}^{u}
\right)
=
\mathcal{N}\!\left(
\boldsymbol{y}_t^{u};
\sqrt{1-\beta_t}\,\boldsymbol{y}_{t-1}^{u}
+
\left(1-\sqrt{1-\beta_t}\right)\widetilde{\boldsymbol{y}}^{u},
\beta_t \boldsymbol{I}
\right).
\end{equation}
}
In practice, we sample $\boldsymbol{y}_t^{u}$ from $\boldsymbol{y}_0^{u}$ at an arbitrary timestep $t$:
{\small
\begin{equation}\label{eq:forward_2}
q\!\left(
\boldsymbol{y}_t^{u} \mid \boldsymbol{y}_0^{u}, \widetilde{\boldsymbol{y}}^{u}
\right)
=
\mathcal{N}\!\left(
\boldsymbol{y}_t^{u};
\sqrt{\bar{\alpha}_t}\,\boldsymbol{y}_0^{u}
+
\left(1-\sqrt{\bar{\alpha}_t}\right)\widetilde{\boldsymbol{y}}^{u},
(1-\bar{\alpha}_t)\boldsymbol{I}
\right),
\end{equation}
}where ${\alpha}_t = 1-\beta_t$ and $\bar{\alpha}_t = \prod_{i=1}^{t}{\alpha}_i$.
From the mean term in Eq.~\eqref{eq:forward_1}, the forward process can be interpreted as a task-wise interpolation between the ground-truth values $\boldsymbol{y}_0^{u}$ and the prior knowledge $\widetilde{\boldsymbol{y}}_0^{u}$, progressively shifting from the former to the latter during diffusion.

It is also crucial to incorporate the prior knowledge $\widetilde{\boldsymbol{y}}_0^{u}$ into the reverse process. Given the forward process in Eq.~\eqref{eq:forward_1}, the corresponding tractable posterior of forward process is given by
\begin{equation}\label{eq:reverse}
q\!\left(
\boldsymbol{y}_{t-1}^{u}
\mid
\boldsymbol{y}_{t}^{u},
\boldsymbol{y}_{0}^{u},
\widetilde{\boldsymbol{y}}^{u}
\right)
=
\mathcal{N}\!\left(
\boldsymbol{y}_{t-1}^{u};
\gamma_0 \boldsymbol{y}_{0}^{u}
+
\gamma_1 \boldsymbol{y}_{t}^{u}
+
\gamma_2 \widetilde{\boldsymbol{y}}^{u},
\tilde{\beta}^{t}\boldsymbol{I}
\right),
\end{equation}
where
$\gamma_0
=
\frac{\beta^{t}\sqrt{\bar{\alpha}^{t-1}}}{1-\bar{\alpha}^{t}},
$ $
\gamma_1
=
\frac{\left(1-\bar{\alpha}^{t-1}\right)\sqrt{{\alpha}^{t}}}{1-\bar{\alpha}^{t}},
$ $\gamma_2
=
1
+
\frac{\left(\sqrt{\bar{\alpha}^{t}}-1\right)
\left(\sqrt{{\alpha}^{t}}+\sqrt{\bar{\alpha}^{t-1}}\right)}
{1-\bar{\alpha}^{t}},
$ and $
\tilde{\beta}^{t}
=
\frac{1-\bar{\alpha}^{t-1}}{1-\bar{\alpha}^{t}}\beta^{t}.$
Detailed derivation is provided in Appendix~\ref{subsec:appendix_of_derivation_for_forward}.

\subsubsection{Task-conditioned Denoiser}
\label{subsubsec:task_condition}
\begin{figure}[htbp]
  \centering
  \includegraphics[width=\columnwidth,
  trim=1.6cm 0cm 3cm 0cm,
  clip]{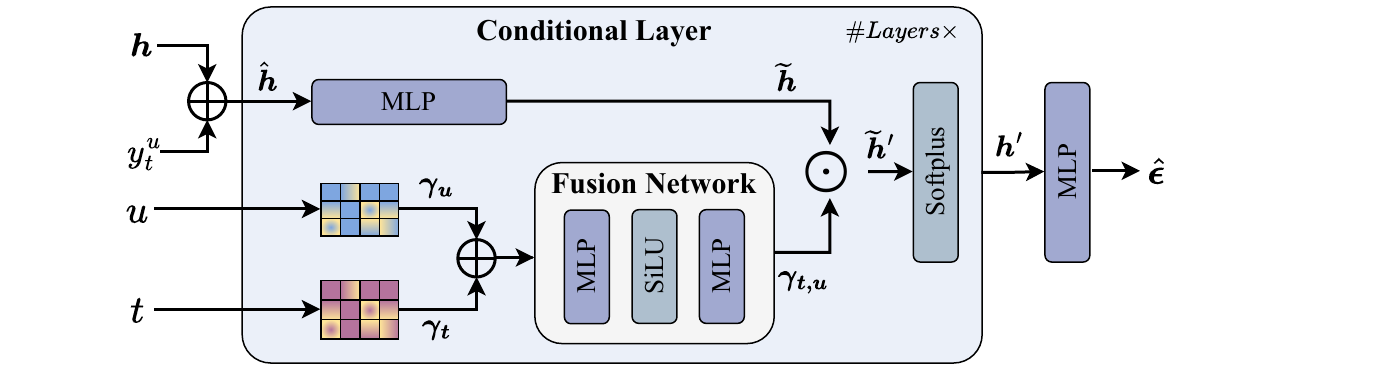}
  \caption{Task-conditioned denoiser.}
  \label{fig:task_condition}
\end{figure}

To support multi-task prediction, we incorporate task-aware signals as an additional condition to guide task-specific predictions. 
Rather than injecting task information into the forward and reverse diffusion processes or concatenating it with the input, we directly modulate the denoising network, as illustrated in Fig.~\ref{fig:task_condition}.
This design allows the diffusion model to adapt its denoising dynamics to different tasks, resulting in more stable optimization and improved task-specific modeling.


We propose a task-conditioned denoiser module (\denoisNet), a conditional denoising network that explicitly incorporates both the diffusion timestep and task-specific information via element-wise modulation.
Given a region embedding $\boldsymbol{h}$, a diffusion timestep $t \in \{0, \cdots, T \}$, a task index $u \in \{ 1, \cdots, U \}$, and the current noisy observation ${y}_t^u$ at timestep $t$ for the downstream task $u$, we concatenate $\boldsymbol{h}$ and ${y}_t^u$ along the feature dimension to form the initial input $\hat{\boldsymbol{h}} \in \mathbb{R}^{d+1}$, i.e.,  $\hat{\boldsymbol{h}} = [\boldsymbol{h} \: \Vert \: {y}_t^u]$.
The resulting input $\hat{\boldsymbol{h}}$ is then fed into a conditional layer, where an MLP is first applied to map it to a hidden representation $\widetilde{\boldsymbol{h}} \in \mathbb{R}^{d_{dn}}$, i.e., $\widetilde{\boldsymbol{h}} = \text{MLP}(\hat{\boldsymbol{h}})$.

To condition the denoising process on the diffusion timestep, we introduce a learnable timestep embedding  $\boldsymbol{\gamma_t} \in \mathbb{R}^{d_{dn}}$. In parallel, task-specific information is encoded using a task embedding $\boldsymbol{\gamma_u} \in \mathbb{R}^{d_{dn}}$. These two embeddings are concatenated and passed through a lightweight fusion network, denoted as $\text{Fusion}(\cdot)$.
This produces a unified modulation vector  $\boldsymbol{\gamma_{t, u}} \in \mathbb{R}^{d_{dn}}$, given by $\boldsymbol{\gamma_{t, u}} = \text{Fusion} ([\boldsymbol{\gamma_t} \: \Vert \:\boldsymbol{\gamma_u}])$.
The output of the conditional layer is then obtained via element-wise modulation, 
{\small
\begin{equation}
    \widetilde{\boldsymbol{h}}' = \boldsymbol{\gamma_{t, u}} \odot \widetilde{\boldsymbol{h}},
\end{equation}
}allowing the same linear transformation to exhibit different behaviors across diffusion timesteps and downstream tasks. 

\denoisNet\ consists of a sequence of three such conditional layers, each followed by a Softplus activation function, yielding intermediate representations $\boldsymbol{h}'_1, \boldsymbol{h}'_2$, and $\boldsymbol{h}'_3$. Finally, the output of the last conditional layer is used to predict noise through an MLP, i.e., $\hat{\boldsymbol{\epsilon}} = \text{MLP}(\boldsymbol{h}'_3)$, where $\hat{\boldsymbol{\epsilon}}$ serves as the predicted noise.


\subsubsection{Model Training and Inference.}
Given the predicted noise 
$\hat{\boldsymbol{\epsilon}}$, we optimize \diffM\ by minimizing the mean squared error between it and the ground-truth noise $\boldsymbol{\epsilon}$, defined as
\begin{equation} \label{eq:noise_loss}
\mathcal{L}_{noise} = \mathbb{E}
\|\hat{\boldsymbol{\epsilon}} - \boldsymbol{\epsilon}\|^2.
\end{equation}
The detailed training procedure is summarized in Algorithm~\ref{alg:training}.
During inference, we generate new predictions by iteratively denoising samples initialized as $\boldsymbol{y}_T^u \sim \mathcal{N}\!\bigl(\widetilde{\boldsymbol{y}}^u, \boldsymbol{I}\bigr)$ according to Eq.~\ref{eq:reverse}.
The complete inference procedure is summarized in Algorithm~\ref{alg:inference}.

\begin{algorithm}[t]
\caption{\diffM\ Training}
\label{alg:training}
\small
\setstretch{0.6}
\begin{algorithmic}[1]
\State Generate $\widetilde{\boldsymbol{y}}^u$ for task $u$ from \priorGen
\Repeat
    \State Let $u \in \{1,\ldots,U\}$ denote the task index
    \State Draw ${y}_0^u \sim q(\boldsymbol{y}_0^{u} \mid \boldsymbol{h}, u)$
    \State Draw $t \sim \mathrm{Uniform}(\{1,\ldots,T\})$
    \State Draw $\boldsymbol{\epsilon} \sim \mathcal{N}(\mathbf{0}, \boldsymbol{I})$
    \State Generate $\boldsymbol{y}_t^u$ via the forward diffusion process in Eq.~\ref{eq:forward_2}:
    \[
    \boldsymbol{y}_t^u = \sqrt{\bar{\alpha}_t}\, \boldsymbol{y}_0^u
\;+\;
\sqrt{1-\bar{\alpha}_t}\, \boldsymbol{\epsilon}
\;+\;
\bigl(1-\sqrt{\bar{\alpha}_t}\bigr)\, \widetilde{\boldsymbol{y}}^u
    \]
    \State Generate predicted noise: $\hat{\boldsymbol{\epsilon}} = \text{\denoisNet} (\boldsymbol{h}, {y}_t^u, t, u)$
    \State Compute noise estimation loss in Eq.~\ref{eq:noise_loss}
    \State Update parameters by descending:
    $\nabla_\theta \mathcal{L}_{noise}$
\Until{Convergence}
\end{algorithmic}
\end{algorithm}

\begin{algorithm}[t]
\caption{\diffM\ Inference}
\label{alg:inference}
\small
\setstretch{0.6}
\begin{algorithmic}[1]
\State $\boldsymbol{y}_T^u \sim \mathcal{N}\!\bigl(\widetilde{\boldsymbol{y}}^u, \boldsymbol{I}\bigr)$
\For{$t = T$ \textbf{to} $1$}
    \State Generate predicted noise $\hat{\boldsymbol{\epsilon}}$ using \denoisNet
    \State Calculate reparameterized $\hat{\boldsymbol{y}}_0^u$:
    \[
    \hat{\boldsymbol{y}}_0^u
    =
    \frac{1}{\sqrt{\bar{\alpha}_t}}
    \Bigl(
    \boldsymbol{y}_t^u
    -
    (1-\sqrt{\bar{\alpha}_t}) \widetilde{\boldsymbol{y}}^u
    -
    \sqrt{1-\bar{\alpha}_t}\,
    \hat{\boldsymbol{\epsilon}}
    \Bigr)
    \]
    \State \textbf{if} $t > 1$: draw $\boldsymbol{v} \sim \mathcal{N}(\mathbf{0},\boldsymbol{I})$
    \State \hspace{2em}$
    \begin{aligned}
    \boldsymbol{y}_{t-1}^u
    &= \gamma_0 \hat{\boldsymbol{y}}_0^u
    + \gamma_1 \boldsymbol{y}_t^u
    + \gamma_2 \widetilde{\boldsymbol{y}}^u
    + \sqrt{\tilde{\beta}_t}\, \boldsymbol{v}
    \end{aligned}
    $
    \State \textbf{else:}
    \State \hspace{2em}$
    \begin{aligned}
        \boldsymbol{y}_{t-1}^u
        =
        \hat{\boldsymbol{y}}_0^u
    \end{aligned}
    $
\EndFor
\State \Return $\boldsymbol{y}_0^u$
\end{algorithmic}
\end{algorithm}

\section{Experiments}
\label{sec:exp}

We run experiments to verify: 
(\textbf{Q1})~the embedding quality of \model\ as compared with the state-of-the-art (SOTA) models on six downstream tasks under both cross-city and same-city settings,
(\textbf{Q2})~the applicability of \model\ across diverse urban environments,
(\textbf{Q3})~the general applicability of our proposed \diffM\ module when incorporated into existing models,
(\textbf{Q4})~the impact of our model components,
(\textbf{Q5})~the adaptability of \diffM\ to new tasks,
(\textbf{Q6})~prediction quality of \diffM, cross-country applicability of \model, and parameter impact.

\begin{table*}[htbp]
\centering
\caption{Overall Cross-city Prediction Accuracy Results (`$\uparrow$' indicates that larger values are preferred. The best results are in boldface, and the second-best results are underlined. Same for the tables below).}
\label{tab:overall_result}
\setlength{\tabcolsep}{4pt}
\renewcommand{\arraystretch}{1.15}
\resizebox{\textwidth }{!}{
\begin{tabular}{lccccccccc}

\toprule [0.4ex] \\[-3.5ex]

\multirow{2}{*}{\textbf{\makecell[l]{CHI \& SF (X) \\ $\rightarrow$ NYC (Y)}}}
& HREP~\cite{HREP}
& RegionDCL~\cite{RegionDCL}
& UrbanCLIP~\cite{urbanclip}
& CityFM~\cite{cityFM}
& GeoHG~\cite{GeoHG}
& GURPP~\cite{GURPP}
& FlexiReg~\cite{FlexiReg}
& \textbf{\model}
& \multirow{2}{*}{\textbf{Improvement}} \\

\cmidrule(lr){2-2} \cmidrule(lr){3-3} \cmidrule(lr){4-4} \cmidrule(lr){5-5} \cmidrule(lr){6-6} \cmidrule(lr){7-7} \cmidrule(lr){8-8}   \cmidrule(lr){9-9}
    
& $R^{2} \uparrow$ & $R^{2} \uparrow$ & $R^{2} \uparrow$ & $R^{2} \uparrow$ & $R^{2}  \uparrow$ & $R^{2} \uparrow$ & $R^{2} \uparrow$ & $R^{2} \uparrow$ \\
\hline

Crime
& 0.301$\,\pm\,$0.037
& 0.211$\,\pm\,$0.028
& 0.241$\,\pm\,$0.005
& 0.297$\,\pm\,$0.017
& 0.339$\,\pm\,$0.047
& 0.545$\,\pm\,$0.039
& \underline{0.663$\,\pm\,$0.024}
& \textbf{0.724$\,\pm\,$0.008}
& \red{\textbf{+9.20\%}} \\

Check-in
& 0.478$\,\pm\,$0.041
& 0.434$\,\pm\,$0.014
& 0.426$\,\pm\,$0.031
& 0.445$\,\pm\,$0.022
& 0.329$\,\pm\,$0.058
& 0.716$\,\pm\,$0.021
& \underline{0.754$\,\pm\,$0.009}
& \textbf{0.781$\,\pm\,$0.012}
& \red{\textbf{+3.58\%}} \\

Service Call
& 0.173$\,\pm\,$0.031
& 0.071$\,\pm\,$0.016
& 0.155$\,\pm\,$0.012
& 0.124$\,\pm\,$0.011
& 0.231$\,\pm\,$0.039
& 0.247$\,\pm\,$0.055
& \underline{0.513$\,\pm\,$0.031}
& \textbf{0.589$\,\pm\,$0.021}
& \red{\textbf{+14.81\%}} \\

Population
& 0.262$\,\pm\,$0.044
& 0.185$\,\pm\,$0.010
& 0.224$\,\pm\,$0.011
& 0.219$\,\pm\,$0.009
& 0.206$\,\pm\,$0.022
& 0.477$\,\pm\,$0.024
& \underline{0.569$\,\pm\,$0.018}
& \textbf{0.626$\,\pm\,$0.016}
& \red{\textbf{+10.02\%}} \\

Carbon
& 0.136$\,\pm\,$0.026
& 0.025$\,\pm\,$0.016
& 0.051$\,\pm\,$0.033
& 0.085$\,\pm\,$0.036
& 0.139$\,\pm\,$0.057
& 0.009$\,\pm\,$0.024
& \underline{0.324$\,\pm\,$0.022}
& \textbf{0.389$\,\pm\,$0.027}
& \red{\textbf{+20.06\%}} \\

Nightlight
& 0.028$\,\pm\,$0.027
& 0.011$\,\pm\,$0.029
& 0.324$\,\pm\,$0.022
& 0.026$\,\pm\,$0.014
& 0.137$\,\pm\,$0.023
& -0.042$\,\pm\,$0.063
& \underline{0.442$\,\pm\,$0.011}
& \textbf{0.492$\,\pm\,$0.013}
& \red{\textbf{+11.31\%}} \\

\bottomrule \toprule [0.4ex] \\[-3.5ex]

\multirow{2}{*}{\textbf{\makecell[l]{NYC \& SF (X) \\ $\rightarrow$ CHI (Y)}}}
& HREP~\cite{HREP}
& RegionDCL~\cite{RegionDCL}
& UrbanCLIP~\cite{urbanclip}
& CityFM~\cite{cityFM}
& GeoHG~\cite{GeoHG}
& GURPP~\cite{GURPP}
& FlexiReg~\cite{FlexiReg}
& \textbf{\model}
& \multirow{2}{*}{\textbf{Improvement}} \\

\cmidrule(lr){2-2} \cmidrule(lr){3-3} \cmidrule(lr){4-4} \cmidrule(lr){5-5} \cmidrule(lr){6-6} \cmidrule(lr){7-7} \cmidrule(lr){8-8}   \cmidrule(lr){9-9}

& $R^{2} \uparrow$ & $R^{2} \uparrow$ & $R^{2} \uparrow$ & $R^{2} \uparrow$ & $R^{2}  \uparrow$ & $R^{2} \uparrow$ & $R^{2} \uparrow$ & $R^{2} \uparrow$ \\
\hline

Crime
& 0.402$\,\pm\,$0.037
& 0.159$\,\pm\,$0.015
& 0.411$\,\pm\,$0.010
& 0.277$\,\pm\,$0.032
& 0.442$\,\pm\,$0.026
& 0.477$\,\pm\,$0.024
& \underline{0.708$\,\pm\,$0.027}
& \textbf{0.769$\,\pm\,$0.018}
& \red{\textbf{+8.62\%}} \\

Check-in
& 0.544$\,\pm\,$0.045
& 0.416$\,\pm\,$0.007
& 0.087$\,\pm\,$0.007
& 0.534$\,\pm\,$0.011
& 0.343$\,\pm\,$0.045
& 0.289$\,\pm\,$0.063
& \underline{0.729$\,\pm\,$0.018}
& \textbf{0.802$\,\pm\,$0.032}
& \red{\textbf{+10.01\%}} \\

Service Call
& 0.458$\,\pm\,$0.041
& 0.363$\,\pm\,$0.012
& 0.462$\,\pm\,$0.017
& 0.291$\,\pm\,$0.005
& 0.391$\,\pm\,$0.021
& 0.396$\,\pm\,$0.028
& \underline{0.669$\,\pm\,$0.019}
& \textbf{0.713$\,\pm\,$0.021}
& \red{\textbf{+6.58\%}} \\

Population
& 0.327$\,\pm\,$0.044
& 0.259$\,\pm\,$0.009
& 0.211$\,\pm\,$0.006
& 0.264$\,\pm\,$0.018
& 0.296$\,\pm\,$0.031
& 0.484$\,\pm\,$0.019
& \underline{0.733$\,\pm\,$0.022}
& \textbf{0.810$\,\pm\,$0.016}
& \red{\textbf{+10.50\%}} \\

Carbon
& 0.065$\,\pm\,$0.033
& -0.002$\,\pm\,$0.024
& 0.094$\,\pm\,$0.057
& 0.055$\,\pm\,$0.008
& 0.182$\,\pm\,$0.027
& 0.112$\,\pm\,$0.046
& \underline{0.467$\,\pm\,$0.034}
& \textbf{0.531$\,\pm\,$0.027}
& \red{\textbf{+13.70\%}} \\

Nightlight
& 0.159$\,\pm\,$0.056
& 0.127$\,\pm\,$0.017
& 0.721$\,\pm\,$0.011
& 0.074$\,\pm\,$0.011
& 0.093$\,\pm\,$0.027
& 0.110$\,\pm\,$0.019
& \underline{0.859$\,\pm\,$0.011}
& \textbf{0.891$\,\pm\,$0.023}
& \red{\textbf{+3.73\%}} \\

\bottomrule \toprule [0.4ex] \\[-3.5ex]

\multirow{2}{*}{\textbf{\makecell[l]{NYC \& CHI (X) \\ $\rightarrow$ SF (Y)}}}
& HREP~\cite{HREP}
& RegionDCL~\cite{RegionDCL}
& UrbanCLIP~\cite{urbanclip}
& CityFM~\cite{cityFM}
& GeoHG~\cite{GeoHG}
& GURPP~\cite{GURPP}
& FlexiReg~\cite{FlexiReg}
& \textbf{\model}
& \multirow{2}{*}{\textbf{Improvement}} \\

\cmidrule(lr){2-2} \cmidrule(lr){3-3} \cmidrule(lr){4-4} \cmidrule(lr){5-5} \cmidrule(lr){6-6} \cmidrule(lr){7-7} \cmidrule(lr){8-8}   \cmidrule(lr){9-9}

& $R^{2} \uparrow$ & $R^{2} \uparrow$ & $R^{2} \uparrow$ & $R^{2} \uparrow$ & $R^{2}  \uparrow$ & $R^{2} \uparrow$ & $R^{2} \uparrow$ & $R^{2} \uparrow$ \\
\hline

Crime
& 0.491$\,\pm\,$0.036
& 0.455$\,\pm\,$0.012
& 0.252$\,\pm\,$0.022
& 0.375$\,\pm\,$0.018
& 0.234$\,\pm\,$0.041
& 0.513$\,\pm\,$0.031
& \underline{0.599$\,\pm\,$0.017}
& \textbf{0.814$\,\pm\,$0.008}
& \red{\textbf{+35.89\%}} \\

Check-in
& 0.429$\,\pm\,$0.041
& 0.361$\,\pm\,$0.014
& 0.379$\,\pm\,$0.031
& 0.318$\,\pm\,$0.022
& 0.223$\,\pm\,$0.058
& \underline{0.728$\,\pm\,$0.021}
& 0.723$\,\pm\,$0.009
& \textbf{0.783$\,\pm\,$0.012}
& \red{\textbf{+7.56\%}} \\

Service Call
& 0.303$\,\pm\,$0.042
& 0.280$\,\pm\,$0.007
& 0.171$\,\pm\,$0.012
& 0.376$\,\pm\,$0.019
& 0.199$\,\pm\,$0.046
& 0.268$\,\pm\,$0.046
& \underline{0.476$\,\pm\,$0.015}
& \textbf{0.565$\,\pm\,$0.021}
& \red{\textbf{+18.70\%}} \\

Population
& 0.183$\,\pm\,$0.031
& -0.003$\,\pm\,$0.011
& -0.414$\,\pm\,$0.021
& 0.019$\,\pm\,$0.021
& 0.106$\,\pm\,$0.031
& 0.067$\,\pm\,$0.036
& \underline{0.431$\,\pm\,$0.011}
& \textbf{0.601$\,\pm\,$0.015}
& \red{\textbf{+39.44\%}} \\

Carbon
& 0.168$\,\pm\,$0.032
& 0.061$\,\pm\,$0.017
& 0.237$\,\pm\,$0.012
& 0.078$\,\pm\,$0.013
& 0.184$\,\pm\,$0.031
& 0.222$\,\pm\,$0.037
& \underline{0.584$\,\pm\,$0.025}
& \textbf{0.665$\,\pm\,$0.017}
& \red{\textbf{+13.70\%}} \\

Nightlight
& 0.270$\,\pm\,$0.038
& 0.156$\,\pm\,$0.011
& 0.644$\,\pm\,$0.027
& 0.072$\,\pm\,$0.015
& 0.085$\,\pm\,$0.036
& 0.392$\,\pm\,$0.039
& \underline{0.799$\,\pm\,$0.011}
& \textbf{0.845$\,\pm\,$0.012}
& \red{\textbf{+5.76\%}} \\

\bottomrule
\end{tabular}
}
\end{table*}

\subsection{Experimental Settings}
\label{subsec:exp_settings}
\textbf{Dataset.}
We use data from three cities from : New York City (\textbf{NYC})~\cite{nycOpendata}, Chicago (\textbf{CHI})~\cite{chiOpendata}, and San  (\textbf{SF})~\cite{sfOpendata}. For each city, we collect data on region division, POI information, and six downstream tasks (crime, check-in, service call, population, carbon emission, and nightlight). 
Additional details are provided in Appendix~\ref{subsec:appendix_of_dataset_statistics}.

\textbf{Competitors.}
We compare with seven models including two SOTA models: \textbf{HREP}~\cite{HREP}, \textbf{RegionDCL}~\cite{RegionDCL}, \textbf{UrbanCLIP}~\cite{urbanclip}, \textbf{CityFM}~\cite{cityFM}, \textbf{GeoHG}~\cite{GeoHG}, \textbf{GURPP} (SOTA)~\cite{GURPP}, and \textbf{FlexiReg} (SOTA)~\cite{FlexiReg}. Appendix~\ref{subsec:appendix_of_baselines} details these models. Implementation details and hyperparameter settings of all models tested (including \model) are in Appendix~\ref{subsec:appendix_of_hypepara_settings}.

\textbf{Evaluation procedure.} 
We evaluate representation learning models under a cross-city setting by training each model on data from all but one city (i.e., the target city) and applying it to the held-out target city to generate region embeddings, which are then used for downstream prediction for the target city.
We consider six downstream tasks: crime, check-in, service call, population, carbon emission, and nightlight counts. 
\model\ employs a single jointly trained downstream predictor for all tasks, whereas baseline methods require task-specific predictors.
We assess the quality of region embeddings through downstream task performance using mean absolute error (\textbf{MAE}), root mean square error (\textbf{RMSE}), and coefficient of determination ($\boldsymbol{R^2}$).

\subsection{Overall Results (Q1)}
\label{subsec:overall_results}

Table~\ref{tab:overall_result} reports the overall cross-city prediction accuracy, where $\mathbf{X}$ denotes the source cities and $\mathbf{Y}$ denotes the target city. Here, we only report $R^2$ for conciseness, as the performance in MAE and RMSE resembles (same below). Full results can be found in Appendix~\ref{sec:appendix_of_cross_city_result}. 
We make the following observations.

(1)~Our model \model\ outperforms all competitors across three target cities and all six downstream tasks, improving $R^2$ by up to 35.89\% over the best baseline FlexiReg. This is attributed to our novel model design:
(i)~Our random-walk–based embedding learning module reduces sensitivity to city-specific global structures while capturing transferable local structural and functional patterns, thereby improving cross-city generalization.
(ii)~Our diffusion-based downstream learning module jointly models multiple tasks via conditioning mechanisms, enabling the sharing of knowledge across tasks compared to independent task learning.

(2)~Baselines that adapt region embeddings to downstream tasks via prompting (HREP, GURPP, and FlexiReg) generally outperform the others, as task-specific prompts inject task-relevant information and enable more effective representation adaptation.
Nevertheless, these methods are still outperformed by \model, as their city-centric embedding learning suffers when the models are learned for one city and applied to another to geenerate the embeddings.

(3)~The remaining baselines are inferior for the following reasons. 
RegionDCL relies on building footprints, which provide limited semantic information on regional functionalities.
UrbanCLIP uses satellite imagery and vision–language model-generated textual descriptions. Its effectiveness is constrained by the quality of the generated text, which may suffer from hallucination. 
CityFM and GeoHG adopt contrastive learning among geospatial entities or regions within a single city. 
Because of the city-level contrastive objectives, the resulting region embeddings tend to encode city-specific relationships and fail the generalize across cities.

\paragraph{Same-city Results.}
Since all baseline models were designed for training and testing on the same city, we also report  results under such same-city settings in Table~\ref{tab:same_city_full_p1} and Table~\ref{tab:same_city_full_p2} in Appendix~\ref{sec:appendix_of_same_city_result}.
The SOTA model FlexiReg is strong in such settings, while \model\ is on par or even more accurate than FlexiReg, even though \model\ uses much fewer features (i.e., only POIs and region neighborhood features vs. POIs, region neighborhood features, satellite images, street view images, and LLM-based region descriptions). 
This demonstrates the applicability of \model\ across different training and test settings.  




\subsection{Applicability to Suburban Areas (Q2)}
\label{subsec:suburban_area}

We further evaluate the applicability of \model\ across diverse urban environments with regions in Staten Island. Staten Island is a relatively underdeveloped area of New York City and differs substantially from Manhattan, which was used in the experiments above. All experiments use the same protocol as in Manhattan.

\begin{table}[htbp]
\captionsetup{justification=centering}
\caption{Model Applicability to Suburban Areas.}
\label{tab:suburban_crossCity}
\setlength{\tabcolsep}{2pt}
\renewcommand{\arraystretch}{1.0}
\resizebox{\columnwidth}{!}{
\begin{tabular}{l | c c c c}
\toprule
\multirow{2}{*}{\textbf{\makecell[l]{CHI \& SF (X) $\rightarrow$ \\ Staten Island (Y)}}}
& \textbf{Check-in}
& \textbf{Population}
& \textbf{Carbon}
& \textbf{Nightlight} \\

\cmidrule(lr){2-2} \cmidrule(lr){3-3} \cmidrule(lr){4-4} \cmidrule(lr){5-5} 

& $R^{2} \uparrow$ & $R^{2} \uparrow$ & $R^{2} \uparrow$ & $R^{2} \uparrow$ \\
\midrule

HREP~\cite{HREP}
& 0.148$\,\pm\,$0.031
& 0.181$\,\pm\,$0.024
& 0.136$\,\pm\,$0.038
& 0.226$\,\pm\,$0.033 \\

RegionDCL~\cite{RegionDCL}
& 0.098$\,\pm\,$0.027
& 0.056$\,\pm\,$0.018
& 0.106$\,\pm\,$0.038
& 0.394$\,\pm\,$0.023 \\

UrbanCLIP~\cite{urbanclip}
& 0.061$\,\pm\,$0.019
& 0.092$\,\pm\,$0.014
& 0.514$\,\pm\,$0.018
& 0.577$\,\pm\,$0.019 \\

CityFM~\cite{cityFM}
& 0.023$\,\pm\,$0.067
& 0.073$\,\pm\,$0.017
& 0.013$\,\pm\,$0.023
& 0.063$\,\pm\,$0.026 \\

GeoHG~\cite{GeoHG}
& 0.057$\,\pm\,$0.022
& 0.027$\,\pm\,$0.015
& 0.051$\,\pm\,$0.038
& 0.077$\,\pm\,$0.027 \\

GURPP~\cite{GURPP}
& 0.145$\,\pm\,$0.054
& 0.358$\,\pm\,$0.035
& 0.319$\,\pm\,$0.045
& 0.446$\,\pm\,$0.029 \\

FlexiReg~\cite{FlexiReg}
& \underline{0.407$\,\pm\,$0.027}
& \underline{0.389$\,\pm\,$0.019}
& \underline{0.609$\,\pm\,$0.017}
& \underline{0.869$\,\pm\,$0.010} \\

\midrule
\textbf{\model}
& \textbf{0.533$\,\pm\,$0.031}
& \textbf{0.603$\,\pm\,$0.025}
& \textbf{0.781$\,\pm\,$0.005}
& \textbf{0.945$\,\pm\,$0.005} \\

\midrule
\textbf{Improvement}
& \textbf{\red{+31.0\%}}
& \textbf{\red{+55.0\%}}
& \textbf{\red{+28.2\%}}
& \textbf{\red{+8.7\%}} \\

\bottomrule
\end{tabular}
}
\end{table}

\begin{table*}[t]
\centering
\caption{Prediction Accuracy Results When Powering Existing Models with Our \diffM\ Module (NYC).}
\label{tab:app_diffusionM_partial}

\setlength{\tabcolsep}{3pt}
\renewcommand{\arraystretch}{1.1}
\resizebox{\textwidth }{!}{
\begin{tabular}{l |cl | cl | cl | cl}

\toprule [0.4ex] \\[-3.5ex]

\multirow{2}{*}{\textbf{NYC}}
& \multicolumn{1}{c}{HREP}
& \multicolumn{1}{c|}{\textbf{HREP-DiffCT}}
& \multicolumn{1}{c}{UrbanCLIP}
& \multicolumn{1}{c|}{\textbf{UrbanCLIP-DiffCT}}
& \multicolumn{1}{c}{HAFusion}
& \multicolumn{1}{c|}{\textbf{HAFusion-DiffCT}}
& \multicolumn{1}{c}{GURPP}
& \multicolumn{1}{c}{\textbf{GURPP-DiffCT}}\\

\cmidrule(lr){2-2} \cmidrule(lr){3-3} \cmidrule(lr){4-4} \cmidrule(lr){5-5} \cmidrule(lr){6-6} \cmidrule(lr){7-7} \cmidrule(lr){8-8}   \cmidrule(lr){9-9}
    
& \multicolumn{1}{c}{$R^2\uparrow$}
& \multicolumn{1}{c|}{$R^2\uparrow$}
& \multicolumn{1}{c}{$R^2\uparrow$}
& \multicolumn{1}{c|}{$R^2\uparrow$}
& \multicolumn{1}{c}{$R^2\uparrow$}
& \multicolumn{1}{c|}{$R^2\uparrow$}
& \multicolumn{1}{c}{$R^2\uparrow$}
& \multicolumn{1}{c}{$R^2\uparrow$} \\

\hline

Crime
& 0.681$\pm$0.014
& \textbf{0.784$\pm$0.013} \,(\red{+15.1\%})
& 0.267$\pm$0.012
& \textbf{0.471$\pm$0.039} \,(\red{+76.4\%})
& 0.734$\pm$0.015
& \textbf{0.823$\pm$0.011} \,(\red{+12.1\%})
& 0.589$\pm$0.039
& \textbf{0.722$\pm$0.029} \,(\red{+22.6\%}) \\

Check-in
& 0.700$\pm$0.022
& \textbf{0.850$\pm$0.016} \,(\red{+21.4\%})
& 0.458$\pm$0.005
& \textbf{0.642$\pm$0.025} \,(\red{+40.2\%})
& 0.844$\pm$0.012
& \textbf{0.900$\pm$0.009} \,(\red{+6.6\%})
& 0.757$\pm$0.019
& \textbf{0.837$\pm$0.017} \,(\red{+10.6\%}) \\

Service Call
& 0.398$\pm$0.019
& \textbf{0.589$\pm$0.014} \,(\red{+49.0\%})
& 0.232$\pm$0.005
& \textbf{0.369$\pm$0.033} \,(\red{+59.1\%})
& 0.493$\pm$0.014
& \textbf{0.667$\pm$0.006} \,(\red{+35.3\%})
& 0.405$\pm$0.048
& \textbf{0.528$\pm$0.009} \,(\red{+30.4\%}) \\

Population
& 0.571$\pm$0.021
& \textbf{0.667$\pm$0.019} \,(\red{+16.8\%})
& 0.276$\pm$0.002
& \textbf{0.392$\pm$0.024} \,(\red{+42.0\%})
& 0.616$\pm$0.019
& \textbf{0.692$\pm$0.019} \,(\red{+12.3\%})
& 0.381$\pm$0.026
& \textbf{0.420$\pm$0.021} \,(\red{+10.2\%}) \\

Carbon
& 0.184$\pm$0.050
& \textbf{0.373$\pm$0.036} \,(\red{+102.7\%})
& 0.113$\pm$0.013
& \textbf{0.238$\pm$0.033} \,(\red{+110.6\%})
& 0.189$\pm$0.019
& \textbf{0.358$\pm$0.017} \,(\red{+89.4\%})
& 0.021$\pm$0.001
& \textbf{0.204$\pm$0.019} \,(\red{+871.4\%}) \\

Nightlight
& -0.026$\pm$0.024
& \textbf{0.167$\pm$0.011} \,(\red{+742.3\%})
& 0.337$\pm$0.019
& \textbf{0.434$\pm$0.026} \,(\red{+28.8\%})
& 0.035$\pm$0.013
& \textbf{0.171$\pm$0.026} \,(\red{+388.6\%})
& -0.012$\pm$0.005
& \textbf{0.160$\pm$0.009} \,(\red{+1433.3\%}) \\

\bottomrule
\end{tabular}
}
\end{table*}

We report the $R^2$ results in Table~\ref{tab:suburban_crossCity} for the check-in, population, carbon, and nightlight count prediction tasks. Crime and service call data are unavailable for Staten Island.
\model\ achieves even larger improvements, up to 55\%, on Staten Island than on Manhattan. 
This gain stems from \model’s region-centric design, which reduces sensitivity to city-specific structural variations. 
These findings further demonstrate \model’s ability to transfer knowledge from data-rich cities to data-scarce urban areas.

\paragraph{Same-city Results} 
We report the same-city results on Staten Island in Table~\ref{tab:suburban_sameCity} in Appendix~\ref{sec:appendix_of_suburban_area}. 
Even under this more favorable setting, the baseline models perform poorly due to severe data scarcity. This observation indicates that city-specific models struggle in data-limited scenarios, further highlighting the importance of cross-city modeling and the advantages of our solution.




\subsection{Applicability of \diffM\ (Q3)}
\label{subsec:app_of_diffusion}

To show the general applicability of the heterogeneous conditional diffusion-based cross-task learning (\diffM) module, we integrate it with four baseline models, i.e., HREP, UrbanCLIP, HAFusion, and GURPP. We denote the resulting models as \textbf{HREP-DiffCT}, \textbf{UrbanCLIP-DiffCT}, \textbf{HAFusion-DiffCT}, and \textbf{GURPP-DiffCT}.

Since these baseline models are originally designed for the same-city setting, we evaluate all variants under this setting using six downstream tasks. Following the same experimental protocol as above, a single predictor is trained jointly for all downstream tasks. The results in Table~\ref{tab:app_diffusionM_partial} show that \diffM\ consistently enhances the performance of all baseline models (full results can be found in Appendix~\ref{sec:appendix_of_app_diffusionM}). Compared with the vanilla models, the \diffM-enhanced variants achieve accuracy improvements of up to 1433.3\%. 
These results confirm both the effectiveness and the applicability of \diffM, and further demonstrate its ability to enhance model accuracy  under the same-city setting.

Note that these baseline models generate region embeddings based on diverse region features and training strategies. The consistent gains indicate that \diffM\ is model- and feature-agnostic. It can consistently improve downstream task effectiveness regardless of how the underlying region embeddings are generated.


\subsection{Ablation Study (Q4)}
\label{subsec:ablation_study}

We study the effectiveness of model components with the following variants: 
(i)~\textbf{\model-w/o-RL+CL} replaces the reconstruction loss for input features of randomly masked nodes with contrastive loss;
(ii)~\textbf{\model-w/o-EM+CA} replaces the element-wise modulation with a cross-attention mechanism to integrate task embeddings $\boldsymbol{\gamma_u}$ into the denoising network;
(iii)~\textbf{\model}\textbf{-w/o-EM+C} replaces the element-wise modulation with direct concatenation of task embeddings $\boldsymbol{\gamma_u}$ to the denoising network input;
(iv)~\textbf{\model-w/o-Retr} randomly selects top-$K$ regions from the information repository instead of retrieving them based on region embedding similarity;
(v)~\textbf{\model-w/o-Prior} replaces the generated prior knowledge with a standard Gaussian prior in the diffusion process; 
and (vi)~\textbf{\model-w/o-DiffM} removes the diffusion module from the downstream task learning stage. 

\begin{figure}[htbp]
    \centering
    \begin{subfigure}[b]{\columnwidth}
        \centering
        \includegraphics[width=\columnwidth]{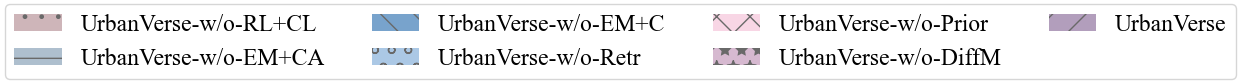}
    \end{subfigure}
    
    \begin{subfigure}[b]{\columnwidth}
        \centering
        \includegraphics[width=\columnwidth]{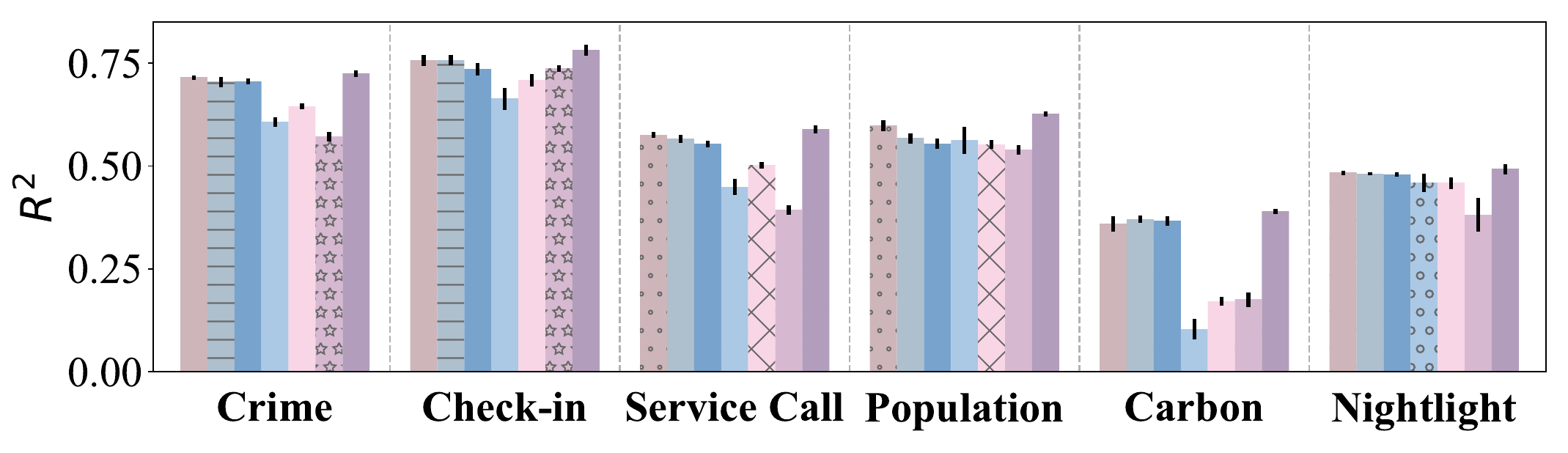}
    \end{subfigure} 

    \caption{Ablation study results (NYC).}
    \label{fig:ablation_study}
\end{figure}

We again repeat the experiments and present the results in Fig.~\ref{fig:ablation_study}.
As expected, \model\ consistently outperforms all  variants, underscoring the contribution of each component to the overall effectiveness of \model. There are further observations:

(1)~\model-w/o-DiffM consistently performs the worst across all tasks. This suggests that modeling the full conditional distribution of the target variable, rather than producing a single point estimate, is helpful for accurate urban prediction. Moreover, diffusion models naturally capture uncertainty and multiple plausible outcomes, leading to more robust and reliable predictions.

(2)~\model-w/o-Prior performs worse than \model, highlighting the importance of incorporating prior knowledge that captures the relationship between region embeddings and downstream task values to effectively guide the denoising process. Meanwhile, \model-w/o-Retr underperforms \model-w/o-Prior in most cases, indicating that naively incorporated prior knowledge may introduce noise and fail to improve performance.
%

(3)~\model-w/o-LM+C and \model-w/o-LM+CA are both outperformed by \model, indicating that element-wise modulation is more effective than direct concatenation or cross-attention for injecting task embeddings as conditional information. This design provides more stable and noise-robust conditioning while avoiding potential overfitting of attention-based mechanisms.




\subsection{Adaptability to New Downstream Tasks (Q5)}
\label{subsec:adaptability_to_new_tasks}


As \diffM\ is designed to support multiple downstream tasks, we further evaluate its adaptability to new tasks under two training strategies: (i)~\textbf{Fine-tuning:} \diffM\ is first trained on five of the six tasks and then adapted to the remaining target task using new data from the task;
and (ii)~\textbf{Training from scratch:} \diffM\ is retrained using data from all tasks, including the target task.

\begin{figure}[htbp]
  \centering
  \includegraphics[width=\columnwidth,]{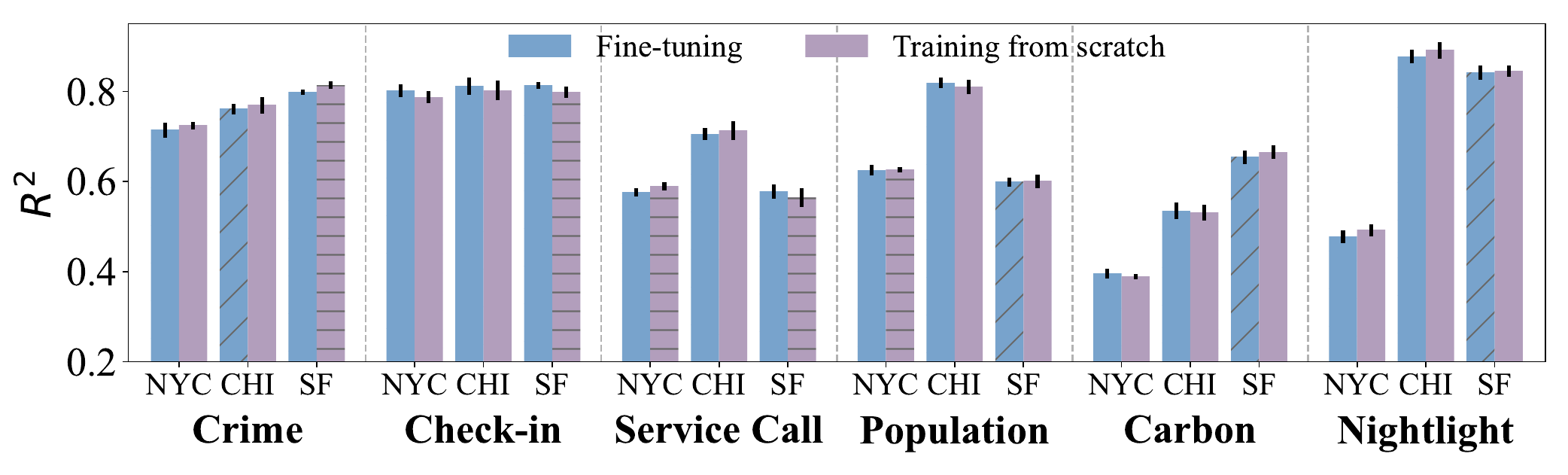}
  \caption{Adaptability to new downstream tasks.}
  \label{fig:adapt_new_task}
\end{figure}

Overall, as shown in Fig~\ref{fig:adapt_new_task}, the accuracy gap between the two strategies is marginal, i.e., below 0.02 in $R^2$, while fine-tuning requires lower computational cost than a full retrain.
These results highlight the strong adaptability of \diffM\ to new downstream tasks. It can efficiently accommodate new tasks via lightweight fine-tuning while preserving high accuracy.

\subsection{Additional Results (Q6)}
\label{subsec:additional_exp}
We conduct additional experiments with results reported in Appendix~\ref{sec:additional_exp_results}, including model efficiency, prediction quality of \diffM\ module, cross-country applicability of \model,
 and model sensitivity to key hyper-parameters.

\section{Conclusion}
\label{sec:conclusion}

We proposed a model named \model\ to generalize urban representation learning beyond city- and task-specific settings, towards a foundation-style model for urban analytics. 
For cross-city generalization, we proposed \RWM, 
which shifts the model design from city-centric to region-centric. Regions are represented as graph nodes, and random walks are used to construct region sequences that encode local and neighborhood structural information.
For cross-task generalization, we proposed \diffM, 
which jointly models multiple downstream  prediction tasks by conditioning on region-specific prior knowledge and task semantics.
Extensive experiments on three real-world datasets show that \model\ outperforms SOTA models by up to 35.89\% in prediction accuracy under cross-city settings. Moreover, integrating our \diffM\ module into existing models yields consistency improvements in prediction accuracy under same-city settings.

\bibliographystyle{ACM-Reference-Format}
\bibliography{ref}

@incollection{davis2011remarks,
  title={Remarks on Some Nonparametric Estimates of a Density Function},
  author={Davis, Richard A. and Lii, Keh-Shin and Politis, Dimitris N.},
  booktitle={Selected Works of Murray Rosenblatt},
  pages={95--100},
  year={2011},
  publisher={Springer}
}

@online{nycOpendata,
  author = {{New York Dataset}},
  title = {\url{https://opendata.cityofnewyork.us/}},
year = {2020}
}

@online{chiOpendata,
  author = {{Chicago Dataset}},
  title = {https://data.cityofchicago.org/},
year = {2020}
}

@online{sfOpendata,
  author = {{San Francisco Dataset}},
  title = {\url{https://datasf.org/opendata/}},
year = {2020}
}

@inproceedings{HREP,
  title={{Heterogeneous Region Embedding with Prompt Learning}},
  author={Zhou, Silin and He, Dan and Chen, Lisi and Shang, Shuo and Han, Peng},
  booktitle={AAAI},
  pages={4981--4989},
  year={2023}
}

@inproceedings{cityFM,
  title={{City Foundation Models for Learning General Purpose Representations from OpenStreetMap}},
  author={Balsebre, Pasquale and Huang, Weiming and Cong, Gao and Li, Yi},
  booktitle={CIKM},
  pages = {87--97},
  year={2024}
}

@inproceedings{RegionDCL,
  title={{Urban Region Representation Learning with OpenStreetMap Building Footprints}},
  author={Li, Yi and Huang, Weiming and Cong, Gao and Wang, Hao and Wang, Zheng},
  booktitle={KDD},
  pages={1363--1373},
  year={2023}
}

@inproceedings{MGFN,
  title={{Multi-Graph Fusion Networks for Urban Region Embedding}},
  author={Wu, Shangbin and Yan, Xu and Fan, Xiaoliang and Pan, Shirui and Zhu, Shichao and Zheng, Chuanpan and Cheng, Ming and Wang, Cheng},
  booktitle={IJCAI},
  pages={2312--2318},
  year={2022}
}

@inproceedings{MVURE,
  title={{Multi-View Joint Graph Representation Learning for Urban Region Embedding}},
  author={Zhang, Mingyang and Li, Tong and Li, Yong and Hui, Pan},
  booktitle={IJCAI},
  pages={4431--4437},
  year={2020}
}

@inproceedings{urbanclip,
  title={{UrbanCLIP: Learning Text-Enhanced Urban Region Profiling with Contrastive Language-Image Pretraining from the Web}},
  author={Yan, Yibo and Wen, Haomin and Zhong, Siru and Chen, Wei and Chen, Haodong and Wen, Qingsong and Zimmermann, Roger and Liang, Yuxuan},
  booktitle={WWW},
  pages={4006--4017},
  year={2024}
}

@inproceedings{ReCP,
  title={Urban Region Embedding via Multi-View Contrastive Prediction},
  author={Li, Zechen and Huang, Weiming and Zhao, Kai and Yang, Min and Gong, Yongshun and Chen, Meng},
  booktitle={AAAI},
  pages={8724--8732},
  year={2024}
}

@inproceedings{GeoHG,
  title={{Space-aware Socioeconomic Indicator Inference with Heterogeneous Graphs}},
  author={Zou, Xingchen and Huang, Jiani and Hao, Xixuan and Yang, Yuhao and Wen, Haomin and Yan, Yibo and Huang, Chao and Chen, Chao and Liang, Yuxuan},
  booktitle={SIGSPATIAL},
  pages={244--256},
  year={2025}
}

@inproceedings{GURPP,
  title={{Urban Region Pre-training and Prompting: A Graph-based Approach}},
  author={Jin, Jiahui and Song, Yifan and Kan, Dong and Zhu, Haojia and Sun, Xiangguo and Li, Zhicheng and Sun, Xigang and Zhang, Jinghui},
  booktitle={KDD},
  pages={1071--1082},
  year={2025}
}

@inproceedings{FlexiReg,
  title={FlexiReg: Flexible Urban Region Representation Learning},
  author={Sun, Fengze and Chang, Yanchuan and Tanin, Egemen and Karunasekera, Shanika and Qi, Jianzhong},
  booktitle={KDD},
  pages={2702--2713},
  year={2025}
}

@online{osm,
  author = {{OpenStreetMap}},
  title = {\url{https://www.openstreetmap.org/}},
year = {2024}
}

@online{checkinData,
  author = {{Foursquare Dataset}},
  title = {\url{https://sites.google.com/site/yangdingqi/home/foursquare-dataset}},
year = {2015}
}

@online{WorldPop,
  author = {{WorldPop}},
  title = {\url{https://www.worldpop.org/}},
year = {2024}
}

@online{ODIAC,
  author = {{ODIAC}},
  title = {\url{https://db.cger.nies.go.jp/dataset/ODIAC/}},
year = {2025}
}

@dataset{nightlight,
  author    = {Li, Xuecao and Zhou, Yuyu and Zhao, Min and Zhao, Xia},
  title     = {Harmonization of DMSP and VIIRS Nighttime Light Data from 1992--2020 at the Global Scale},
  year      = {2020},
  publisher = {figshare},
  doi       = {10.6084/m9.figshare.9828827.v5}
}

@inproceedings{HGT,
  title={{Heterogeneous Graph Transformer}},
  author={Hu, Ziniu and Dong, Yuxiao and Wang, Kuansan and Sun, Yizhou},
  booktitle={WWW},
  pages={2704--2710},
  year={2020}
}

@inproceedings{grover2016node2vec,
  title={{Node2vec: Scalable Feature Learning for Networks}},
  author={Grover, Aditya and Leskovec, Jure},
  booktitle={KDD},
  pages={855--864},
  year={2016}
}

@inproceedings{devlin2019bert,
  title={{BERT: Pre-training of Deep Bidirectional Transformers for Language Understanding}},
  author={Devlin, Jacob and Chang, Ming-Wei and Lee, Kenton and Toutanova, Kristina},
  booktitle={NAACL-HLT},
  pages={4171--4186},
  year={2019}
}

@inproceedings{attention,
  title={{Attention is All You Need}},
  author={Vaswani, Ashish and Shazeer, Noam and Parmar, Niki and Uszkoreit, Jakob and Jones, Llion and Gomez, Aidan N and Kaiser, {\L}ukasz and Polosukhin, Illia},
  booktitle={NeurIPS},
  pages={6000--6010},
  year={2017}
}

@inproceedings{ho2020denoising,
    author      = {Ho, Jonathan and Jain, Ajay and Abbeel, Pieter},
    booktitle   = {NeurIPS},
    pages       = {6840--6851},
    title       = {{Denoising Diffusion Probabilistic Models}},
    year        = {2020}
}

@inproceedings{han2022card,
  title={{CARD: Classification and Regression Diffusion Models}},
  author={Han, Xizewen and Zheng, Huangjie and Zhou, Mingyuan},
  booktitle={NeurIPS},
  pages={18100--18115},
  year={2022}
}

@article{zheng2014urban,
  title={{Urban Computing: Concepts, Methodologies, and Applications}},
  author={Zheng, Yu and Capra, Licia and Wolfson, Ouri and Yang, Hai},
  journal={ACM Transactions on Intelligent Systems and Technology},
  volume={5},
  number={3},
  pages={1--55},
  year={2014}
}

@inproceedings{zhang2017urban,
  title={{Deep Spatio-Temporal Residual Networks for Citywide Crowd Flows Prediction}},
  author={Zhang, Junbo and Zheng, Yu and Qi, Dekang},
  booktitle={AAAI},
  pages={1655--1661},
  year={2017}
}

@inproceedings{kdd1,
  title={{UniST: A Prompt-empowered Universal Model for Urban Spatio-temporal Prediction}},
  author={Yuan, Yuan and Ding, Jingtao and Feng, Jie and Jin, Depeng and Li, Yong},
  booktitle={KDD},
  pages={4095--4106},
  year={2024}
}

@inproceedings{kdd2,
  title={{Multi-task Learning for Routing Problem with Cross-problem Zero-shot Generalization}},
  author={Liu, Fei and Lin, Xi and Wang, Zhenkun and Zhang, Qingfu and Xialiang, Tong and Yuan, Mingxuan},
  booktitle={KDD},
  pages={1898--1908},
  year={2024}
}

@inproceedings{HAFusion,
  title={{Urban Region Representation Learning with Attentive Fusion}},
  author={Sun, Fengze and Qi, Jianzhong and Chang, Yanchuan and Fan, Xiaoliang and Karunasekera, Shanika and Tanin, Egemen},
  booktitle={ICDE},
  pages={4409--4421},
  year={2024}
}

@article{CDAE,
  title={{Learning Urban Community Structures: A Collective Embedding Perspective with Periodic Spatial-Temporal Mobility Graphs}},
  author={Wang, Pengyang and Fu, Yanjie and Zhang, Jiawei and Li, Xiaolin and Lin, Dan},
  journal={ACM Transactions on Intelligent Systems and Technology},
  volume={9},
  number={6},
  pages={63:1--63:28},
  year={2018}
}

@inproceedings{MP-VN,
  title={{Efficient Region Embedding with Multi-View Spatial Networks: A Perspective of Locality-Constrained Spatial Autocorrelations}},
  author={Fu, Yanjie and Wang, Pengyang and Du, Jiadi and Wu, Le and Li, Xiaolin},
  booktitle={AAAI},
  pages={906--913},
  year={2019}
}

@inproceedings{CGAL,
  title={{Unifying Inter-Region Autocorrelation and Intra-Region Structures for Spatial Embedding via Collective Adversarial Learning}},
  author={Zhang, Yunchao and Fu, Yanjie and Wang, Pengyang and Li, Xiaolin and Zheng, Yu},
  booktitle={KDD},
  pages={1700--1708},
  year={2019}
}

@article{ReMVC,
  title={{Region Embedding With Intra and Inter-View Contrastive Learning}},
  author={Zhang, Liang and Long, Cheng and Cong, Gao},
  journal={IEEE Transactions on Knowledge and Data Engineering},
  year={2023},
  volume = {35},
number = {9},
  pages={9031--9036},
}

@inproceedings{Region2Vec,
  title={{Urban Region Profiling via Multi-Graph Representation Learning}},
  author={Luo, Yan and Chung, Fu-lai and Chen, Kai},
  booktitle={CIKM},
  pages={4294--4298},
  year={2022}
}

@inproceedings{DLCL,
  title={{Beyond Geo-First Law: Learning Spatial Representations via Integrated Autocorrelations and Complementarity}},
  author={Du, Jiadi and Zhang, Yunchao and Wang, Pengyang and Leopold, Jennifer and Fu, Yanjie},
  booktitle={ICDM},
  pages={160--169},
  year={2019}
}

@inproceedings{urban2vec,
  title={{Urban2Vec: Incorporating Street View Imagery and POIs for Multi-Modal Urban Neighborhood Embedding}},
  author={Wang, Zhecheng and Li, Haoyuan and Rajagopal, Ram},
  booktitle={AAAI},
  pages={1013--1020},
  year={2020}
}

@article{m3g,
  title={{Learning Neighborhood Representation from Multi-Modal Multi-Graph: Image, Text, Mobility Graph and Beyond}},
  author={Huang, Tianyuan and Wang, Zhecheng and Sheng, Hao and Ng, Andrew Y. and Rajagopal, Ram},
  journal={arXiv preprint arXiv:2105.02489},
  year={2021}
}

@article{RAW,
  title={{Regions are Who Walk Them: a Large Pre-trained Spatiotemporal Model Based on Human Mobility for Ubiquitous Urban Sensing}},
  author={Zhang, Ruixing and Han, Liangzhe and Sun, Leilei and Liu, Yunqi and Wang, Jibin and Lv, Weifeng},
  journal={arXiv preprint arXiv:2311.10471},
  year={2023}
}

@inproceedings{CGAP,
  title     = {{CGAP: Urban Region Representation Learning with Coarsened Graph Attention Pooling}},
  author    = {Xu, Zhuo and Zhou, Xiao},
  booktitle = {IJCAI},
  pages     = {7518--7526},
  year      = {2024}
}

@inproceedings{MuseCL,
  title     = {{MuseCL: Predicting Urban Socioeconomic Indicators via Multi-Semantic Contrastive Learning}},
  author    = {Yong, Xixian and Zhou, Xiao},
  booktitle = {IJCAI},
  pages     = {7536--7544},
  year      = {2024}
}

@article{MMGR,
  title={{Geographic Mapping with Unsupervised Multi-modal Representation Learning from VHR Images and POIs}},
  author={Bai, Lubin and Huang, Weiming and Zhang, Xiuyuan and Du, Shihong and Cong, Gao and Wang, Haoyu and Liu, Bo},
  journal={ISPRS Journal of Photogrammetry and Remote Sensing},
  volume={201},
  pages={193--208},
  year={2023},
  publisher={Elsevier}
}

@article{hyperregion,
  title={HyperRegion: Integrating Graph and Hypergraph Contrastive Learning for Region Embeddings},
  author={Deng, Mingyu and Chen, Chao and Zhang, Wanyi and Zhao, Jie and Yang, Wei and Guo, Suiming and Pu, Huayan and Luo, Jun},
  journal={IEEE Transactions on Mobile Computing},
  year={2024},
  volume = {24},
    number = {5},
    pages = {3667--3684},
  publisher={IEEE}
}

@article{BPURF,
  title={Boundary Prompting: Elastic Urban Region Representation via Graph-based Spatial Tokenization},
  author={Zhu, Haojia and Jin, Jiahui and Kan, Dong and Shen, Rouxi and Wang, Ruize and Sun, Xiangguo and Zhang, Jinghui},
  journal={arXiv preprint arXiv:2503.07991},
  year={2025}
}

@inproceedings{perozzi2014deepwalk,
  title={DeepWalk: Online Learning of Social Representations},
  author={Perozzi, Bryan and Al-Rfou, Rami and Skiena, Steven},
  booktitle={KDD},
  pages={701--710},
  year={2014}
}

@inproceedings{nlp,
  title={Language Models are Few-shot Learners},
  author={Brown, Tom and Mann, Benjamin and Ryder, Nick and Subbiah, Melanie and Kaplan, Jared D and Dhariwal, Prafulla and Neelakantan, Arvind and Shyam, Pranav and Sastry, Girish and Askell, Amanda and Agarwal, Sandhini and Herbert-Voss, Ariel and Krueger, Gretchen and Henighan, Tom and Child, Rewon and Ramesh, Aditya and Ziegler, Daniel and Wu, Jeffrey and Winter, Clemens and Hesse, Chris and Chen, Mark and Sigler, Eric and Litwin, Mateusz and Gray, Scott and Chess, Benjamin and Clark, Jack and Berner, Christopher and McCandlish, Sam and Radford, Alec and Sutskever, Ilya and Amodei, Dario},
  booktitle={NeurIPS},
  pages={1877--1901},
  year={2020}
}

@article{nlp2,
  title={{LLaMA: Open and Efficient Foundation Language Models}},
  author={Touvron, Hugo and Lavril, Thibaut and Izacard, Gautier and Martinet, Xavier and Lachaux, Marie-Anne and Lacroix, Timoth{\'e}e and Rozi{\`e}re, Baptiste and Goyal, Naman and Hambro, Eric and Azhar, Faisal and Rodriguez, Aurelien and Joulin, Armand and Grave, Edouard and Lample, Guillaume},
  journal={arXiv preprint arXiv:2302.13971},
  year={2023}
}

@inproceedings{cv,
  author      = {Alexey Dosovitskiy and
                  Lucas Beyer and
                  Alexander Kolesnikov and
                  Dirk Weissenborn and
                  Xiaohua Zhai and
                  Thomas Unterthiner and
                  Mostafa Dehghani and
                  Matthias Minderer and
                  Georg Heigold and
                  Sylvain Gelly and
                  Jakob Uszkoreit and
                  Neil Houlsby},
  title        = {An Image is Worth 16x16 Words: Transformers for Image Recognitionat Scale},
  booktitle    = {ICLR},
  year         = {2021},
}

@inproceedings{cv2,
  title={Masked Autoencoders are Scalable Vision Learners},
  author={He, Kaiming and Chen, Xinlei and Xie, Saining and Li, Yanghao and Doll{\'a}r, Piotr and Girshick, Ross},
  booktitle={CVPR},
  pages={16000--16009},
  year={2022}
}

@inproceedings{hgnn,
  title={{Heterogeneous Graph Neural Network}},
  author={Zhang, Chuxu and Song, Dongjin and Huang, Chao and Swami, Ananthram and Chawla, Nitesh V},
  booktitle={KDD},
  pages={793--803},
  year={2019}
}

@article{MoRA,
  title={MobCLIP: Learning General-purpose Geospatial Representation at Scale},
  author={Wen, Ya and Cai, Jixuan and Ma, Qiyao and Li, Linyan and Chen, Xinhua and Webster, Chris and Zhou, Yulun},
  journal={arXiv preprint arXiv:2506.01297},
  year={2025}
}

@article{MGRL4RE,
author = {Chen, Meng and Li, Zechen and Jia, Hongwei and Shao, Xin and Zhao, Jun and Gao, Qiang and Yang, Min and Yin, Yilong},
title = {MGRL4RE: A Multi-Graph Representation Learning Approach for Urban Region Embedding},
volume = {16},
number = {2},
year = {2025},
articleno = {49},
numpages = {23},
journal = {ACM Transactions on Intelligent Systems and Technology}
}

@inproceedings{MVGCL,
  title={{Multi-view Graph Contrastive Learning for Urban Region Representation}},
  author={Zhang, Yu and Xu, Yonghui and Cui, Lizhen and Yan, Zhongmin},
  booktitle={IJCNN},
  pages={1--8},
  year={2023}
}

@inproceedings{diff_st_1,
  title={Diffusion Transformers as Open-World Spatiotemporal Foundation Models},
  author={Yuan, Yuan and Han, Chonghua and Ding, Jingtao and Zhang, Guozhen and Jin, Depeng and Li, Yong},
  booktitle={NeurIPS},
  year={2025}
}

@inproceedings{diff_st_2,
  title={{Stochastic Diffusion: A Diffusion based Model for Stochastic Time Series Forecasting}},
  author={Liu, Yuansan and Wijewickrema, Sudanthi and Hu, Dongting and Bester, Christofer and O'Leary, Stephen and Bailey, James},
  booktitle={KDD},
  pages={1939--1950},
  year={2025}
}

@inproceedings{diff_st_3,
  title={{ControlTraj: Controllable Trajectory Generation with Topology-constrained Diffusion Model}},
  author={Zhu, Yuanshao and Yu, James Jianqiao and Zhao, Xiangyu and Liu, Qidong and Ye, Yongchao and Chen, Wei and Zhang, Zijian and Wei, Xuetao and Liang, Yuxuan},
  booktitle={KDD},
  pages={4676--4687},
  year={2024}
}

@inproceedings{diff_st_4,
  title={{Predict, Refine, Synthesize: Self-guiding Diffusion Models for Probabilistic Time Series Forecasting}},
  author={Kollovieh, Marcel and Ansari, Abdul Fatir and Bohlke-Schneider, Michael and Zschiegner, Jasper and Wang, Hao and Wang, Yuyang Bernie},
  booktitle={NeurIPS},
  pages={28341--28364},
  year={2023}
}

@inproceedings{diff_image_1,
author = {Dhariwal, Prafulla and Nichol, Alex},
title = {{Diffusion Models Beat GANs on Image Synthesis}},
year = {2021},
booktitle = {NeurIPS},
pages = {8780--8794}
}

@inproceedings{diff_image_2,
  title={{Scalable Diffusion Models with Transformers}},
  author={Peebles, William and Xie, Saining},
  booktitle={ICCV},
  pages={4195--4205},
  year={2023}
}

@inproceedings{
  diff_image_3,
  title={{Score-Based Generative Modeling through Stochastic Differential Equations}},
  author={Yang Song and Jascha Sohl-Dickstein and Diederik P. Kingma and Abhishek Kumar and Stefano Ermon and Ben Poole},
  booktitle={ICLR},
  year={2021},
}

@inproceedings{gat,
  title={{Graph Attention Networks}},
  author={Veli{\v{c}}kovi{\'c}, Petar and Cucurull, Guillem and Casanova, Arantxa and Romero, Adriana and Lio, Pietro and Bengio, Yoshua},
  booktitle={ICLR},
  year={2018}
}

@inproceedings{su2025generalising,
  title={Generalising Traffic Forecasting to Regions without Traffic Observations},
  author={Su, Xinyu and Sarvi, Majid and Liu, Feng and Tanin, Egemen and Qi, Jianzhong},
  booktitle={AAAI},
  year={2026}
}

@inproceedings{ijcai25su, 
author = {Su, Xinyu and Liu, Feng and Chang, Yanchuan and Tanin, Egemen and Sarvi, Majid and Qi, Jianzhong}, title = {DualCast: Disentangling Aperiodic Events from Traffic Series with a Dual-Branch Model}, 
booktitle = {IJCAI}, 
year = {2025} }

@online{sgOpendata,
  author = {{Singapore Dataset}},
  title = {\url{https://data.gov.sg}},
year = {2019}
}

@online{lxRegion,
  author = {{Lisbon Region Boundary}},
  title = {\url{https://gadm.org/index.html}},
year = {2024}
}

@article{chang,
author = {Chang, Yanchuan and Tanin, Egemen and Cong, Gao and Jensen, Christian S. and Qi, Jianzhong},
title = {Trajectory Similarity Measurement: An Efficiency Perspective},
year = {2024},
journal = {VLDB},
}

\appendix

\section{Related Work}
\label{sec:appendic_of_related_work}

Existing studies typically adopt a two-stage learning framework, consisting of a region embedding learning stage followed by a downstream task learning stage.

\paragraph{Region Embedding Learning} Generic region embeddings are learned in a self-supervised manner from a variety of regional features, typically following  three training paradigms. 

The first paradigm focuses on  reconstructing predefined correlations between regions, such as POI distribution similarity or taxi flow statistics ~\cite{DLCL, MP-VN, CDAE, CGAL, MVURE, CGAP, MGRL4RE, MGFN, RAW, Region2Vec}.
For example, MVURE~\cite{MVURE} constructs four graphs (where each region is a node) based on human mobility (source and destination) and POI category and check-in similarities. It applies Graph Attention Networks (GATs)~\cite{gat} to learn embeddings, which are then aggregated via weighted summation.
MGFN~\cite{MGFN} builds multiple hourly mobility graphs and clusters them into mobility pattern graphs using time-weighted distances, modeling intra- and inter-pattern correlations through self-attention to generate region embeddings.
CGAP~\cite{CGAP} learns region embeddings by hierarchically pooling the graph of region nodes via attention to extract global urban features. Such features are integrated back into region representations through a global attention mechanism.

The second paradigm adopts contrastive learning by constructing positive and negative samples based on region similarity derived from input features (e.g., geographic proximity or POI distributions), or model-specific design~\cite{urban2vec, ReMVC, MVGCL, cityFM, m3g, MMGR, MoRA, MuseCL, ReCP, RegionDCL, urbanclip, hyperregion, GeoHG}.
For example, 
ReMVC~\cite{ReMVC} learns region embeddings via self-supervised contrastive learning both within and across multiple feature views, and concatenates view-specific embeddings.
RegionDCL~\cite{RegionDCL} partitions building footprints into non-overlapping groups and applies contrastive learning at both the building-group and region levels. The resulting building-group embeddings are aggregated to form region representations.
UrbanCLIP~\cite{urbanclip} leverages a vision–language model to generate textual descriptions of region-level satellite imagery. It  learns region embeddings by applying contrastive learning to the resulting image–text pairs.
CityFM~\cite{cityFM} learns embeddings of diverse OpenStreetMap entities by optimizing multiple contrastive objectives spanning text, vision, and road contexts, which are then aggregated to produce region representations.
ReCP~\cite{ReCP} replaces fusion modules with a contrastive framework that maximizes inter-view mutual information while minimizing conditional entropy.
GeoHG~\cite{GeoHG} constructs a heterogeneous graph from imagery and POI data and applies contrastive learning with a heterogeneous Graph Neural Network (GNN) to derive region embeddings.
HyperRegion~\cite{hyperregion} integrates graph and hypergraph contrastive learning over multi-modal urban data to capture both pairwise and higher-order regional interactions.

The third paradigm combines reconstruction and contrastive objectives to further enhance representation learning~\cite{HAFusion, HREP, FlexiReg, GURPP, BPURF}. 
For example, 
HAFusion~\cite{HAFusion} learns region embeddings from human mobility, POI, and land-use data. It then applies attention-based fusion at both the region and feature view levels to integrate multi-view information and capture higher-order regional correlations.
HREP~\cite{HREP} learns region embeddings from human mobility, POI, and geographic neighborhood features, and adapts them to different downstream tasks by concatenating randomly initialized learnable prompt embeddings.
FlexiReg~\cite{FlexiReg} learns fine-grained grid-cell embeddings and aggregates them to obtain region embeddings. The learned region embeddings are further adapted to different downstream tasks by incorporating task-relevant features guided by the target tasks.
GURPP~\cite{GURPP} constructs a heterogeneous graph from diverse region features and applies a heterogeneous graph transformer to learn region embeddings, which are then adapted to downstream tasks via task-specific prompting.

Most existing methods adopt a city-centric design, modeling the entire city and optimizing city-specific objectives. The learned embeddings tend to overfit to city-dependent characteristics, limiting their generalization across cities (when used without retraining for each city). In contrast, our \RWM\ module adopts a region-centric design that focuses on features local to target regions and the structural context of their neighboring regions, rather than the entire city, enabling cross-city generalization.

\paragraph{Downstream Task learning} Most existing methods apply learned region embeddings to downstream tasks by training separate task-specific predictors, treating each task (and city) independently. Several recent works~\cite{HREP, FlexiReg, GURPP} improve task performance by injecting task-relevant information to adapt region embeddings for individual tasks.
This paradigm still relies on independent predictors for different tasks, limiting cross-task generalization. In contrast, our \diffM\ module jointly models multiple downstream tasks within a unified conditional diffusion framework, enabling shared learning across tasks while explicitly modeling task-dependent output distributions.

Diffusion models, originally successful in image generation~\cite{diff_image_1, diff_image_2, diff_image_3}, have recently demonstrated strong performance in urban spatiotemporal prediction~\cite{diff_st_1, diff_st_2, diff_st_3, diff_st_4}. For example, UrbanDiT~\cite{diff_st_1} proposes a foundation model for urban spatio-temporal learning that unifies grid- and graph-based urban data within a diffusion transformer framework to model complex spatio-temporal dynamics. 
StochDiff~\cite{diff_st_2} introduces a diffusion-based time series forecasting model that integrates the diffusion process directly into sequential modeling, enabling step-wise learning of temporal dynamics. 
These successes suggest that modeling the full conditional distribution of targets is more effective than point estimation, motivating the design of \diffM.

\section{Notation}
\label{subsec:appendix_of_symbols}
We summarize the frequently used symbols in Table~\ref{tab:symbols}.

\begin{table}[ht]
\centering
\caption{Frequently Used Symbols.}
\label{tab:symbols}
\resizebox{\columnwidth }{!}{
    \begin{tabular}{cl}
    \toprule
    \textbf{Symbol} & \textbf{Description} \\ 
    \midrule
     $S$ & A spatial area of interest  \\ \hline
    $R$ & A set of regions (non-overlapping space partitions)  \\ \hline 
    $N$ & The number of regions \\ \hline
    $C$ & A set of spatial grid cells (basic space partition units) \\ \hline 
    $n$ & The number of grid cells \\ \hline
    $\boldsymbol{p}_i$ & POI feature of cell $c_i$ \\ \hline 
    $\boldsymbol{gn}_i$ & Geographic neighbor feature of cell  $c_i$ \\ \hline
    $\mathbf{X}$ & The embeddings of grid cells \\ \hline 
    $\mathbf{H}$ & The embeddings of regions \\  \hline
    $\mathbf{Z}^e$ & The encoder output embeddings of  \RWM\ \\  \hline
    $\mathbf{Z}^d$ & The decoder output embeddings of \RWM\ \\  \hline
    $\boldsymbol{y}$ & Ground-truth values for downstream tasks \\  \hline
    $\widetilde{\boldsymbol{y}}$ &  Generated prior knowledge for downstream tasks \\  \hline
    $\hat{\boldsymbol{\epsilon}}, \boldsymbol{\epsilon}$ & Predicted noise and ground-truth noise \\  \hline
    $t$  & A timestep in diffusion process\\ \hline
    $u$  & A task identifier \\
    \bottomrule
    \end{tabular}
}
\end{table}

\section{Details of Grid Cell Construction}
\label{subsec:appendix_of_cell_construction}

\begin{figure}[htbp]
  \centering
  \includegraphics[width=\columnwidth]{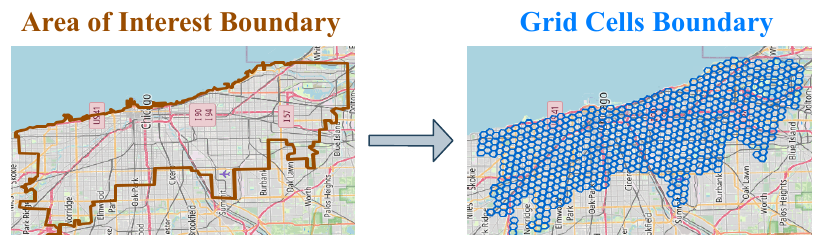}
  \caption{Grid cell construction.}
  \label{fig:grid_cell_construction}
\end{figure}

We partition the input spatial area of interest into a set $C$ of non-overlapping grid cells, where $c_i \in C$ denotes the $i$-th cell. The  grid cells provide a finer-grained spatial partition than regions, enabling flexible region formation. We employ a hexagonal grid with an edge length of  150 meters to partition the area, as illustrated in Fig.~\ref{fig:grid_cell_construction}.

\section{Details of Adaptive Region Embedding Learning}
\label{subse:appendix_of_adaRegionGen}

Following FlexiReg~\cite{FlexiReg}, we generate the region embeddings $\mathbf{H} = \{\boldsymbol{h}_{i}\}^{N}_{i=1}$ by aggregating the grid cell embeddings $\mathbf{X} = \{ \boldsymbol{x}_{i}\}_{i=1}^{n}$, weighted by the size of the overlapping area of a cell and the target regions.

Given a region $r_j$, we find the set of grid cells that overlap with (or are fully contained by) $r_j$, denoted as $\mathcal{C}_{r_j} = \{ c_1, \cdots, c_i, \cdots \}$. 
We compute the overlapping ratio between  $r_j$ and each $c_i \in \mathcal{C}_{r_j}$ based on the size of their overlapping area, denoted as $\omega_{{r_j} \cap {c_i}}$: 
\begin{equation}\label{eq:aggregation_1}
    \omega_{{r_j} \cap {c_i}} = \frac{\mathrm{Area}(r_j \cap c_i)}{\mathrm{Area}(c_i)},
\end{equation}
Here, $\cap$ denotes spatial intersection, and $\mathrm{Area}(\cdot)$ computes the size of an area.

This ratio reflects the contribution of $c_i$ to the region embedding of $r_j$. The region embedding $\boldsymbol{h}_j$ is then computed as the sum of the cell embeddings $\boldsymbol{x}_i$ weighted by their respective overlapping ratios: 
{\small
\begin{equation}\label{eq:aggregation_2}
\boldsymbol{h}_j = \sum_{c_i \in \mathcal{C}_{r_j}} \omega_{{r_j} \cap {c_i}} \cdot \boldsymbol{x}_i,
\end{equation}
}

\section{Denoising Diffusion Probabilistic Model}
\label{subsec:appendix_of_ddpm}

The Denoising Diffusion Probabilistic Model (DDPM)~\cite{ho2020denoising} comprises a forward and a reverse process. In the \textbf{forward process}, Gaussian noise is gradually added to a data sample $\boldsymbol{y}_0 \sim q(\boldsymbol{y}_0)$ over $t$ steps to produce $\bold y_t$:
{\small
\begin{align}
q(\boldsymbol{y}_t \mid \boldsymbol{y}_{t-1})
&= \mathcal{N}\!\left(
\boldsymbol{y}_t;
\sqrt{1-\beta_t}\,\boldsymbol{y}_{t-1},
\beta_t \mathbf{I}
\right), \\
q(\boldsymbol{y}_t \mid \boldsymbol{y}_0)
&= \mathcal{N}\!\left(
\boldsymbol{y}_t;
\sqrt{\bar{\alpha}_t}\,\boldsymbol{y}_0,
(1-\bar{\alpha}_t)\mathbf{I}
\right), \label{eq:forward_process}
\end{align}
}where $\beta_t$ is a noise level controller, ${\alpha}_t = 1- \beta_t$, and $\bar{\alpha}_t = \prod_{i=1}^{t} \alpha_i$. In the \textbf{reverse process}, a denoising neural network $p_\theta(\boldsymbol{y}_{t-1} \mid \boldsymbol{y}_t)$ is trained to approximate the true posterior $q(\boldsymbol{y}_{t-1} \mid \boldsymbol{y}_t, \boldsymbol{y}_0)$, which can be derived and expressed as a Gaussian distribution:
{\small
\begin{equation}
    q(\boldsymbol{y}_{t-1} \mid \boldsymbol{y}_t, \boldsymbol{y}_0) =
\mathcal{N}\!\left(
\boldsymbol{y}_{t-1};
\tilde{\boldsymbol{\mu}}_t(\boldsymbol{y}_t, \boldsymbol{y}_0),
\tilde{\beta}_t \mathbf{I}
\right).
\end{equation}
}where $\tilde{\boldsymbol{\mu}}_t(\boldsymbol{y}_t, \boldsymbol{y}_0)
=
\frac{\sqrt{\bar{\alpha}_{t-1}}\,\beta_t}{1-\bar{\alpha}_t}\,\boldsymbol{y}_0
+
\frac{\sqrt{\alpha_t}\,(1-\bar{\alpha}_{t-1})}{1-\bar{\alpha}_t}\,\boldsymbol{y}_t,$ and 
$\tilde{\beta}_t
=
\frac{1-\bar{\alpha}_{t-1}}{1-\bar{\alpha}_t}\,\beta_t.$

The model is trained by minimizing the KL divergence between the true posterior distribution $q(\boldsymbol{y}_{t-1} \mid \boldsymbol{y}_t, \boldsymbol{y}_0)$ and the model approximation $p_\theta(\boldsymbol{y}_{t-1} \mid \boldsymbol{y}_t)$. 
This training objective can be further simplified to minimizing the mean squared error between the actual noise $\boldsymbol{\epsilon}$ added to $\boldsymbol{y}_0$ and the noise predicted by the denoising network $\boldsymbol{\epsilon}_\theta$:
{\small
\begin{equation}
    \mathcal{L}_{\mathrm{simple}}(\theta)
=
\mathbb{E}_{\boldsymbol{y}_0, t, \boldsymbol{\epsilon}}
\bigl[
\|\boldsymbol{\epsilon} - \boldsymbol{\epsilon}_\theta(\boldsymbol{y}_t, t)\|^2
\bigr],
\end{equation}
}where $t$ and $\boldsymbol{\epsilon} \sim \mathcal{N}(\mathbf{0}, \boldsymbol{I})$ are randomly sampled, and $\boldsymbol{y}_t$ is constructed by perturbing $\boldsymbol{y}_0$ according to Eq.~\ref{eq:forward_process}.
Once trained, new samples are generated by iteratively denoising $\boldsymbol{y}_T \sim \mathcal{N}(\mathbf{0}, \boldsymbol{I})$ to recover the original data distribution.

\section{Derivation for Forward Process Posteriors}
\label{subsec:appendix_of_derivation_for_forward}

We derive the mean and variance of the forward process posteriors 
$q\!\left(\boldsymbol{y}_{t-1}^{u} \mid \boldsymbol{y}_t^{u}, \boldsymbol{y}_0^{u}, \widetilde{\boldsymbol{y}}^{u}\right)$ in Eq.\ref{eq:reverse}:

{\small
\begin{equation*}
q\!\left(
\boldsymbol{y}_{t-1}^{u}
\mid
\boldsymbol{y}_{t}^{u},
\boldsymbol{y}_{0}^{u},
\widetilde{\boldsymbol{y}}^{u}
\right)
\end{equation*}}
{\small
\begin{equation}
\;\propto\;
q\!\left(
\boldsymbol{y}_{t}^{u}
\mid
\boldsymbol{y}_{t-1}^{u},
\widetilde{\boldsymbol{y}}^{u}
\right)
\,
q\!\left(
\boldsymbol{y}_{t-1}^{u}
\mid
\boldsymbol{y}_{0}^{u},
\widetilde{\boldsymbol{y}}^{u}
\right) \text{(based on Eq.\ref{eq:forward_1} and Eq.\ref{eq:forward_2})}
\end{equation}}
{\small
\begin{equation}
\propto
\exp\!\left(
-\frac{1}{2}
\begin{aligned}
\Bigg[
&\frac{
\left(
\boldsymbol{y}_{t}^{u}
-
(1-\sqrt{\alpha_t})\widetilde{\boldsymbol{y}}^{u}
-
\sqrt{\alpha_t}\boldsymbol{y}_{t-1}^{u}
\right)^2
}{\beta_t}
\\
&+
\frac{
\left(
\boldsymbol{y}_{t-1}^{u}
-
\sqrt{\bar{\alpha}_{t-1}}\boldsymbol{y}_{0}^{u}
-
(1-\sqrt{\bar{\alpha}_{t-1}})\widetilde{\boldsymbol{y}}^{u}
\right)^2
}{1-\bar{\alpha}_{t-1}}
\Bigg]
\end{aligned}
\right)
\end{equation}}

{\small
\begin{equation}
\propto
\exp\!\left(
-\frac{1}{2}
\begin{aligned}
\Bigg(
&\frac{
\alpha_t \big(\boldsymbol{y}_{t-1}^{u}\big)^2
-2\sqrt{\alpha_t}
\Big(
\boldsymbol{y}_t^{u}
-
(1-\sqrt{\alpha_t})\widetilde{\boldsymbol{y}}^{u}
\Big)
\boldsymbol{y}_{t-1}^{u}
}{\beta_t}
\\
&+
\frac{
\big(\boldsymbol{y}_{t-1}^{u}\big)^2
-2\Big(
\sqrt{\bar{\alpha}_{t-1}}\boldsymbol{y}_0^{u}
+
(1-\sqrt{\bar{\alpha}_{t-1}})\widetilde{\boldsymbol{y}}^{u}
\Big)
\boldsymbol{y}_{t-1}^{u}
}{1-\bar{\alpha}_{t-1}}
\Bigg)
\end{aligned}
\right)
\end{equation}
}

{\small
\begin{equation}
    =
\exp\!\left(
-\frac{1}{2}
\begin{aligned}
\Bigg(
& D_1 \big(\boldsymbol{y}_{t-1}^{u}\big)^2
-
2 D_2 \boldsymbol{y}_{t-1}^{u}
\Bigg)
\end{aligned}
\right),
\end{equation}}
where 
{\small
\begin{align}
D_1
&=
\frac{\alpha_t(1-\bar{\alpha}_{t-1})+\beta_t}{\beta_t(1-\bar{\alpha}_{t-1})}
=
\frac{1-\bar{\alpha}_t}{\beta_t(1-\bar{\alpha}_{t-1})},
\\[6pt]
D_2
&=
\frac{\sqrt{\bar{\alpha}_{t-1}}}{1-\bar{\alpha}_{t-1}}
\boldsymbol{y}_0^{u}
+
\frac{\sqrt{\alpha_t}}{\beta_t}
\boldsymbol{y}_t^{u}
+
\left(
\frac{\sqrt{\alpha_t}(\sqrt{\alpha_t}-1)}{\beta_t}
+
\frac{1-\sqrt{\bar{\alpha}_{t-1}}}{1-\bar{\alpha}_{t-1}}
\right)
\widetilde{\boldsymbol{y}}^{u},
\end{align}
}and we have the posterior variance:
{\small
\begin{equation}
\tilde{\beta}_t
=
\frac{1}{D_1}
=
\frac{1-\bar{\alpha}_{t-1}}{1-\bar{\alpha}_t}\,\beta_t.
\end{equation}
}Meanwhile, the coefficients of the terms in the posterior mean are obtained through dividing each coefficient in $D_2$ by $D_1$: 

{\small
\begin{equation}
\gamma_0
=
\frac{\sqrt{\bar{\alpha}_{t-1}}}{1-\bar{\alpha}_{t-1}}
\cdot
\frac{1}{D_1}
=
\frac{\beta_t\sqrt{\bar{\alpha}_{t-1}}}{1-\bar{\alpha}_t},
\end{equation}
}

{\small
\begin{equation}
\gamma_1
=
\frac{\sqrt{\alpha_t}}{\beta_t}
\cdot
\frac{1}{D_1}
=
\frac{1-\bar{\alpha}_{t-1}}{1-\bar{\alpha}_t}
\sqrt{\alpha_t},
\end{equation}
}

{\small
\begin{align}
\gamma_2
&=
\left(
\frac{\sqrt{\alpha_t}(\sqrt{\alpha_t}-1)}{\beta_t}
+
\frac{1-\sqrt{\bar{\alpha}_{t-1}}}{1-\bar{\alpha}_{t-1}}
\right)
\cdot
\frac{1}{D_1}
\\[6pt]
&=
\frac{
\alpha_t
-
\bar{\alpha}_t
-
\sqrt{\alpha_t}(1-\bar{\alpha}_{t-1})
+
\beta_t
-
\beta_t\sqrt{\bar{\alpha}_{t-1}}
}{
1-\bar{\alpha}_t
}
\\[6pt]
&=
1
+
\frac{(\sqrt{\bar\alpha_t}-1)\big(\sqrt{\alpha_t}+\sqrt{\bar{\alpha}_{t-1}}\big)}
{1-\bar{\alpha}_t},
\end{align}
}which together give us the posterior mean:
\begin{equation}
\tilde{\mu}\!\left(
\boldsymbol{y}_t^{u},
\boldsymbol{y}_0^{u},
\widetilde{\boldsymbol{y}}^{u}
\right)
=
\gamma_0 \boldsymbol{y}_0^{u}
+
\gamma_1 \boldsymbol{y}_t^{u}
+
\gamma_2 \widetilde{\boldsymbol{y}}^{u}.
\end{equation}

\section{Details of Experimental Datasets}
\label{subsec:appendix_of_dataset_statistics}

We use real data from three cities: New York City (\textbf{NYC})~\cite{nycOpendata}, Chicago (\textbf{CHI})~\cite{chiOpendata}, and San Francisco (\textbf{SF})~\cite{sfOpendata}. For each city, we collect data on region division, POI information and six region indicators (crime, check-in, service call, population, carbon emission, and nightlight). 
Among these indicators, crime, check-in, and service call capture dynamic human activity, while population, carbon emission, and nightlight represent static socioeconomic attributes.

Table~\ref{tab:datasets} summarizes the  NYC, CHI, and SF datasets. Each dataset consists of region boundaries obtained from open data portals, POIs extracted from OpenStreetMap~\cite{osm} with 15 representative category labels, including \emph{educational institutions, commercial and industrial properties, accommodation, cultural and recreational venues, healthcare and medical facilities, entertainment venues, places of worship, food and drink establishments, parking facilities, transportation and transit facilities, residential properties, camping and outdoor recreation sites, sports and recreation facilities, financial services, and others.}. Crime and service call records are retrieved from open data portals, while check-in records are obtained from a Foursquare dataset~\cite{checkinData}. Each record contains location and time information, with counts aggregated at the region level. Population data  is sourced from WorldPop~\cite{WorldPop}. Carbon emission estimates are sourced from the Open-Data Inventory for Anthropogenic Carbon dioxide (ODIAC)~\cite{ODIAC}. Nightlight data are obtained from the harmonized DMSP–VIIRS dataset~\cite{nightlight}.

\begin{table}[ht]
\centering
\caption{Dataset Statistics: New York City, Chicago, and San Francisco.}
\label{tab:datasets}
\setlength{\tabcolsep}{2pt}
\resizebox{\columnwidth}{!}{
\begin{tabular}{lrrr}
\toprule
&\textbf{NYC~\cite{nycOpendata}} & \textbf{CHI~\cite{chiOpendata}} & \textbf{SF~\cite{sfOpendata}} \\ \midrule
\#regions & 180 & 77 & 175 \\
\midrule

\#grid cells & 438 & 720 & 1032 \\
\midrule

\#POIs & 24,496 & 57,891 & 28,578 \\

\#POI categories & 15 & 15 & 15 \\
\midrule

{\#crime records} & 35,335 & 18,200  & 48,489 \\
(data collection time) & unknown & 12/2022 - 12/2022 & 01/2022 - 12/2022 \\
\midrule

{\#check-ins} & 106,902 & 167,232 & 87,750 \\
(data collection time) & 04/2012 - 09/2013 & 04/2012 - 09/2013 & 04/2012 - 09/2013 \\
\midrule

{\#service calls} & 516,187 & 24,350 & 34,385 \\ 
(data collection time) & 01/2023 - 03/2023 & 12/2022 - 12/2022 & 01/2022 - 12/2022 \\ 
\midrule

Population counts & 1,540,692 & 2,508,984 & 801,251 \\
(data collection time) & 2020 & 2020 & 2020 \\ 
\midrule

Carbon emissions & 293,353 & 76,703 & 111,509 \\
(data collection time) & 11/2023 - 12/2023 & 11/2023 - 12/2023 & 11/2023 - 12/2023 \\ 
\midrule

Nightlight & 35,209 & 102,516 & 46,420 \\
(data collection time) & 2020 & 2020 & 2020 \\

\bottomrule
\end{tabular}
}
\vspace{-1mm}
\end{table}

\begin{table*}[htbp]
\centering
\caption{Overall \emph{Cross-city} Prediction Accuracy Results on \emph{Crime, Check-in, and Service Call} Prediction Tasks (`$\uparrow$' indicates that large values are preferred. The best results are in boldface, and the second-best results are underlined. Same below).}
\label{tab:cross_city_full_p1}

\setlength{\tabcolsep}{2pt}
\renewcommand{\arraystretch}{1}
\resizebox{\textwidth }{!}{
\begin{tabular}{l|ccc|ccc|ccc}

\toprule [0.4ex] \\[-3.5ex]

\multirow{2}{*}{\textbf{\makecell[l]{CHI \& SF (X) \\ $\rightarrow$ NYC (Y)}}} &
\multicolumn{3}{c|}{Crime} & \multicolumn{3}{c|}{Check-in} & \multicolumn{3}{c}{Service Call}  \\

\cmidrule(lr){2-4} \cmidrule(lr){5-7} \cmidrule(lr){8-10}
    
&MAE $\downarrow$ & RMSE $\downarrow$ & $R^{2} \uparrow$ 
& MAE $\downarrow$ & RMSE $\downarrow$ & $R^{2} \uparrow$ 
 & MAE $\downarrow$ & RMSE $\downarrow$ & $R^{2} \uparrow$ \\
\midrule

HREP~\cite{HREP}
& 91.2$\,\pm\,$3.1
& 123.6$\,\pm\,$3.4
& 0.301$\,\pm\,$0.037
& 382.2$\,\pm\,$12.5
& 591.3$\,\pm\,$16.5
& 0.478$\,\pm\,$0.031
& 1638$\,\pm\,$43
& 2493$\,\pm\,$52
& 0.173$\,\pm\,$0.029 \\

RegionDCL~\cite{RegionDCL}
& 100.7$\,\pm\,$1.6
& 131.3$\,\pm\,$2.4
& 0.211$\,\pm\,$0.028
& 398.1$\,\pm\,$5.5
& 615.3$\,\pm\,$9.2
& 0.434$\,\pm\,$0.017
& 1842$\,\pm\,$15
& 2643$\,\pm\,$20
& 0.071$\,\pm\,$0.014 \\

UrbanCLIP~\cite{urbanclip}
& 98.5$\,\pm\,$1.2
& 128.9$\,\pm\,$2.3
& 0.241$\,\pm\,$0.005
& 394.1$\,\pm\,$4.8
& 619.9$\,\pm\,$6.3
& 0.426$\,\pm\,$0.008
& 1693$\,\pm\,$21
& 2520$\,\pm\,$32
& 0.155$\,\pm\,$0.015 \\

CityFM~\cite{cityFM}
& 93.5$\,\pm\,$1.4
& 124.1$\,\pm\,$2.3
& 0.297$\,\pm\,$0.017
& 388.1$\,\pm\,$17.7
& 608.7$\,\pm\,$29.8
& 0.445$\,\pm\,$0.024
& 1778$\,\pm\,$9
& 2567$\,\pm\,$16
& 0.124$\,\pm\,$0.011 \\

GeoHG~\cite{GeoHG}
& 88.6$\,\pm\,$3.6
& 120.1$\,\pm\,$5.9
& 0.339$\,\pm\,$0.047
& 420.7$\,\pm\,$33.1
& 674.8$\,\pm\,$62.3
& 0.329$\,\pm\,$0.041
& 1587$\,\pm\,$43
& 2404$\,\pm\,$61
& 0.231$\,\pm\,$0.039 \\

GURPP~\cite{GURPP}
& 71.9$\,\pm\,$4.1
& 99.6$\,\pm\,$5.5
& 0.545$\,\pm\,$0.039
& 278.6$\,\pm\,$15.6
& 435.4$\,\pm\,$27.1
& 0.716$\,\pm\,$0.034
& 1644$\,\pm\,$34
& 2379$\,\pm\,$88
& 0.247$\,\pm\,$0.055 \\

FlexiReg~\cite{FlexiReg}
& \underline{66.5$\,\pm\,$2.7}
& \underline{85.8$\,\pm\,$3.1}
& \underline{0.663$\,\pm\,$0.024}
& \underline{272.9$\,\pm\,$4.9}
& \underline{405.2$\,\pm\,$7.3}
& \underline{0.754$\,\pm\,$0.008}
& \underline{1260$\,\pm\,$25}
& \underline{1913$\,\pm\,$61}
& \underline{0.513$\,\pm\,$0.021} \\

\midrule

\textbf{\model} 
& \textbf{57.4$\,\pm\,$1.6}
& \textbf{78.2$\,\pm\,$1.1}
& \textbf{0.724$\,\pm\,$0.008}
& \textbf{240.7$\,\pm\,$5.9}
& \textbf{383.2$\,\pm\,$10.8}
& \textbf{0.781$\,\pm\,$0.013}
& \textbf{1126$\,\pm\,$15}
& \textbf{1756$\,\pm\,$20}
& \textbf{0.589$\,\pm\,$0.009} \\

\midrule
\textbf{Improvement}
& \red{\textbf{+13.68\%}}
& \red{\textbf{+8.86\%}}
& \red{\textbf{+9.20\%}}
& \red{\textbf{+11.80\%}}
& \red{\textbf{+5.43\%}}
& \red{\textbf{+3.58\%}}
& \red{\textbf{+10.63\%}}
& \red{\textbf{+8.21\%}}
& \red{\textbf{+14.81\%}} \\

\bottomrule \toprule [0.4ex] \\[-3.5ex]

\multirow{2}{*}{\textbf{\makecell[l]{NYC \& SF (X) \\ $\rightarrow$ CHI (Y)}}} &
\multicolumn{3}{c|}{Crime} & \multicolumn{3}{c|}{Check-in} & \multicolumn{3}{c}{Service Call}  \\

\cmidrule(lr){2-4} \cmidrule(lr){5-7} \cmidrule(lr){8-10}
    
&MAE $\downarrow$ & RMSE $\downarrow$ & $R^{2} \uparrow$ 
& MAE $\downarrow$ & RMSE $\downarrow$ & $R^{2} \uparrow$ 
 & MAE $\downarrow$ & RMSE $\downarrow$ & $R^{2} \uparrow$ \\
\midrule

HREP~\cite{HREP}
& 104.9$\,\pm\,$4.2
& 136.2$\,\pm\,$5.1
& 0.402$\,\pm\,$0.037
& 1880$\,\pm\,$85
& 3595$\,\pm\,$164
& 0.544$\,\pm\,$0.045
& 189.9$\,\pm\,$8.7
& 264.1$\,\pm\,$10.1
& 0.458$\,\pm\,$0.041 \\

RegionDCL~\cite{RegionDCL}
& 120.6$\,\pm\,$1.9
& 161.5$\,\pm\,$2.4
& 0.159$\,\pm\,$0.015
& 2293$\,\pm\,$17
& 4068$\,\pm\,$23
& 0.416$\,\pm\,$0.007
& 199.6$\,\pm\,$9.8
& 286.2$\,\pm\,$7.2
& 0.363$\,\pm\,$0.012 \\

UrbanCLIP~\cite{urbanclip}
& 103.8$\,\pm\,$2.2
& 135.2$\,\pm\,$3.1
& 0.411$\,\pm\,$0.010
& 2671$\,\pm\,$19
& 5172$\,\pm\,$19
& 0.087$\,\pm\,$0.007
& 186.6$\,\pm\,$7.1
& 263.1$\,\pm\,$9.8
& 0.462$\,\pm\,$0.039 \\

CityFM~\cite{cityFM}
& 119.9$\,\pm\,$3.4
& 149.7$\,\pm\,$4.3
& 0.277$\,\pm\,$0.032
& 1953$\,\pm\,$19
& 3688$\,\pm\,$31
& 0.534$\,\pm\,$0.011
& 202.5$\,\pm\,$2.7
& 280.1$\,\pm\,$9.1
& 0.389$\,\pm\,$0.019 \\

GeoHG~\cite{GeoHG}
& 98.7$\,\pm\,$1.7
& 131.6$\,\pm\,$3.1
& 0.442$\,\pm\,$0.026
& 2307$\,\pm\,$97
& 4311$\,\pm\,$150
& 0.343$\,\pm\,$0.045
& 225.1$\,\pm\,$5.8
& 295.8$\,\pm\,$7.4
& 0.321$\,\pm\,$0.035 \\

GURPP~\cite{GURPP}
& 97.4$\,\pm\,$1.3
& 127.4$\,\pm\,$2.8
& 0.477$\,\pm\,$0.024
& 2173$\,\pm\,$134
& 4484$\,\pm\,$204
& 0.289$\,\pm\,$0.063
& 200.9$\,\pm\,$5.7
& 278.9$\,\pm\,$6.5
& 0.396$\,\pm\,$0.028 \\

FlexiReg~\cite{FlexiReg}
& \underline{67.5$\,\pm\,$2.9}
& \underline{95.1$\,\pm\,$4.5}
& \underline{0.708$\,\pm\,$0.027}
& \underline{1263$\,\pm\,$62}
& \underline{2767$\,\pm\,$90}
& \underline{0.729$\,\pm\,$0.018}
& \underline{148.7$\,\pm\,$7.7}
& \underline{206.3$\,\pm\,$8.9}
& \underline{0.669$\,\pm\,$0.019} \\

\midrule

\textbf{\model} 
& \textbf{59.3$\,\pm\,$2.9}
& \textbf{84.6$\,\pm\,$4.3}
& \textbf{0.769$\,\pm\,$0.018}
& \textbf{1143.6$\,\pm\,$84}
& \textbf{2360$\,\pm\,$191}
& \textbf{0.802$\,\pm\,$0.032}
& \textbf{130.6$\,\pm\,$5.7}
& \textbf{192.2$\,\pm\,$7.1}
& \textbf{0.713$\,\pm\,$0.021} \\

\midrule
\textbf{Improvement}
& \red{\textbf{+12.15\%}}
& \red{\textbf{+11.04\%}}
& \red{\textbf{+8.62\%}}
& \red{\textbf{+9.45\%}}
& \red{\textbf{+14.71\%}}
& \red{\textbf{+10.01\%}}
& \red{\textbf{+12.17\%}}
& \red{\textbf{+6.83\%}}
& \red{\textbf{+6.58\%}} \\

\bottomrule \toprule [0.4ex] \\[-3.5ex]

\multirow{2}{*}{\textbf{\makecell[l]{NYC \& CHI (X) \\ $\rightarrow$ SF (Y)}}} &
\multicolumn{3}{c|}{Crime} & \multicolumn{3}{c|}{Check-in} & \multicolumn{3}{c}{Service Call}  \\

\cmidrule(lr){2-4} \cmidrule(lr){5-7} \cmidrule(lr){8-10}
    
&MAE $\downarrow$ & RMSE $\downarrow$ & $R^{2} \uparrow$ 
& MAE $\downarrow$ & RMSE $\downarrow$ & $R^{2} \uparrow$ 
 & MAE $\downarrow$ & RMSE $\downarrow$ & $R^{2} \uparrow$ \\
\midrule

HREP~\cite{HREP}
& 144.9$\,\pm\,$6.4
& 225.7$\,\pm\,$7.6
& 0.491$\,\pm\,$0.036
& 427.2$\,\pm\,$19.9
& 751.3$\,\pm\,$32.1
& 0.429$\,\pm\,$0.041
& 115.4$\,\pm\,$3.8
& 190.3$\,\pm\,$5.9
& 0.303$\,\pm\,$0.042 \\

RegionDCL~\cite{RegionDCL}
& 154.7$\,\pm\,$1.5
& 233.6$\,\pm\,$2.5
& 0.455$\,\pm\,$0.012
& 524.3$\,\pm\,$4.6
& 794.8$\,\pm\,$8.6
& 0.361$\,\pm\,$0.014
& 117.5$\,\pm\,$0.5
& 193.6$\,\pm\,$1.1
& 0.280$\,\pm\,$0.007 \\

UrbanCLIP~\cite{urbanclip}
& 178.6$\,\pm\,$5.8
& 277.4$\,\pm\,$15.7
& 0.252$\,\pm\,$0.022
& 488.6$\,\pm\,$13.1
& 784.7$\,\pm\,$21.7
& 0.379$\,\pm\,$0.031
& 127.9$\,\pm\,$0.8
& 207.9$\,\pm\,$1.5
& 0.171$\,\pm\,$0.012 \\

CityFM~\cite{cityFM}
& 166.5$\,\pm\,$2.2
& 251.7$\,\pm\,$3.6
& 0.375$\,\pm\,$0.018
& 549.9$\,\pm\,$8.2
& 827.2$\,\pm\,$14.1
& 0.318$\,\pm\,$0.022
& 114.8$\,\pm\,$3.6
& 180.9$\,\pm\,$5.8
& 0.376$\,\pm\,$0.019 \\

GeoHG~\cite{GeoHG}
& 183.8$\,\pm\,$7.5
& 281.6$\,\pm\,$12.9
& 0.234$\,\pm\,$0.041
& 589.8$\,\pm\,$12.3
& 896.1$\,\pm\,$33.1
& 0.223$\,\pm\,$0.058
& 126.4$\,\pm\,$3.1
& 204.2$\,\pm\,$7.8
& 0.199$\,\pm\,$0.046 \\

GURPP~\cite{GURPP}
& 143.5$\,\pm\,$6.1
& 220.6$\,\pm\,$6.9
& 0.513$\,\pm\,$0.031
& 321.3$\,\pm\,$13.3
& \underline{518.3$\,\pm\,$20.3}
& \underline{0.728$\,\pm\,$0.021}
& 123.2$\,\pm\,$5.6
& 195.1$\,\pm\,$6.1
& 0.268$\,\pm\,$0.046 \\

FlexiReg~\cite{FlexiReg}
& \underline{133.5$\,\pm\,$3.1}
& \underline{200.3$\,\pm\,$4.2}
& \underline{0.599$\,\pm\,$0.017}
& \underline{320.7$\,\pm\,$9.6}
& 523.5$\,\pm\,$8.5
& 0.723$\,\pm\,$0.009
& \underline{96.5$\,\pm\,$1.9}
& \underline{165.2$\,\pm\,$2.4}
& \underline{0.476$\,\pm\,$0.015} \\

\midrule

\textbf{\model} 
& \textbf{93.9$\,\pm\,$3.3}
& \textbf{136.6$\,\pm\,$4.1}
& \textbf{0.814$\,\pm\,$0.008}
& \textbf{275.2$\,\pm\,$11.1}
& \textbf{462.9$\,\pm\,$13.9}
& \textbf{0.783$\,\pm\,$0.012}
& \textbf{84.5$\,\pm\,$2.1}
& \textbf{150.4$\,\pm\,$3.7}
& \textbf{0.565$\,\pm\,$0.021} \\

\midrule
\textbf{Improvement}
& \red{\textbf{+29.67\%}}
& \red{\textbf{+31.80\%}}
& \red{\textbf{+35.89\%}}
& \red{\textbf{+14.19\%}}
&\red{\textbf{+11.97\%}}
& \red{\textbf{+7.56\%}}
& \red{\textbf{+12.44\%}}
& \red{\textbf{+8.96\%}}
& \red{\textbf{+18.70\%}} \\

\bottomrule
\end{tabular}
}
\end{table*}

\begin{table*}[htbp]
\centering
\caption{Overall \emph{Cross-city} Prediction Accuracy Results on \emph{Population, Carbon, and Nightlight} Prediction Tasks.}
\label{tab:cross_city_full_p2}

\setlength{\tabcolsep}{2pt}
\renewcommand{\arraystretch}{1}
\resizebox{\textwidth }{!}{
\begin{tabular}{l|ccc|ccc|ccc}

\toprule [0.4ex] \\[-3.5ex]

\multirow{2}{*}{\textbf{\makecell[l]{CHI \& SF (X) \\ $\rightarrow$ NYC (Y)}}} &
\multicolumn{3}{c|}{Population} & \multicolumn{3}{c|}{Carbon} & \multicolumn{3}{c}{Nightlight}  \\

\cmidrule(lr){2-4} \cmidrule(lr){5-7} \cmidrule(lr){8-10}
    
&MAE $\downarrow$ & RMSE $\downarrow$ & $R^{2} \uparrow$ 
& MAE $\downarrow$ & RMSE $\downarrow$ & $R^{2} \uparrow$ 
 & MAE $\downarrow$ & RMSE $\downarrow$ & $R^{2} \uparrow$ \\
\midrule

HREP~\cite{HREP}
& 3595$\,\pm\,$91
& 4542$\,\pm\,$97
& 0.262$\,\pm\,$0.034
& 668$\,\pm\,$17
& 1223$\,\pm\,$21
& 0.136$\,\pm\,$0.026
& 56.6$\,\pm\,$0.8
& 76.6$\,\pm\,$1.1
& 0.028$\,\pm\,$0.027 \\

RegionDCL~\cite{RegionDCL}
& 3801$\,\pm\,$46
& 4774$\,\pm\,$57
& 0.185$\,\pm\,$0.012
& 826$\,\pm\,$9
& 1298$\,\pm\,$11
& 0.025$\,\pm\,$0.016
& 56.5$\,\pm\,$0.9
& 77.2$\,\pm\,$1.1
& 0.011$\,\pm\,$0.029 \\

UrbanCLIP~\cite{urbanclip}
& 3596$\,\pm\,$69
& 4657$\,\pm\,$81
& 0.224$\,\pm\,$0.014
& 783$\,\pm\,$4
& 1282$\,\pm\,$12
& 0.051$\,\pm\,$0.007
& 50.8$\,\pm\,$0.6
& 63.8$\,\pm\,$1.6
& 0.324$\,\pm\,$0.022 \\

CityFM~\cite{cityFM}
& 3601$\,\pm\,$19
& 4672$\,\pm\,$27
& 0.219$\,\pm\,$0.008
& 761$\,\pm\,$7
& 1258$\,\pm\,$11
& 0.085$\,\pm\,$0.013
& 56.9$\,\pm\,$0.4
& 76.7$\,\pm\,$0.6
& 0.026$\,\pm\,$0.014 \\

GeoHG~\cite{GeoHG}
& 3583$\,\pm\,$149
& 4711$\,\pm\,$169
& 0.206$\,\pm\,$0.041
& 645$\,\pm\,$33
& 1219$\,\pm\,$69
& 0.139$\,\pm\,$0.046
& 55.3$\,\pm\,$0.9
& 72.2$\,\pm\,$1.1
& 0.137$\,\pm\,$0.023 \\

GURPP~\cite{GURPP}
& 2978$\,\pm\,$81
& 3823$\,\pm\,$98
& 0.477$\,\pm\,$0.031
& 854$\,\pm\,$39
& 1309$\,\pm\,$46
& 0.009$\,\pm\,$0.032
& 61.2$\,\pm\,$1.2
& 79.3$\,\pm\,$2.6
& -0.042$\,\pm\,$0.063 \\

FlexiReg~\cite{FlexiReg}
& \underline{2720$\,\pm\,$39}
& \underline{3470$\,\pm\,$40}
& \underline{0.569$\,\pm\,$0.011}
& \underline{585$\,\pm\,$24}
& \underline{1081$\,\pm\,$31}
& \underline{0.324$\,\pm\,$0.019}
& \underline{47.1$\,\pm\,$0.7}
& \underline{58.0$\,\pm\,$1.1}
& \underline{0.442$\,\pm\,$0.011} \\

\midrule

\textbf{\model}
& \textbf{2489$\,\pm\,$31}
& \textbf{3235$\,\pm\,$36}
& \textbf{0.626$\,\pm\,$0.006}
& \textbf{562$\,\pm\,$16}
& \textbf{1028$\,\pm\,$14}
& \textbf{0.389$\,\pm\,$0.006}
& \textbf{44.1$\,\pm\,$1.1}
& \textbf{55.3$\,\pm\,$0.7}
& \textbf{0.492$\,\pm\,$0.013} \\

\midrule
\textbf{Improvement}
& \red{\textbf{+8.49\%}}
& \red{\textbf{+6.77\%}}
& \red{\textbf{+10.02\%}}
& \red{\textbf{+3.93\%}}
& \red{\textbf{+4.90\%}}
& \red{\textbf{+20.06\%}}
& \red{\textbf{+6.37\%}}
& \red{\textbf{+4.66\%}}
& \red{\textbf{+11.31\%}} \\

\bottomrule \toprule [0.4ex] \\[-3.5ex]

\multirow{2}{*}{\textbf{\makecell[l]{NYC \& SF (X) \\ $\rightarrow$ CHI (Y)}}} &
\multicolumn{3}{c|}{Population} & \multicolumn{3}{c|}{Carbon} & \multicolumn{3}{c}{Nightlight}  \\

\cmidrule(lr){2-4} \cmidrule(lr){5-7} \cmidrule(lr){8-10}
    
&MAE $\downarrow$ & RMSE $\downarrow$ & $R^{2} \uparrow$ 
& MAE $\downarrow$ & RMSE $\downarrow$ & $R^{2} \uparrow$ 
 & MAE $\downarrow$ & RMSE $\downarrow$ & $R^{2} \uparrow$ \\
\midrule

HREP~\cite{HREP}
& 13327$\,\pm\,$575
& 17002$\,\pm\,$630
& 0.327$\,\pm\,$0.044
& 402.3$\,\pm\,$10.9
& 601.1$\,\pm\,$12.3
& 0.065$\,\pm\,$0.033
& 396.2$\,\pm\,$15.9
& 656.1$\,\pm\,$20.7
& 0.159$\,\pm\,$0.056 \\

RegionDCL~\cite{RegionDCL}
& 13757$\,\pm\,$123
& 17838$\,\pm\,$178
& 0.259$\,\pm\,$0.009
& 428.1$\,\pm\,$8.8
& 622.1$\,\pm\,$7.4
& -0.002$\,\pm\,$0.024
& 478.4$\,\pm\,$8.3
& 668.6$\,\pm\,$16.6
& 0.127$\,\pm\,$0.017 \\

UrbanCLIP~\cite{urbanclip}
& 14178$\,\pm\,$60
& 18418$\,\pm\,$67
& 0.211$\,\pm\,$0.006
& 391.4$\,\pm\,$10.8
& 591.4$\,\pm\,$21.6
& 0.094$\,\pm\,$0.057
& 245.8$\,\pm\,$8.1
& 377.4$\,\pm\,$11.6
& 0.721$\,\pm\,$0.011 \\

CityFM~\cite{cityFM}
& 13932$\,\pm\,$175
& 17697$\,\pm\,$236
& 0.264$\,\pm\,$0.018
& 407.1$\,\pm\,$1.6
& 604.1$\,\pm\,$3.2
& 0.055$\,\pm\,$0.008
& 425.4$\,\pm\,$5.5
& 688.7$\,\pm\,$13.4
& 0.074$\,\pm\,$0.011 \\

GeoHG~\cite{GeoHG}
& 13912$\,\pm\,$221
& 17397$\,\pm\,$373
& 0.296$\,\pm\,$0.031
& 382.8$\,\pm\,$5.9
& 560.7$\,\pm\,$9.3
& 0.182$\,\pm\,$0.027
& 414.3$\,\pm\,$18.7
& 681.5$\,\pm\,$29.1
& 0.093$\,\pm\,$0.027 \\

GURPP~\cite{GURPP}
& 11751$\,\pm\,$323
& 14895$\,\pm\,$365
& 0.484$\,\pm\,$0.019
& 389.4$\,\pm\,$16.9
& 586.1$\,\pm\,$14.7
& 0.112$\,\pm\,$0.046
& 402.8$\,\pm\,$13.1
& 675.4$\,\pm\,$17.4
& 0.110$\,\pm\,$0.019 \\

FlexiReg~\cite{FlexiReg}
& \underline{7767$\,\pm\,$508}
& \underline{10697$\,\pm\,$428}
& \underline{0.733$\,\pm\,$0.022}
& \underline{285.8$\,\pm\,$28.9}
& \underline{453.7$\,\pm\,$18.3}
& \underline{0.467$\,\pm\,$0.034}
& \underline{173.2$\,\pm\,$6.4}
& \underline{268.6$\,\pm\,$10.5}
& \underline{0.859$\,\pm\,$0.011} \\

\midrule

\textbf{\model}
& \textbf{7037$\,\pm\,$395}
& \textbf{9026$\,\pm\,$367}
& \textbf{0.810$\,\pm\,$0.016}
& \textbf{258.6$\,\pm\,$13.1}
& \textbf{425.3$\,\pm\,$17.1}
& \textbf{0.531$\,\pm\,$0.027}
& \textbf{165.8$\,\pm\,$11.5}
& \textbf{234.9$\,\pm\,$24.5}
& \textbf{0.891$\,\pm\,$0.023} \\

\midrule
\textbf{Improvement}
& \red{\textbf{+9.40\%}}
& \red{\textbf{+15.62\%}}
& \red{\textbf{+10.50\%}}
& \red{\textbf{+9.51\%}}
& \red{\textbf{+6.26\%}}
& \red{\textbf{+13.70\%}}
& \red{\textbf{+4.27\%}}
& \red{\textbf{+12.55\%}}
& \red{\textbf{+3.73\%}} \\

\bottomrule \toprule [0.4ex] \\[-3.5ex]

\multirow{2}{*}{\textbf{\makecell[l]{NYC \& CHI (X) \\ $\rightarrow$ SF (Y)}}} &
\multicolumn{3}{c|}{Population} & \multicolumn{3}{c|}{Carbon} & \multicolumn{3}{c}{Nightlight}  \\

\cmidrule(lr){2-4} \cmidrule(lr){5-7} \cmidrule(lr){8-10}
    
&MAE $\downarrow$ & RMSE $\downarrow$ & $R^{2} \uparrow$ 
& MAE $\downarrow$ & RMSE $\downarrow$ & $R^{2} \uparrow$ 
 & MAE $\downarrow$ & RMSE $\downarrow$ & $R^{2} \uparrow$ \\
\midrule

HREP~\cite{HREP}
& 1398$\,\pm\,$33
& 1806$\,\pm\,$42
& 0.183$\,\pm\,$0.031
& 271.4$\,\pm\,$5.8
& 399.1$\,\pm\,$7.4
& 0.168$\,\pm\,$0.032
& 77.3$\,\pm\,$2.2
& 117.2$\,\pm\,$2.9
& 0.270$\,\pm\,$0.038 \\

RegionDCL~\cite{RegionDCL}
& 1567$\,\pm\,$18
& 2001$\,\pm\,$31
& -0.003$\,\pm\,$0.011
& 282.2$\,\pm\,$2.5
& 424.1$\,\pm\,$3.7
& 0.061$\,\pm\,$0.017
& 81.1$\,\pm\,$0.6
& 126.1$\,\pm\,$1.1
& 0.156$\,\pm\,$0.011 \\

UrbanCLIP~\cite{urbanclip}
& 1710$\,\pm\,$13
& 2376$\,\pm\,$27
& -0.414$\,\pm\,$0.021
& 261.6$\,\pm\,$1.7
& 382.2$\,\pm\,$3.9
& 0.237$\,\pm\,$0.012
& 63.6$\,\pm\,$0.6
& 81.8$\,\pm\,$2.9
& 0.644$\,\pm\,$0.027 \\

CityFM~\cite{cityFM}
& 1583$\,\pm\,$19
& 1990$\,\pm\,$29
& 0.019$\,\pm\,$0.021
& 279.8$\,\pm\,$2.1
& 417.3$\,\pm\,$4.8
& 0.078$\,\pm\,$0.013
& 86.8$\,\pm\,$1.0
& 130.3$\,\pm\,$1.3
& 0.072$\,\pm\,$0.015 \\

GeoHG~\cite{GeoHG}
& 1457$\,\pm\,$41
& 1889$\,\pm\,$53
& 0.106$\,\pm\,$0.031
& 269.1$\,\pm\,$8.3
& 394.7$\,\pm\,$7.3
& 0.184$\,\pm\,$0.031
& 85.1$\,\pm\,$2.3
& 128.3$\,\pm\,$2.5
& 0.085$\,\pm\,$0.036 \\

GURPP~\cite{GURPP}
& 1517$\,\pm\,$30
& 1930$\,\pm\,$36
& 0.067$\,\pm\,$0.036
& 270.5$\,\pm\,$6.5
& 385.8$\,\pm\,$9.4
& 0.222$\,\pm\,$0.037
& 75.3$\,\pm\,$3.6
& 106.9$\,\pm\,$3.7
& 0.392$\,\pm\,$0.039 \\

FlexiReg~\cite{FlexiReg}
& \underline{1170$\,\pm\,$7}
& \underline{1507$\,\pm\,$13}
& \underline{0.431$\,\pm\,$0.011}
& \underline{197.7$\,\pm\,$4.8}
& \underline{281.9$\,\pm\,$8.5}
& \underline{0.584$\,\pm\,$0.025}
& \underline{48.3$\,\pm\,$0.9}
& \underline{61.5$\,\pm\,$1.7}
& \underline{0.799$\,\pm\,$0.011} \\

\midrule

\textbf{\model}
& \textbf{957.4$\,\pm\,$22.9}
& \textbf{1263$\,\pm\,$23}
& \textbf{0.601$\,\pm\,$0.015}
& \textbf{183.7$\,\pm\,$4.8}
& \textbf{253.2$\,\pm\,$6.5}
& \textbf{0.665$\,\pm\,$0.017}
& \textbf{41.0$\,\pm\,$1.1}
& \textbf{54.0$\,\pm\,$2.1}
& \textbf{0.845$\,\pm\,$0.012} \\

\midrule
\textbf{Improvement}
& \red{\textbf{+18.17\%}}
& \red{\textbf{+16.19\%}}
& \red{\textbf{+39.44\%}}
& \red{\textbf{+7.08\%}}
& \red{\textbf{+10.18\%}}
& \red{\textbf{+13.87\%}}
& \red{\textbf{+15.11\%}}
& \red{\textbf{+12.20\%}}
& \red{\textbf{+5.76\%}} \\

\bottomrule
\end{tabular}
}
\end{table*}

\section{Details of Baseline Models}
\label{subsec:appendix_of_baselines}

We compare with the following models, including two SOTA models GURPP~\cite{GURPP} and FlexiReg~\cite{FlexiReg}:

\begin{itemize}
\item {\textbf{HREP~\cite{HREP}}} uses human mobility, POI, and geographic neighbor features to generate region embeddings. During downstream task learning, it introduces randomly initialized, learnable prompt embeddings, which are concatenated with the region embeddings to adapt them to different downstream tasks.

\item {\textbf{RegionDCL~\cite{RegionDCL}}} partitions the buildings within a region into non-overlapping groups. It learns embeddings for these building groups via contrastive learning, considering both intra-group relations (between each group and its constituent buildings) and inter-level relations (between each group and its corresponding region). These group embeddings are then aggregated to form the region embeddings.

\item {\textbf{UrbanCLIP~\cite{urbanclip}}} generates fine-grained textual descriptions for satellite images associated with each region, forming aligned image–text pairs. The model is then trained on these pairs using contrastive learning to produce text-enhanced visual representations of satellite images, which are adopted as embeddings for the corresponding regions.

\item {\textbf{CityFM~\cite{cityFM}}} leverages geospatial entities (e.g., buildings and road segments) extracted from OpenStreetMap and applies contrastive learning with three objectives to learn entity embeddings, including mutual information–based text-to-text, vision–language, and road-based context-to-context objectives. The learned entity embeddings are aggregated to produce the corresponding region embeddings.

\item {\textbf{GeoHG~\cite{GeoHG}}} extracts spatial features from satellite imagery and POI data, including geo-entities, POI categories, and regional positional information. It  constructs a heterogeneous graph based on these features, where different node and edge types model diverse semantic relations (e.g., region-POI associations). The heterogeneous graph is subsequently processed by a Heterogeneous Graph Neural Network (HGNN)~\cite{hgnn} to learn region embeddings.

\item {\textbf{GURPP~\cite{GURPP}}} constructs a heterogeneous graph from multiple region features, with nodes representing roads, POIs, junctions, brands, and regions. The edges encode semantic relations between different node types (e.g., region-road associations).  A Heterogeneous Graph Transformer (HGT)~\cite{HGT} is applied to the graph to learn region embeddings. For downstream tasks, a prompting mechanism is introduced to inject task-specific information into the pre-trained region embeddings.

\item \textbf{FlexiReg~\cite{FlexiReg}} proposes a three-stage learning framework that leverages POI, land-use, geographic neighborhood, satellite imagery, textual, and street-view imagery features. It first partitions the studied area into grid cells and learns cell embeddings from multiple feature modalities. The cell embeddings are then aggregated to produce region embeddings for the target regions. Finally, during downstream task learning, the region embeddings are further adapted with additional features as guided by downstream tasks.

\end{itemize}

\section{Model Hyperparameter Settings}
\label{subsec:appendix_of_hypepara_settings}
All models were trained and tested on a machine equipped with an NVIDIA Tesla A100 GPU and 80 GB of memory.

For the competitor models, we follow the parameter settings recommended in their respective papers. Except for HREP, none of the competitor models require special configurations, as their training procedures are independent of the number of regions. For HREP, we adopt the settings described in the FlexiReg paper~\cite{FlexiReg} and reduce the number of GNN layers on the CHI dataset.

For our random walk-based embedding learning module, we follow Node2Vec~\cite{grover2016node2vec} to sample random walks. The random walk parameters are set to $p = 1.0$ and $q = 0.1$ to control the depth preference of the walks. Based on the results in Appendix~\ref{subsubsec:para_num_RW}, we sample $k = 8$ independent walks per node, each with a length of $l = 4$. The transformer module consists of three layers, and the masking ratio $\rho$ is set to 0.3. These parameter values are set by a grid search.
The model is trained for 100 epochs using the Adam optimizer with a learning rate of $1\times10^{-7}$.
For our heterogeneous conditional diffusion-based
cross-task learning module, we set the number of diffusion timesteps to $T = 100$ and adopt a linear noise schedule with $\beta_1 = 0.0001$ and $\beta_T = 0.02$, which are often used for diffusion model implementation.  The model is trained for 1,500 epochs using AdamW with a learning rate of $5\times10^{-3}$. Based on the results in Appendix~\ref{subsubsec:para_num_SR}, the number of sampling rounds is set to $\#SR = 10$ by default.

The region embedding dimensionality $d$ is set to 144 for our model, following HREP, GURPP, and FlexiReg. For RegionDCL, UrbanCLIP, CityFM, and GeoHG, the region embedding dimensionalities are set to 64, 768, 1792, and 64, respectively, as recommended in their original papers. According to the embedding dimensionality analysis reported in FlexiReg, these recommended dimensionalities either yield optimal performance or are difficult to adjust due to the model design. Therefore, we adopt the dimensionalities specified in the original papers for these models.



\section{Full Results on Overall Model Comparison}
\label{sec:appendix_of_cross_city_result}

Tables~\ref{tab:cross_city_full_p1} and~\ref{tab:cross_city_full_p2} present the complete results on an overall comparison of \model\ against the baseline models, on the three US cities NYC, CHI, and SF for six downstream tasks, evaluated with three performance metrics MAE, RMSE, and $R^2$. It can be seen that \model\ achieves the best results consistently across all  datasets,  downstream tasks, and evaluation metrics.

\section{Model Comparison under Same-City Settings}
\label{sec:appendix_of_same_city_result}

Tables~\ref{tab:same_city_full_p1} and~\ref{tab:same_city_full_p2} report the complete results on a comparison between \model\ and the baseline models under the same-city settings, where all models are trained and evaluated over the same dataset. All baseline methods were designed for such settings. 

Following the same experimental protocol as before, we conduct experiments on three cities NYC, CHI, and SF across six downstream tasks, evaluated using MAE, RMSE, and $R^2$. 
The  baseline models perform better under same-city settings than under cross-city settings, underscoring the difficulties brought by cross-city generalization. 
Still, \model\ achieves the best accuracy in 7 out of the 18 test scenarios and the second-best accuracy in another 10 test scenarios,  demonstrating that \model\ remains highly competitive under the same-city settings.

\begin{table*}[htbp]
\centering
\caption{Overall \emph{Same-city} Prediction Accuracy Results on \emph{Crime, Check-in, and Service Call} Prediction Tasks.}
\label{tab:same_city_full_p1}

\setlength{\tabcolsep}{2pt}
\renewcommand{\arraystretch}{1}
\resizebox{\textwidth }{!}{
\begin{tabular}{l|ccc|ccc|ccc}

\toprule [0.4ex] \\[-3.5ex]

\multirow{2}{*}{\textbf{\makecell[c]{NYC}}} &
\multicolumn{3}{c|}{Crime} & \multicolumn{3}{c|}{Check-in} & \multicolumn{3}{c}{Service Call}  \\

\cmidrule(lr){2-4} \cmidrule(lr){5-7} \cmidrule(lr){8-10}
    
&MAE $\downarrow$ & RMSE $\downarrow$ & $R^{2} \uparrow$ 
& MAE $\downarrow$ & RMSE $\downarrow$ & $R^{2} \uparrow$ 
& MAE $\downarrow$ & RMSE $\downarrow$ & $R^{2} \uparrow$ \\
\midrule

HREP~\cite{HREP}
& 62.8$\,\pm\,$2.1
& 83.1$\,\pm\,$2.3
& 0.681$\,\pm\,$0.014
& 276.3$\,\pm\,$11.7
& 448.2$\,\pm\,$16.9
& 0.700$\,\pm\,$0.022
& 1430$\,\pm\,$29
& 2127$\,\pm\,$33
& 0.398$\,\pm\,$0.019 \\

RegionDCL~\cite{RegionDCL}
& 98.7$\,\pm\,$0.7
& 127.9$\,\pm\,$1.1
& 0.251$\,\pm\,$0.025
& 371.2$\,\pm\,$10.3
& 595.5$\,\pm\,$5.9
& 0.471$\,\pm\,$0.023
& 1783$\,\pm\,$22
& 2597$\,\pm\,$18
& 0.103$\,\pm\,$0.026 \\

UrbanCLIP~\cite{urbanclip}
& 97.4$\,\pm\,$2.6
& 126.1$\,\pm\,$1.9
& 0.267$\,\pm\,$0.012
& 393.6$\,\pm\,$5.9
& 602.4$\,\pm\,$3.1
& 0.458$\,\pm\,$0.005
& 1409$\,\pm\,$7
& 2401$\,\pm\,$15
& 0.232$\,\pm\,$0.005 \\

CityFM~\cite{cityFM}
& 95.5$\,\pm\,$1.4
& 122.4$\,\pm\,$1.8
& 0.315$\,\pm\,$0.010
& 380.2$\,\pm\,$3.8
& 594.9$\,\pm\,$6.4
& 0.471$\,\pm\,$0.011
& 1780$\,\pm\,$27
& 2577$\,\pm\,$19
& 0.117$\,\pm\,$0.013 \\

GeoHG~\cite{GeoHG}
& 79.5$\,\pm\,$1.5
& 116.9$\,\pm\,$2.1
& 0.374$\,\pm\,$0.019
& 388.1$\,\pm\,$11.9
& 621.5$\,\pm\,$34.1
& 0.422$\,\pm\,$0.032
& 1484$\,\pm\,$29
& 2325$\,\pm\,$37
& 0.281$\,\pm\,$0.023 \\

GURPP~\cite{GURPP}
& 72.9$\,\pm\,$2.3
& 94.7$\,\pm\,$4.6
& 0.589$\,\pm\,$0.039
& 267.0$\,\pm\,$15.2
& 403.3$\,\pm\,$16.5
& 0.757$\,\pm\,$0.019
& 1480$\,\pm\,$33
& 2115$\,\pm\,$84
& 0.405$\,\pm\,$0.048 \\

FlexiReg~\cite{FlexiReg}
& \textbf{50.4$\,\pm\,$1.12}
& \textbf{67.6$\,\pm\,$1.52}
& \textbf{0.789$\,\pm\,$0.009}
& \textbf{187.3$\,\pm\,$5.01}
& \textbf{287.5$\,\pm\,$7.57}
& \textbf{0.876$\,\pm\,$0.006}
& \underline{1131$\,\pm\,$46}
& \textbf{1726$\,\pm\,$45}
& \textbf{0.601$\,\pm\,$0.021} \\

\midrule

\model\
& \underline{57.9$\,\pm\,$1.4}
& \underline{78.3$\,\pm\,$2.1}
& \underline{0.719$\,\pm\,$0.008}
& \underline{239.3$\,\pm\,$4.9}
& \underline{378.2$\,\pm\,$9.4}
& \underline{0.786$\,\pm\,$0.012}
& \textbf{1120$\,\pm\,$19}
& \underline{1747$\,\pm\,$36}
& \underline{0.594$\,\pm\,$0.011} \\


\bottomrule \toprule [0.4ex] \\[-3.5ex]

\multirow{2}{*}{\textbf{\makecell[c]{CHI}}} &
\multicolumn{3}{c|}{Crime} & \multicolumn{3}{c|}{Check-in} & \multicolumn{3}{c}{Service Call}  \\

\cmidrule(lr){2-4} \cmidrule(lr){5-7} \cmidrule(lr){8-10}
    
&MAE $\downarrow$ & RMSE $\downarrow$ & $R^{2} \uparrow$ 
& MAE $\downarrow$ & RMSE $\downarrow$ & $R^{2} \uparrow$ 
 & MAE $\downarrow$ & RMSE $\downarrow$ & $R^{2} \uparrow$ \\
\midrule

HREP~\cite{HREP}
& 88.3$\,\pm\,$6.38
& 114.4$\,\pm\,$5.48
& 0.578$\,\pm\,$0.41
& 1702$\,\pm\,$79
& 3300$\,\pm\,$100
& 0.628$\,\pm\,$0.023
& 185.7$\,\pm\,$6.1
& 262.2$\,\pm\,$5.84
& 0.468$\,\pm\,$0.021 \\

RegionDCL~\cite{RegionDCL} 
& 121.7$\,\pm\,$2.3
& 159.6$\,\pm\,$2.8
& 0.179$\,\pm\,$0.028
& 2427$\,\pm\,$123
& 4184$\,\pm\,$136
& 0.402$\,\pm\,$0.042
& 195.7$\,\pm\,$7.6
& 267.1$\,\pm\,$10.1
& 0.445$\,\pm\,$0.041 \\

UrbanCLIP~\cite{urbanclip} 
& 101.6$\,\pm\,$0.6
& 134.7$\,\pm\,$1.7
& 0.416$\,\pm\,$0.006
& 2611$\,\pm\,$29
& 4884$\,\pm\,$72
& 0.186$\,\pm\,$0.024
& 183.2$\,\pm\,$0.9
& 256.3$\,\pm\,$1.8
& 0.491$\,\pm\,$0.003 \\

CityFM~\cite{cityFM}
& 121.6$\,\pm\,$1.8
& 157.1$\,\pm\,$2.8
& 0.205$\,\pm\,$0.018
& 1980$\,\pm\,$64
& 3362$\,\pm\,$109
& 0.614$\,\pm\,$0.025
& 198.3$\,\pm\,$3.7
& 280.1$\,\pm\,$6.1
& 0.391$\,\pm\,$0.027 \\

GeoHG~\cite{GeoHG} 
& 87.8$\,\pm\,$2.9
& 134.3$\,\pm\,$4.8
& 0.502$\,\pm\,$0.029
& 2215$\,\pm\,$70
& 4089$\,\pm\,$133
& 0.439$\,\pm\,$0.027
& 201.1$\,\pm\,$7.4
& 273.6$\,\pm\,$12.7
& 0.419$\,\pm\,$0.035 \\

GURPP~\cite{GURPP}
& 94.0$\,\pm\,$1.7
& 122.1$\,\pm\,$2.9
& 0.557$\,\pm\,$0.021
& 2473$\,\pm\,$124
& 4165$\,\pm\,$215
& 0.388$\,\pm\,$0.054
& 197.1$\,\pm\,$4.3
& 270.2$\,\pm\,$5.7
& 0.433$\,\pm\,$0.025 \\

FlexiReg~\cite{FlexiReg}
& \underline{61.7$\,\pm\,$3.5}
& \underline{85.1$\,\pm\,$4.2}
& \underline{0.766$\,\pm\,$0.022}
& \textbf{922$\,\pm\,$76}
& \textbf{1775$\,\pm\,$198}
& \textbf{0.891$\,\pm\,$0.024}
& \textbf{121.1$\,\pm\,$7.4}
& \textbf{178.2$\,\pm\,$9.5}
& \textbf{0.753$\,\pm\,$0.026} \\

\midrule
\model\
& \textbf{59.2$\,\pm\,$1.9}
& \textbf{83.4$\,\pm\,$3.8}
& \textbf{0.776$\,\pm\,$0.016}
& \underline{1119$\,\pm\,$71}
& \underline{2339$\,\pm\,$167}
& \underline{0.805$\,\pm\,$0.026}
& \underline{131.4$\,\pm\,$4.6}
& \underline{193.6$\,\pm\,$8.4}
& \underline{0.709$\,\pm\,$0.019} \\


\bottomrule \toprule [0.4ex] \\[-3.5ex]

\multirow{2}{*}{\textbf{\makecell[c]{SF}}} &
\multicolumn{3}{c|}{Crime} & \multicolumn{3}{c|}{Check-in} & \multicolumn{3}{c}{Service Call}  \\

\cmidrule(lr){2-4} \cmidrule(lr){5-7} \cmidrule(lr){8-10}
    
&MAE $\downarrow$ & RMSE $\downarrow$ & $R^{2} \uparrow$ 
& MAE $\downarrow$ & RMSE $\downarrow$ & $R^{2} \uparrow$ 
 & MAE $\downarrow$ & RMSE $\downarrow$ & $R^{2} \uparrow$ \\
\midrule

HREP~\cite{HREP}
& 124.4$\,\pm\,$2.3
& 196.9$\,\pm\,$3.9
& 0.612$\,\pm\,$0.014
& 330.9$\,\pm\,$9.3
& 606.7$\,\pm\,$25.8
& 0.629$\,\pm\,$0.032
& 103.4$\,\pm\,$3.2
& 167.4$\,\pm\,$4.6
& 0.461$\,\pm\,$0.029 \\

RegionDCL~\cite{RegionDCL} 
& 156.3$\,\pm\,$1.2
& 242.3$\,\pm\,$3.2
& 0.413$\,\pm\,$0.021
& 398.8$\,\pm\,$9.9
& 748.1$\,\pm\,$17.8
& 0.437$\,\pm\,$0.024
& 116.6$\,\pm\,$1.8
& 196.7$\,\pm\,$1.2
& 0.256$\,\pm\,$0.018 \\

UrbanCLIP~\cite{urbanclip} 
& 171.1$\,\pm\,$1.0
& 269.8$\,\pm\,$2.6
& 0.283$\,\pm\,$0.014
& 380.3$\,\pm\,$2.6
& 813.5$\,\pm\,$3.3
& 0.334$\,\pm\,$0.002
& 106.9$\,\pm\,$1.1
& 192.1$\,\pm\,$1.2
& 0.292$\,\pm\,$0.008 \\

CityFM~\cite{cityFM} 
& 168.3$\,\pm\,$0.6
& 259.8$\,\pm\,$1.5
& 0.334$\,\pm\,$0.008
& 428.3$\,\pm\,$2.7
& 839.3$\,\pm\,$4.7
& 0.298$\,\pm\,$0.008
& 105.1$\,\pm\,$1.3
& 178.5$\,\pm\,$1.6
& 0.396$\,\pm\,$0.008 \\

GeoHG~\cite{GeoHG}
& 166.8$\,\pm\,$2.2
& 262.5$\,\pm\,$3.4
& 0.311$\,\pm\,$0.019
& 437.8$\,\pm\,$7.9
& 845.6$\,\pm\,$6.4
& 0.277$\,\pm\,$0.012
& 118.0$\,\pm\,$3.3
& 197.4$\,\pm\,$7.9
& 0.251$\,\pm\,$0.039 \\

GURPP~\cite{GURPP}
& 139.1$\,\pm\,$5.6
& 211.8$\,\pm\,$6.9
& 0.552$\,\pm\,$0.027
& \underline{261.6$\,\pm\,$15.1}
& \underline{417.1$\,\pm\,$20.5}
& \underline{0.824$\,\pm\,$0.016}
& 105.8$\,\pm\,$4.3
& 184.2$\,\pm\,$5.8
& 0.348$\,\pm\,$0.035 \\

FlexiReg~\cite{FlexiReg}
& \underline{98.6$\,\pm\,$3.9}
& \underline{163.7$\,\pm\,$4.3}
& \underline{0.732$\,\pm\,$0.014}
& \textbf{229.4$\,\pm\,$8.1}
& \textbf{375.2$\,\pm\,$34.9}
& \textbf{0.859$\,\pm\,$0.011}
& \textbf{79.9$\,\pm\,$4.6}
& \textbf{136.5$\,\pm\,$3.8}
& \textbf{0.641$\,\pm\,$0.021} \\

\midrule

\model\
& \textbf{95.1$\,\pm\,$2.6}
& \textbf{138.0$\,\pm\,$3.8}
& \textbf{0.810$\,\pm\,$0.009}
& 279.2$\,\pm\,$10.3
& 464.5$\,\pm\,$15.8
& 0.781$\,\pm\,$0.013
& \underline{86.9$\,\pm\,$2.5}
& \underline{152.1$\,\pm\,$4.3}
& \underline{0.558$\,\pm\,$0.018} \\


\bottomrule
\end{tabular}
}
\end{table*}

\begin{table*}[htbp]
\centering
\caption{Overall \emph{Same-city} Prediction Accuracy Results on \emph{Population, Carbon, and Nightlight} Prediction Tasks.}
\label{tab:same_city_full_p2}

\setlength{\tabcolsep}{2pt}
\renewcommand{\arraystretch}{1}
\resizebox{\textwidth }{!}{
\begin{tabular}{l|ccc|ccc|ccc}

\toprule [0.4ex] \\[-3.5ex]

\multirow{2}{*}{\textbf{\makecell[c]{NYC}}} &
\multicolumn{3}{c|}{Population} & \multicolumn{3}{c|}{Carbon} & \multicolumn{3}{c}{Nightlight}  \\

\cmidrule(lr){2-4} \cmidrule(lr){5-7} \cmidrule(lr){8-10}
    
&MAE $\downarrow$ & RMSE $\downarrow$ & $R^{2} \uparrow$ 
& MAE $\downarrow$ & RMSE $\downarrow$ & $R^{2} \uparrow$ 
 & MAE $\downarrow$ & RMSE $\downarrow$ & $R^{2} \uparrow$ \\
\midrule

HREP~\cite{HREP}
& 2656$\,\pm\,$59
& 3461$\,\pm\,$83
& 0.571$\,\pm\,$0.021
& 719$\,\pm\,$35
& 1188$\,\pm\,$37
& 0.184$\,\pm\,$0.025
& 58.7$\,\pm\,$1.1
& 78.7$\,\pm\,$1.3
& 0.026$\,\pm\,$0.024 \\

RegionDCL~\cite{RegionDCL}
& 3753$\,\pm\,$47
& 4734$\,\pm\,$59
& 0.198$\,\pm\,$0.019
& 817$\,\pm\,$31
& 1272$\,\pm\,$42
& 0.064$\,\pm\,$0.037
& 55.8$\,\pm\,$0.5
& 75.4$\,\pm\,$1.0
& 0.057$\,\pm\,$0.015 \\

UrbanCLIP~\cite{urbanclip}
& 3338$\,\pm\,$11
& 4499$\,\pm\,$16
& 0.276$\,\pm\,$0.002
& 730$\,\pm\,$13
& 1238$\,\pm\,$19
& 0.113$\,\pm\,$0.013
& 48.9$\,\pm\,$0.3
& 63.2$\,\pm\,$1.0
& 0.337$\,\pm\,$0.019 \\

CityFM~\cite{cityFM}
& 3515$\,\pm\,$18
& 4545$\,\pm\,$26
& 0.261$\,\pm\,$0.002
& 751$\,\pm\,$12
& 1253$\,\pm\,$21
& 0.092$\,\pm\,$0.014
& 56.2$\,\pm\,$0.4
& 75.8$\,\pm\,$0.8
& 0.047$\,\pm\,$0.016 \\

GeoHG~\cite{GeoHG}
& 3578$\,\pm\,$52
& 4685$\,\pm\,$42
& 0.215$\,\pm\,$0.011
& 618$\,\pm\,$10
& 1236$\,\pm\,$16
& 0.117$\,\pm\,$0.009
& 51.7$\,\pm\,$1.0
& 70.5$\,\pm\,$1.6
& 0.176$\,\pm\,$0.038 \\

GURPP~\cite{GURPP}
& 3316$\,\pm\,$75
& 4159$\,\pm\,$87
& 0.381$\,\pm\,$0.026
& 782.3$\,\pm\,$1.2
& 1289$\,\pm\,$3
& 0.021$\,\pm\,$0.001
& 59.1$\,\pm\,$0.2
& 78.1$\,\pm\,$0.3
& 0.012$\,\pm\,$0.005 \\

FlexiReg~\cite{FlexiReg}
& \textbf{2159$\,\pm\,$28}
& \textbf{2822$\,\pm\,$47}
& \textbf{0.715$\,\pm\,$0.010}
& \underline{576$\,\pm\,$21}
& \underline{1063$\,\pm\,$27}
& \underline{0.346$\,\pm\,$0.017}
& \textbf{42.3$\,\pm\,$0.6}
& \textbf{53.6$\,\pm\,$0.8}
& \textbf{0.523$\,\pm\,$0.009} \\

\midrule

\model\
& \underline{2516$\,\pm\,$25}
& \underline{3266$\,\pm\,$41}
& \underline{0.619$\,\pm\,$0.009}
& \textbf{566$\,\pm\,$12}
& \textbf{1031$\,\pm\,$17}
& \textbf{0.385$\,\pm\,$0.008}
& \underline{43.0$\,\pm\,$0.7}
& \underline{54.9$\,\pm\,$0.9}
& \underline{0.498$\,\pm\,$0.013} \\


\bottomrule \toprule [0.4ex] \\[-3.5ex]

\multirow{2}{*}{\textbf{\makecell[c]{CHI}}} &
\multicolumn{3}{c|}{Population} & \multicolumn{3}{c|}{Carbon} & \multicolumn{3}{c}{Nightlight}  \\

\cmidrule(lr){2-4} \cmidrule(lr){5-7} \cmidrule(lr){8-10}
    
&MAE $\downarrow$ & RMSE $\downarrow$ & $R^{2} \uparrow$ 
& MAE $\downarrow$ & RMSE $\downarrow$ & $R^{2} \uparrow$ 
 & MAE $\downarrow$ & RMSE $\downarrow$ & $R^{2} \uparrow$ \\
\midrule

HREP~\cite{HREP}
& 12063$\,\pm\,$539
& 15397$\,\pm\,$832
& 0.447$\,\pm\,$0.061
& 398.2$\,\pm\,$12.1
& 588.9$\,\pm\,$14.6
& 0.102$\,\pm\,$0.039
& 380.9$\,\pm\,$18.8
& 635.6$\,\pm\,$25.4
& 0.210$\,\pm\,$0.037 \\

RegionDCL~\cite{RegionDCL}
& 14289$\,\pm\,$343
& 18653$\,\pm\,$368
& 0.190$\,\pm\,$0.032
& 402.1$\,\pm\,$9.3
& 591.6$\,\pm\,$13.1
& 0.094$\,\pm\,$0.031
& 448.5$\,\pm\,$8.9
& 639.9$\,\pm\,$15.3
& 0.201$\,\pm\,$0.021 \\

UrbanCLIP~\cite{urbanclip}
& 13328$\,\pm\,$69
& 17498$\,\pm\,$74
& 0.288$\,\pm\,$0.006
& 381.9$\,\pm\,$7.6
& 573.5$\,\pm\,$11.4
& 0.149$\,\pm\,$0.025
& 231.9$\,\pm\,$3.8
& 349$\,\pm\,$5.4
& 0.762$\,\pm\,$0.008 \\

CityFM~\cite{cityFM}
& 13904$\,\pm\,$37
& 17704$\,\pm\,$56
& 0.271$\,\pm\,$0.004
& 404.5$\,\pm\,$2.7
& 603.8$\,\pm\,$4.6
& 0.063$\,\pm\,$0.009
& 490.1$\,\pm\,$6.5
& 672.0$\,\pm\,$11.8
& 0.118$\,\pm\,$0.013 \\

GeoHG~\cite{GeoHG}
& 13978$\,\pm\,$68
& 17628$\,\pm\,$133
& 0.277$\,\pm\,$0.028
& 398.3$\,\pm\,$6.9
& 544.2$\,\pm\,$12.3
& 0.233$\,\pm\,$0.018
& 482.6$\,\pm\,$11.6
& 665.7$\,\pm\,$14.5
& 0.134$\,\pm\,$0.012 \\

GURPP~\cite{GURPP}
& 10983$\,\pm\,$179
& 14513$\,\pm\,$215
& 0.510$\,\pm\,$0.031
& 365.4$\,\pm\,$9.7
& 569.9$\,\pm\,$12.4
& 0.159$\,\pm\,$0.021
& 471.2$\,\pm\,$9.6
& 647.7$\,\pm\,$13.5
& 0.180$\,\pm\,$0.017 \\

FlexiReg~\cite{FlexiReg}
& \underline{8126$\,\pm\,$224}
& \underline{11395$\,\pm\,$255}
& \underline{0.698$\,\pm\,$0.014}
& \underline{258.8$\,\pm\,$11.2}
& \underline{421.7$\,\pm\,$15.4}
& \underline{0.539$\,\pm\,$0.012}
& \textbf{142.1$\,\pm\,$7.3}
& \textbf{202.3$\,\pm\,$12.7}
& \textbf{0.935$\,\pm\,$0.011} \\

\midrule

\model\
& \textbf{7460$\,\pm\,$289}
& \textbf{9390$\,\pm\,$317}
& \textbf{0.795$\,\pm\,$0.011}
& \textbf{255.8$\,\pm\,$9.2}
& \textbf{420.7$\,\pm\,$14.5}
& \textbf{0.542$\,\pm\,$0.015}
& \underline{166.4$\,\pm\,$10.9}
& \underline{238.8$\,\pm\,$21.7}
& \underline{0.889$\,\pm\,$0.009} \\


\bottomrule \toprule [0.4ex] \\[-3.5ex]

\multirow{2}{*}{\textbf{\makecell[c]{SF}}} &
\multicolumn{3}{c|}{Population} & \multicolumn{3}{c|}{Carbon} & \multicolumn{3}{c}{Nightlight}  \\

\cmidrule(lr){2-4} \cmidrule(lr){5-7} \cmidrule(lr){8-10}
    
&MAE $\downarrow$ & RMSE $\downarrow$ & $R^{2} \uparrow$ 
& MAE $\downarrow$ & RMSE $\downarrow$ & $R^{2} \uparrow$ 
 & MAE $\downarrow$ & RMSE $\downarrow$ & $R^{2} \uparrow$ \\
\midrule

HREP~\cite{HREP}
& 1436$\,\pm\,$40
& 1867$\,\pm\,$45
& 0.127$\,\pm\,$0.023
& 248.3$\,\pm\,$6.4
& 375.4$\,\pm\,$11.2
& 0.264$\,\pm\,$0.022
& 73.9$\,\pm\,$2.3
& 108.2$\,\pm\,$3.1
& 0.378$\,\pm\,$0.041 \\

RegionDCL~\cite{RegionDCL}
& 1513$\,\pm\,$32
& 1971$\,\pm\,$25
& 0.027$\,\pm\,$0.025
& 277.1$\,\pm\,$3.1
& 417.5$\,\pm\,$4.6
& 0.089$\,\pm\,$0.013
& 76.2$\,\pm\,$0.8
& 119.1$\,\pm\,$1.7
& 0.246$\,\pm\,$0.009 \\

UrbanCLIP~\cite{urbanclip}
& 1695$\,\pm\,$17
& 2360$\,\pm\,$34
& 0.395$\,\pm\,$0.005
& 251.9$\,\pm\,$2.8
& 363.4$\,\pm\,$4.1
& 0.311$\,\pm\,$0.012
& 62.6$\,\pm\,$0.7
& 78.5$\,\pm\,$2.6
& 0.673$\,\pm\,$0.021 \\

CityFM~\cite{cityFM}
& 1578$\,\pm\,$14
& 1982$\,\pm\,$28
& 0.023$\,\pm\,$0.003
& 277.6$\,\pm\,$2.7
& 415.2$\,\pm\,$5.6
& 0.082$\,\pm\,$0.009
& 85.4$\,\pm\,$1.1
& 128.4$\,\pm\,$2.1
& 0.088$\,\pm\,$0.014 \\

GeoHG~\cite{GeoHG}
& 1429$\,\pm\,$35
& 1829$\,\pm\,$53
& 0.161$\,\pm\,$0.031
& 265.7$\,\pm\,$3.5
& 379.9$\,\pm\,$5.3
& 0.243$\,\pm\,$0.015
& 84.6$\,\pm\,$1.3
& 126.9$\,\pm\,$2.2
& 0.095$\,\pm\,$0.019 \\

GURPP~\cite{GURPP}
& 1484$\,\pm\,$34
& 1883$\,\pm\,$41
& 0.111$\,\pm\,$0.032
& 258.7$\,\pm\,$9.3
& 358.1$\,\pm\,$12.3
& 0.330$\,\pm\,$0.041
& 71.2$\,\pm\,$3.3
& 103.8$\,\pm\,$4.8
& 0.428$\,\pm\,$0.036 \\

FlexiReg~\cite{FlexiReg}
& \underline{1032$\,\pm\,$29}
& \underline{1441$\,\pm\,$46}
& \underline{0.480$\,\pm\,$0.034}
& \underline{188.4$\,\pm\,$5.1}
& \underline{264.8$\,\pm\,$9.2}
& \underline{0.637$\,\pm\,$0.019}
& \textbf{39.4$\,\pm\,$0.7}
& \textbf{52.7$\,\pm\,$1.4}
& \textbf{0.864$\,\pm\,$0.013} \\

\midrule

\model\
& \textbf{943$\,\pm\,$21}
& \textbf{1252$\,\pm\,$27}
& \textbf{0.607$\,\pm\,$0.013}
& \textbf{185.7$\,\pm\,$3.9}
& \textbf{256.0$\,\pm\,$7.1}
& \textbf{0.658$\,\pm\,$0.014}
& \underline{39.9$\,\pm\,$0.8}
& \underline{54.9$\,\pm\,$1.9}
& \underline{0.840$\,\pm\,$0.009} \\


\bottomrule
\end{tabular}
}
\end{table*}



\section{Additional Results on Model Adaptability to Suburban Area}
\label{sec:appendix_of_suburban_area}

\begin{table}[htbp]
\caption{{Model Applicability to Suburban Areas under Same-city Setting}.}
\label{tab:suburban_sameCity}

\setlength{\tabcolsep}{4pt}
\renewcommand{\arraystretch}{1.05}
\resizebox{\columnwidth}{!}{
\begin{tabular}{l | c c c c}
\toprule
\multirow{2}{*}{ \textbf{\makecell[l]{Staten Island}}}
& \textbf{Check-in}
& \textbf{Population}
& \textbf{Carbon}
& \textbf{Nightlight} \\

\cmidrule(lr){2-2} \cmidrule(lr){3-3} \cmidrule(lr){4-4} \cmidrule(lr){5-5} 

& $R^{2} \uparrow$ & $R^{2} \uparrow$ & $R^{2} \uparrow$ & $R^{2} \uparrow$ \\
\midrule

HREP~\cite{HREP}
& 0.136$\,\pm\,$0.038
& 0.192$\,\pm\,$0.028
& 0.126$\,\pm\,$0.015
& 0.239$\,\pm\,$0.021 \\

RegionDCL~\cite{RegionDCL}
& 0.061$\,\pm\,$0.056
& 0.028$\,\pm\,$0.021
& 0.237$\,\pm\,$0.029
& 0.399$\,\pm\,$0.021 \\

UrbanCLIP~\cite{urbanclip}
& 0.072$\,\pm\,$0.014
& 0.088$\,\pm\,$0.013
& 0.557$\,\pm\,$0.016
& 0.605$\,\pm\,$0.015 \\

CityFM~\cite{cityFM}
& 0.067$\,\pm\,$0.047
& 0.098$\,\pm\,$0.015
& 0.079$\,\pm\,$0.015
& 0.079$\,\pm\,$0.031 \\


GeoHG~\cite{GeoHG}
& 0.094$\,\pm\,$0.035
& 0.023$\,\pm\,$0.012
& 0.047$\,\pm\,$0.028
& 0.103$\,\pm\,$0.049 \\

GURPP~\cite{GURPP}
& 0.153$\,\pm\,$0.038
& 0.384$\,\pm\,$0.027
& 0.366$\,\pm\,$0.038
& 0.488$\,\pm\,$0.033 \\

FlexiReg~\cite{FlexiReg}
& \underline{0.477$\,\pm\,$0.034}
& \underline{0.412$\,\pm\,$0.018}
& \underline{0.716$\,\pm\,$0.012}
& \underline{0.899$\,\pm\,$0.009} \\

\midrule
\textbf{\model}
& \textbf{0.524$\,\pm\,$0.029}
& \textbf{0.589$\,\pm\,$0.021}
& \textbf{0.772$\,\pm\,$0.008}
& \textbf{0.939$\,\pm\,$0.006} \\

\midrule
\textbf{Improvement}
& \textbf{\red{+9.8\%}}
& \textbf{\red{+42.9\%}}
& \textbf{\red{+7.8\%}}
& \textbf{\red{+4.4\%}} \\

\bottomrule
\end{tabular}
}
\end{table}

Table~\ref{tab:suburban_sameCity} reports additional results on the Staten Island dataset, where all models are trained and tested with Staten Island data. A few of the baseline models (e.g., the SOTA models GURPP and FlexiReg) achieve better accuracy under this more favorable setting than under the cross-city setting as reported in Table~\ref{tab:suburban_crossCity}. However, due to the scarcity of available data in this suburban area, there is still a substantial gap in the accuracy of these models and our \model\ model.  

This observation suggests that city-specific models are particularly vulnerable in data-limited environments, thereby underscoring the necessity of cross-city learning and further emphasizing the robustness of our solution.


\section{Full Results on the Applicability of the \diffM\ Module}
\label{sec:appendix_of_app_diffusionM}

Table~\ref{tab:app_diffusionM_full} presents the complete results of integrating \diffM~into existing models. All \diffM~enhanced model variants achieve substantial accuracy improvements, indicating both the effectiveness and the  applicability of \diffM.

\section{Additional Experimental Results}
\label{sec:additional_exp_results}

This section reports additional experimental results on model running time, predictive distribution visualization for sample regions, and the impact of model parameter values.

\begin{table}[ht]
\centering
\caption{Embedding Learning and Test Times (seconds).}
\label{tab:computation_time}
\renewcommand{\arraystretch}{1.1}
\resizebox{1\columnwidth}{!}{
\begin{tabular}{l|rrr|rrr}
\toprule

&\multicolumn{3}{c|}{\textbf{Embedding Learning}} & \multicolumn{3}{c}{\textbf{Downstream Task}} \\ \hline 
& \textbf{NYC} & \textbf{CHI} & \textbf{SF} & \multicolumn{1}{c}{\textbf{NYC}} & \multicolumn{1}{c}{\textbf{CHI}} & \multicolumn{1}{c}{\textbf{SF}} \\ 
\hline

HREP& 97 & 102 & 98 &92 (0.003) &146 (0.005) &91 (0.004)\\ 
\hline

RegionDCL & 2,099 & 475 & 1,925 & \textbf{0.017} (0.001) & \textbf{0.054} (0.002) &\textbf{0.023 (0.001)}\\ 
\hline

UrbanCLIP & 5,817 & 4,911 & 4,001 & 86 (0.005) &86 (0.005) & 84 (0.005)\\ 
\hline

CityFM & 15,301 & 14,571 & 15,427 & 104 (0.006)&102 (0.005) &104 (0.006)\\ 
\hline

GeoHG &\textbf{90} &107 &105 & 96 (0.004) & 95 (0.004) &98 (0.004)\\ 
\hline

GURPP &2882 &1224 &3130 & 121 (0.006) & 92 (0.005) &126 (0.006)\\ 
\hline

FlexiReg &712 &645 &485 & 137 (0.007) &103 (0.006) &142 (0.007)\\ 
\hline

\textbf{\model} &109 & \textbf{93} & \textbf{74} & 198 (0.697) &161 (0.696) &324 (0.706)\\ 


\bottomrule
\end{tabular}
}
\end{table}

\subsection{Model Running Time}
\label{subsec:appendix_of_model_running_time}

Table~\ref{tab:computation_time} reports the running times for embedding learning and downstream task learning. 
Since our experiments are conducted under a cross-city setting with three datasets, models are trained on two of the three cities and evaluated on the remaining one. For example, for NYC embedding learning, each model is trained on the CHI and SF datasets, and we report the corresponding training time.
The downstream task running time includes both model training and inference, with inference times shown in parentheses.

\model\ requires less time for embedding learning due to its region-centric design, which represents each region
independently using multiple random walks originating from the target region,  rather than modeling an entire city as a single graph. As a result, \model’s computational cost is largely independent of the overall graph size, whereas several baseline methods are significantly affected by it. 
For example, FlexiReg starts with learning embeddings for the cells, leading to higher costs due to the much larger cell graph. GURPP also takes extra time because it constructs a large heterogeneous graph with many different node types and edge types.  RegionDCL and CityFM incur extra embedding learning overhead because they model fine-grained geospatial entities (e.g., buildings and road segments) whose number far exceeds the number of regions. 

Meanwhile, \model\ takes additional time for downstream task training because it is jointly trained using data from six tasks, whereas the baseline methods are trained separately for a single task. It is important to note, however, that \model\ is trained only once and can be directly applied to all six downstream tasks without any retraining. As a result, when predictions for multiple tasks are required, \model\ is substantially more time-efficient than the baseline models. \model\ also  significantly reduces prediction errors, as demonstrated in the preceding experiments.
\model\ requires slightly more time for downstream task inference, since the heterogeneous conditional diffusion-based
cross-task learning module needs to sample 10 times for each prediction. Nevertheless, the total inference time remains under one second, which is negligible in practice.

\begin{table}[htbp] 
\caption{Impact of the Number of Cells on  Embedding Learning Time (seconds, based on SF data).}
\label{tab:model_runtime_scalability}
\begin{center}
\setlength{\tabcolsep}{6pt}
\resizebox{\columnwidth }{!}{
\begin{tabular}{l*{7}{c}}
    \toprule
    Number of Cells & 256 & 512 & 1,024 & 2,048 & 4,096 & 8,192 & 16,384 \\
    \midrule
    
   Training time & 27 & 50 & 97 & 191 & 376 & 752 & 1,495\\

    \bottomrule
\end{tabular}
}
\end{center}
\end{table}

We further evaluate \model’s embedding learning time when the number of cells is varied from 256 to 16,384, while fixing the batch size at 1,024. We use the SF dataset for this experiment. SF has 1,032 cells. We first run random walks to construct a sequence for each cell as usual. To obtain datasets with fewer or more cells, we down-sample or up-sample the sequences, respectively (since the focus here is the time of learning rather than the quality). 

As shown in Table~\ref{tab:model_runtime_scalability}, the embedding learning time increases almost linearly with the number of cells. When using 16,384 cells, the total embedding learning time is about 30 minutes. These results demonstrate that our model scales efficiently and is practical for real-world urban-scale applications.


\subsection{Predictive Distribution Visualization}
\label{subsec:appendix_of_pred_distribution_visu}

\begin{figure}[htbp]
    \centering
    \begin{subfigure}[b]{\columnwidth}
        \centering
        \includegraphics[width=0.6\columnwidth]{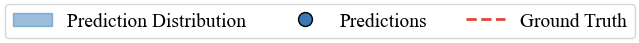}
    \end{subfigure}
    
    \begin{subfigure}[b]{0.32\columnwidth}
        \centering
        \includegraphics[width=0.93\columnwidth]{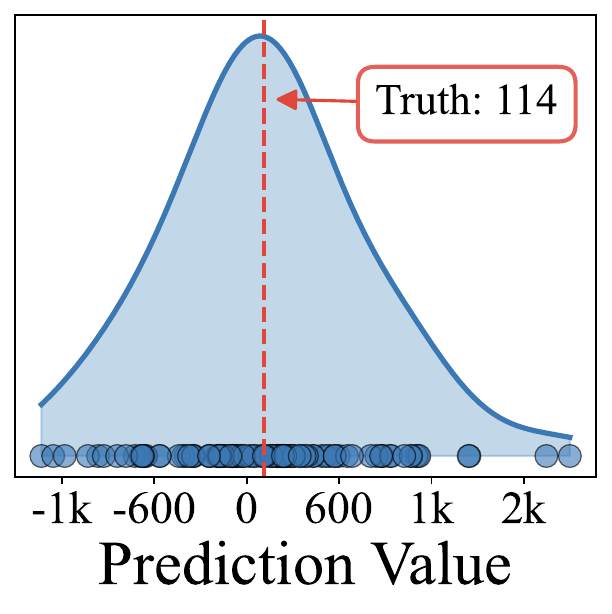}
        \vspace{-2mm}
        \caption{Crime (SF)}
    \end{subfigure} 
    \begin{subfigure}[b]{0.32\columnwidth}
        \centering
        \includegraphics[width=0.96\columnwidth]{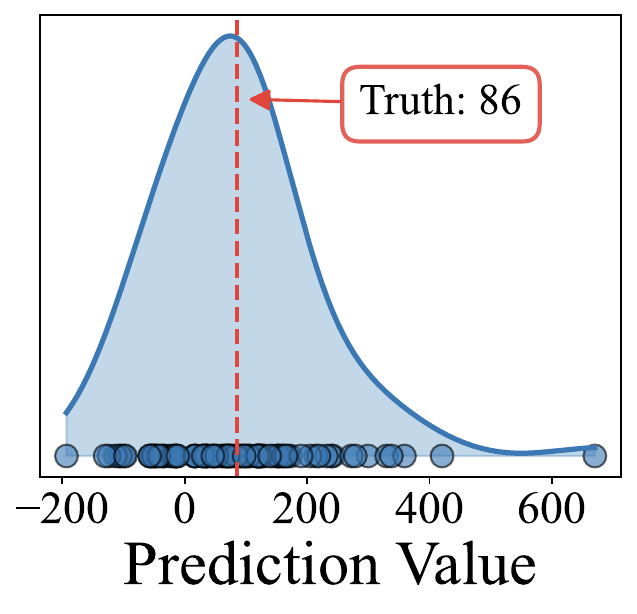}
        \vspace{-2mm}
        \caption{Check-in (NYC)}
    \end{subfigure}
    \begin{subfigure}[b]{0.32\columnwidth}
        \centering
        \includegraphics[width=0.96\columnwidth]{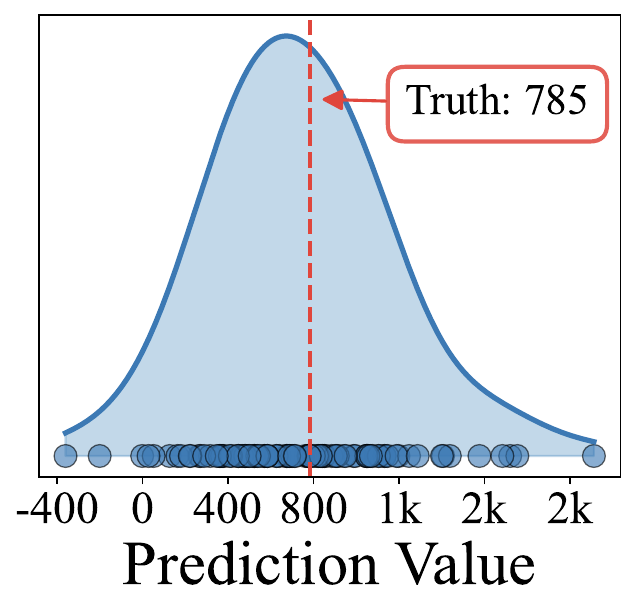}
        \vspace{-2mm}
        \caption{Service Call (NYC)}
    \end{subfigure}

    \begin{subfigure}[b]{0.32\columnwidth}
        \centering
        \includegraphics[width=0.97\columnwidth]{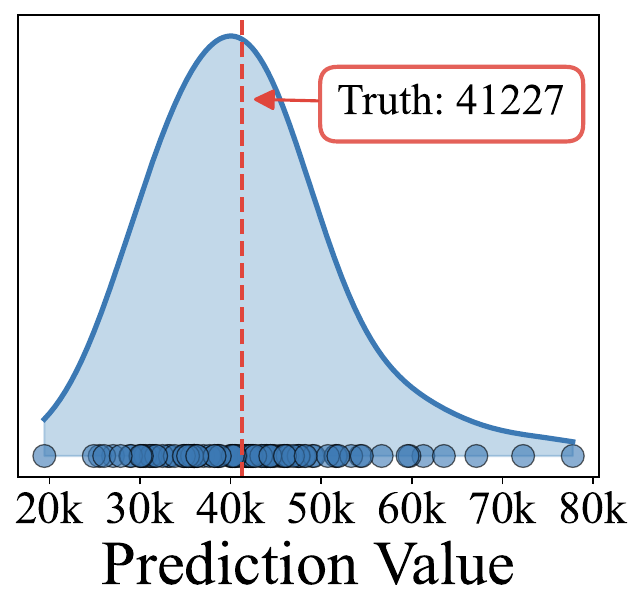}
        \vspace{-2mm}
        \caption{Population (CHI)}
    \end{subfigure} 
    \begin{subfigure}[b]{0.32\columnwidth}
        \centering
        \includegraphics[width=0.93\columnwidth]{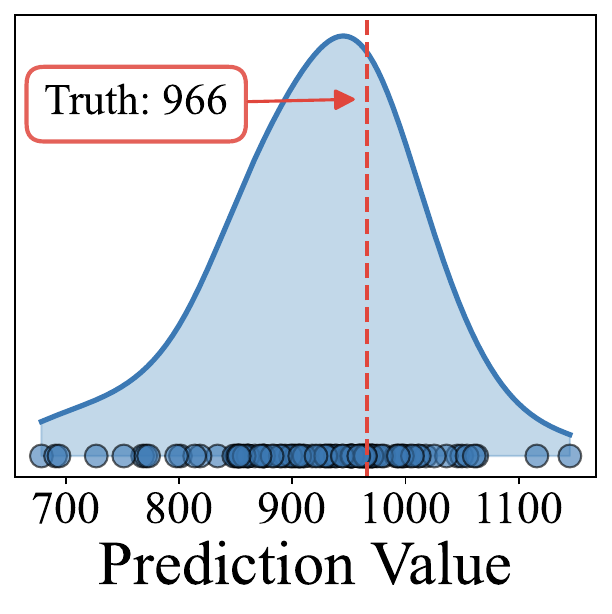}
        \vspace{-2mm}
        \caption{Carbon (CHI)}
    \end{subfigure}
    \begin{subfigure}[b]{0.32\columnwidth}
        \centering
        \includegraphics[width=0.94\columnwidth]{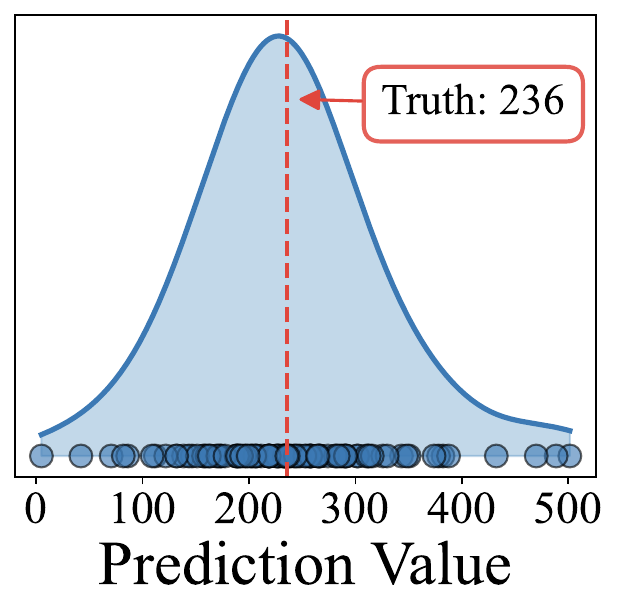}
        \vspace{-2mm}
        \caption{Nightlight (SF)}
    \end{subfigure}
    
    \caption{Predictive distribution of sample regions.}
    \label{fig:pred_distribution}
\end{figure}

We further examine the quality of the predictions generated by \diffM\ and the module's ability to capture uncertainty information. A key advantage of the diffusion process is that it generates a probability distribution over possible outcomes rather than producing a single deterministic prediction. 
To this end, we randomly select six regions from three datasets and let \diffM\ generate 100 stochastic predictions for each region across six different tasks. We then apply the Epanechnikov kernel density estimation method~\cite{davis2011remarks} to visualize the resulting predictive distributions. 

In Fig.~\ref{fig:pred_distribution}, the blue curve represents the distribution of the predictions generated by \diffM; the red dashed line indicates the corresponding ground truth values; and the blue circles denote individual predictions. Based on the results, the predictive distributions are generally unimodal and well-concentrated, indicating that the model produces stable and coherent predictions for individual regions. Importantly, the ground-truth values consistently fall within high-density areas of the predicted distributions, indicating the high quality of the generated distributions by \diffM. Overall, these visualizations highlight the benefit of stochastic prediction in providing not only accurate point estimates but also meaningful uncertainty information, which is particularly important for downstream spatial analysis and decision-making tasks.


\subsection{Cross-country Applicability}
\label{subsec:appendix_cross_country}

We further evaluate the applicability of \model\ over cities outside the USA, including Singapore (Asia) and Lisbon (Europe). 
Details of these two datasets are summrized in Table~\ref{tab:datasets_2}.
As some baseline models cannot be adapted to these cities due to limited region feature data (mobility data) availability, we compare \model\ with the SOTA model FlexiReg, which achieved the best accuracy among baselines in our previous experiments, on check-in and population count prediction tasks. 
We evaluate both models under the cross-city setting, where all models are trained on the NYC, CHI, and SF datasets and then evaluated on a new target city.

\begin{table}[ht]
\centering
\caption{Dataset Statistics: Singapore and Lisbon.}
\label{tab:datasets_2}
\setlength{\tabcolsep}{6pt}
\resizebox{0.9\columnwidth}{!}{
\begin{tabular}{lrr}
\toprule
&\textbf{Singapore~\cite{sgOpendata}} & \textbf{Lisbon~\cite{lxRegion}}
\\ \midrule
\#regions & 324 & 53  \\
\midrule
\#spatial partition units & 748 & 690  \\
\midrule
\#POIs & 65,082 & 43,961  \\
\midrule
\#POI categories & 15 & 15  \\
\midrule

{\#check-ins} & 355,463 & 24,327  \\
(data collection time) & 04/2012 - 09/2013 & 04/2012 - 09/2013 \\
\midrule

Population counts & 4,296,918 & 507,846 \\
(data collection time) & 2020 & 2020 \\

\bottomrule
\end{tabular}
}
\vspace{-1mm}
\end{table}

\begin{table}[htbp] 
\captionsetup{justification=centering}
\caption{\small Prediction Accuracy over Cities in Different Countries}
\label{tab:cross_city}
\setlength{\tabcolsep}{3pt} 
\renewcommand{\arraystretch}{0.9} 
\resizebox{\columnwidth }{!}{
\begin{tabular}{l |c c|c c}
    \toprule [0.4ex] \\[-3ex]

    & \multicolumn{2}{c|}{\textbf{\makecell[c]{NYC \& CHI \& SF (X) \\ $\rightarrow$ Singapore (Y)}}} & \multicolumn{2}{c}{\textbf{\makecell[c]{NYC \& CHI \& SF (X) \\ $\rightarrow$ Lisbon (Y)}}} \\
    \cmidrule(lr){2-3} \cmidrule(lr){4-5}

    & Check-in & Population & Check-in & Population \\
    \midrule
    
    &$R^{2} \uparrow$ & $R^{2} \uparrow$ & $R^{2} \uparrow$ & $R^{2} \uparrow$ \\
    \midrule
    FlexiReg & 0.223 $\pm$ 0.015   & 0.518 $\pm$  0.018
    & 0.716 $\pm$ 0.014 &  0.881 $\pm$ 0.005 
    \\
    \midrule 
    \textbf{\model} & \textbf{ 0.368 $\pm$ 0.016} & \textbf{ 0.625 $\pm$ 0.006} 
    & \textbf{ 0.781 $\pm$ 0.011} & \textbf{0.909 $\pm$ 0.010}
    \\
    \midrule
    \textbf{Improvement}
    & \red{\textbf{+65.0\%}}
    & \red{\textbf{+20.7\%}}
    & \red{\textbf{+9.1\%}}
    & \red{\textbf{+3.2\%}} \\
    \bottomrule
\end{tabular}
}
\end{table}

As shown in Table~\ref{tab:cross_city}, \model\ consistently outperforms FlexiReg across different tasks and cities, achieving improvements of up to 65.0\% in terms of $R^2$. These results demonstrate that \model\ generalizes effectively across diverse urban environments and remains robust when applied to cities from different countries, confirming its strong cross-city (country) applicability.

\subsection{Impact of Model Parameter Values}
\label{subsec:appendix_of_impact_of_parameter_learning}

We study model sensitivity to two key hyper-parameters: the number of random walks ($k$) and the number of sampling rounds ($\#SR$). By default, all experiments in this subsection are evaluated on the NYC dataset, and results are reported using $R^2$ for conciseness.

\subsubsection{Impact of the Number of Random Walks ($k$).} 
\label{subsubsec:para_num_RW}

The number of random walks plays a critical role in balancing structural coverage and representational discriminability of regions. We vary $k$ from 2 to 16 to study its impact. As shown in Table~\ref{tab:para_num_RW}, \model\ achieves comparable accuracy when using 2, 4, or 8 random walks, and the accuracy degrades when $k$ increases to 16. 
When $k$ is too small, each region is characterized by a limited number of sampled paths, making the resulting region embeddings more sensitive to sampling noise and thus exhibiting higher variance. When $k$ is too large, the additional random walks may just overlap with existing ones, creating redundant walks. Feeding these redundant walks into the transformer model causes attention dilution, where the attention weights of informative tokens are dispersed across many repetitive tokens. As a result, region embeddings become overly similar, which ultimately degrades model performance.
Based on these observations, we set $k$ to 8 by default.

\begin{table}[htbp] 
\caption{Impact of $k$ on NYC.}
\label{tab:para_num_RW}
\begin{center}
\setlength{\tabcolsep}{4pt}
\renewcommand{\arraystretch}{1.05}
\resizebox{\columnwidth}{!}{
\begin{tabular}{l*{4}{c}}
    \toprule
    $\#RW$ & 2 & 4 & 8 & 16 \\
    \midrule
    
    Crime
    & 0.721 \pmstd 0.021
    & 0.724 \pmstd 0.008
    & \textbf{0.726 \pmstd 0.007}
    & 0.711 \pmstd 0.011 \\
    
    Check-in
    & 0.776 \pmstd 0.029
    & \textbf{0.781 \pmstd 0.011}
    & 0.779 \pmstd 0.012
    & 0.772 \pmstd 0.009 \\
    
    Service Call
    & 0.585 \pmstd 0.023
    & \textbf{0.589 \pmstd 0.009}
    & 0.584 \pmstd 0.011
    & 0.569 \pmstd 0.012 \\
    
    Population
    & 0.624 \pmstd 0.026
    & \textbf{0.626 \pmstd 0.006}
    & 0.621 \pmstd 0.013
    & 0.613 \pmstd 0.005 \\
    
    Carbon
    & 0.383 \pmstd 0.029
    & 0.389 \pmstd 0.006
    & \textbf{0.390 \pmstd 0.011}
    & 0.363 \pmstd 0.019 \\
    
    Nightlight
    & \textbf{0.498 \pmstd 0.031}
    & 0.492 \pmstd 0.013
    & 0.488 \pmstd 0.011
    & 0.475 \pmstd 0.009 \\
    
    \bottomrule
\end{tabular}
}
\end{center}
\end{table}

\subsubsection{Impact of the Number of Sampling Rounds ($\#SR$).} 
\label{subsubsec:para_num_SR}

The number of sampling rounds impacts both the variance of predictions and the computational costs. We vary $\#SR$ from 1 to 100 to evaluate its impact. As shown in Table~\ref{tab:para_num_SR}, \model~achieves
the best accuracy with different values of $\#SR$ on different downstream tasks. For most tasks, accuracy improves as 
$\#SR$ increases from 1 to 10 and then remains relatively stable when $\#SR$ further increases from 10 to 100.
\begin{table}[htbp] 
\caption{Impact of $\#SR$ ($R^2 \uparrow$ on NYC).}
\label{tab:para_num_SR}
\begin{center}
\setlength{\tabcolsep}{4pt}
\renewcommand{\arraystretch}{1.05}
\resizebox{\columnwidth}{!}{
\begin{tabular}{l*{4}{c}}
    \toprule
    $\#SR$ & 1 & 10 & 50 & 100 \\
    \midrule
    
    Crime
    & 0.645 \pmstd 0.028
    & \textbf{0.724 \pmstd 0.008}
    & 0.715 \pmstd 0.008
    & 0.710 \pmstd 0.006 \\
    
    Check-in
    & 0.754 \pmstd 0.022
    & \textbf{0.781 \pmstd 0.013}
    & 0.776 \pmstd 0.013
    & 0.779 \pmstd 0.009 \\
    
    Service Call
    & 0.538 \pmstd 0.039
    & 0.589 \pmstd 0.009
    & \textbf{0.590 \pmstd 0.011}
    & 0.583 \pmstd 0.009 \\
    
    Population
    & 0.537 \pmstd 0.034
    & \textbf{0.626 \pmstd 0.006}
    & 0.621 \pmstd 0.009
    & \textbf{0.626 \pmstd 0.005} \\
    
    Carbon
    & \textbf{0.401 \pmstd 0.051}
    & 0.389 \pmstd 0.006
    & 0.381 \pmstd 0.005
    & 0.385 \pmstd 0.005 \\
    
    Nightlight
    & 0.430 \pmstd 0.041
    & 0.492 \pmstd 0.013
    & \textbf{0.502 \pmstd 0.010}
    & 0.485 \pmstd 0.008 \\
    
    \bottomrule
\end{tabular}
}
\end{center}
\end{table}

When $\#SR$ is 1, the predictions suffer from high stochastic variance, leading to unstable outputs. Increasing $\#SR$  reduces sampling noise through averaging multiple predictions, leading to more reliable results. However, larger $\#SR$ values also incur higher computational costs without providing consistent performance gains. Therefore, we have set $\#SR$ as 10 by default to balance effectiveness with efficiency.

\begin{table*}[htbp]
\centering
\caption{Overall Prediction Accuracy Results When Powering Existing Models with Our \diffM\ Module.}
\label{tab:app_diffusionM_full}

\setlength{\tabcolsep}{3pt}
\renewcommand{\arraystretch}{1.0}
\resizebox{\textwidth }{!}{
\begin{tabular}{l|ccc|ccc|ccc}

\toprule [0.4ex] \\[-3.5ex]

 &
\multicolumn{3}{c|}{Crime Prediction} & \multicolumn{3}{c|}{Check-in Prediction} & \multicolumn{3}{c}{Service Call Prediction}  \\

\cmidrule(lr){2-4} \cmidrule(lr){5-7} \cmidrule(lr){8-10}
    
&MAE $\downarrow$ & RMSE $\downarrow$ & $R^{2} \uparrow$ 
& MAE $\downarrow$ & RMSE $\downarrow$ & $R^{2} \uparrow$ 
 & MAE $\downarrow$ & RMSE $\downarrow$ & $R^{2} \uparrow$ \\
\midrule

HREP
& 62.8 $\pm$ 2.1 & 83.1 $\pm$ 2.3 & 0.681 $\pm$ 0.014
& 276.3 $\pm$ 11.7 & 448.2 $\pm$ 16.9 & 0.700 $\pm$ 0.022
& 1430 $\pm$ 29 & 2128 $\pm$ 34 & 0.398 $\pm$ 0.019 \\

\textbf{HREP-DiffCT}
& \textbf{50.8 $\pm$ 1.6} & \textbf{68.8 $\pm$ 2.1} & \textbf{0.784 $\pm$ 0.013}
& \textbf{205.9 $\pm$ 7.8} & \textbf{316.5 $\pm$ 17.1} & \textbf{0.850 $\pm$ 0.016}
& \textbf{1188 $\pm$ 31} & \textbf{1756 $\pm$ 33} & \textbf{0.589 $\pm$ 0.014} \\

\midrule
\textbf{Improvement}
& \red{\textbf{+19.1\%}} & \red{\textbf{+17.2\%}} & \red{\textbf{+15.1\%}}
& \red{\textbf{+25.5\%}} & \red{\textbf{+29.4\%}} & \red{\textbf{+21.4\%}}
& \red{\textbf{+16.9\%}} & \red{\textbf{+17.5\%}} & \red{\textbf{+48.0\%}} \\

\bottomrule \toprule

UrbanCLIP
& 97.4 $\pm$ 2.6 & 126.1 $\pm$ 1.9 & 0.267 $\pm$ 0.012
& 393.6 $\pm$ 5.9 & 602.4 $\pm$ 3.1 & 0.458 $\pm$ 0.005
& 1409 $\pm$ 7 & 2401 $\pm$ 16 & 0.232 $\pm$ 0.005 \\

\textbf{UrbanCLIP-DiffCT}
& \textbf{77.6 $\pm$ 3.2} & \textbf{107.4 $\pm$ 4.2} & \textbf{0.471 $\pm$ 0.039}
& \textbf{348.1 $\pm$ 13.5} & \textbf{489.2 $\pm$ 17.7} & \textbf{0.642 $\pm$ 0.025}
& \textbf{1338 $\pm$ 52} & \textbf{2179 $\pm$ 64} & \textbf{0.369 $\pm$ 0.033} \\

\midrule
\textbf{Improvement}
& \red{\textbf{+20.3\%}} & \red{\textbf{+14.8\%}} & \red{\textbf{+76.4\%}}
& \red{\textbf{+11.6\%}} & \red{\textbf{+18.8\%}} & \red{\textbf{+40.2\%}}
& \red{\textbf{+5.0\%}} & \red{\textbf{+9.3\%}} & \red{\textbf{+9.1\%}} \\

\bottomrule \toprule 

HAFusion
& 56.0 $\pm$ 1.3 & 76.1 $\pm$ 2.2 & 0.734 $\pm$ 0.015
& 202.8 $\pm$ 7.2 & 322.8 $\pm$ 12.6 & 0.844 $\pm$ 0.012
& 1274 $\pm$ 20 & 1952 $\pm$ 27 & 0.493 $\pm$ 0.014 \\

\textbf{HAFusion-DiffCT}
& \textbf{46.3 $\pm$ 1.3} & \textbf{62.2 $\pm$ 2.1} & \textbf{0.823 $\pm$ 0.011}
& \textbf{185.6 $\pm$ 7.5} & \textbf{258.2 $\pm$ 11.4} & \textbf{0.900 $\pm$ 0.009}
& \textbf{1167 $\pm$ 23} & \textbf{1582 $\pm$ 14} & \textbf{0.667 $\pm$ 0.006} \\

\midrule
\textbf{Improvement}
& \red{\textbf{+17.3\%}} & \red{\textbf{+18.3\%}} & \red{\textbf{+12.1\%}}
& \red{\textbf{+8.5\%}} & \red{\textbf{+20.0\%}} & \red{\textbf{+6.6\%}}
& \red{\textbf{+8.4\%}} & \red{\textbf{+18.9\%}} & \red{\textbf{+35.3\%}} \\

\bottomrule \toprule 

GURPP
& 72.9 $\pm$ 2.3 & 94.7 $\pm$ 4.6 & 0.589 $\pm$ 0.039
& 267.0 $\pm$ 15.2 & 403.3 $\pm$ 16.5 & 0.757 $\pm$ 0.019
& 1480 $\pm$ 33 & 2115 $\pm$ 84 & 0.405 $\pm$ 0.048 \\

\textbf{GURPP-DiffCT}
& \textbf{60.5 $\pm$ 3.6} & \textbf{77.9 $\pm$ 4.1} & \textbf{0.722 $\pm$ 0.029}
& \textbf{222.8 $\pm$ 12.1} & \textbf{329.8 $\pm$ 17.5} & \textbf{0.837 $\pm$ 0.017}
& \textbf{1317 $\pm$ 23} & \textbf{1882 $\pm$ 29} & \textbf{0.528 $\pm$ 0.009} \\

\midrule
\textbf{Improvement}
& \red{\textbf{+17.0\%}} & \red{\textbf{+17.7\%}} & \red{\textbf{+22.6\%}}
& \red{\textbf{+16.6\%}} & \red{\textbf{+18.2\%}} & \red{\textbf{+10.6\%}}
& \red{\textbf{+11.0\%}} & \red{\textbf{+11.0\%}} & \red{\textbf{+30.4\%}} \\

\bottomrule \toprule [0.4ex] \\[-3.5ex] 

 &
\multicolumn{3}{c|}{Population Prediction} & \multicolumn{3}{c|}{Carbon Prediction} & \multicolumn{3}{c}{Nightlight Prediction}  \\

\cmidrule(lr){2-4} \cmidrule(lr){5-7} \cmidrule(lr){8-10}
    
&MAE $\downarrow$ & RMSE $\downarrow$ & $R^{2} \uparrow$ 
& MAE $\downarrow$ & RMSE $\downarrow$ & $R^{2} \uparrow$ 
 & MAE $\downarrow$ & RMSE $\downarrow$ & $R^{2} \uparrow$ \\
\midrule

HREP
& 2656$\,\pm\,$59 & 3461$\,\pm\,$83 & 0.571$\,\pm\,$0.021
& 718$\,\pm\,$35 & 1187$\,\pm\,$37 & 0.184$\,\pm\,$0.050
& 58.7$\,\pm\,$1.1 & 78.7$\,\pm\,$1.3 & -0.026$\,\pm\,$0.024 \\

\textbf{HREP-DiffCT}
& \textbf{2389$\,\pm\,$51} & \textbf{3051$\,\pm\,$88} & \textbf{0.667$\,\pm\,$0.019}
& \textbf{595$\,\pm\,$22} & \textbf{1041$\,\pm\,$30} & \textbf{0.373$\,\pm\,$0.036}
& \textbf{50.2$\,\pm\,$0.6} & \textbf{70.9$\,\pm\,$0.8} & \textbf{0.167$\,\pm\,$0.011} \\

\midrule
\textbf{Improvement}
& \red{\textbf{+10.05\%}} & \red{\textbf{+11.85\%}} & \red{\textbf{+16.81\%}}
& \red{\textbf{+17.15\%}} & \red{\textbf{+12.31\%}} & \red{\textbf{+102.72\%}}
& \red{\textbf{+14.48\%}} & \red{\textbf{+9.91\%}} & \red{\textbf{+742.31\%}} \\

\bottomrule \toprule

UrbanCLIP
& 3338$\,\pm\,$11 & 4499$\,\pm\,$16 & 0.276$\,\pm\,$0.002
& 729$\,\pm\,$12 & 1238$\,\pm\,$19 & 0.113$\,\pm\,$0.013
& 48.9$\,\pm\,$0.3 & 63.2$\,\pm\,$1.0 & 0.337$\,\pm\,$0.019 \\

\textbf{UrbanCLIP-DiffCT}
& \textbf{3286$\,\pm\,$52} & \textbf{4153$\,\pm\,$87} & \textbf{0.392$\,\pm\,$0.024}
& \textbf{616$\,\pm\,$11} & \textbf{1148$\,\pm\,$24} & \textbf{0.238$\,\pm\,$0.033}
& \textbf{46.9$\,\pm\,$0.9} & \textbf{58.4$\,\pm\,$1.4} & \textbf{0.434$\,\pm\,$0.026} \\

\midrule
\textbf{Improvement}
& \red{\textbf{+1.56\%}} & \red{\textbf{+7.69\%}} & \red{\textbf{+42.03\%}}
& \red{\textbf{+15.57\%}} & \red{\textbf{+7.27\%}} & \red{\textbf{+110.62\%}}
& \red{\textbf{+4.09\%}} & \red{\textbf{+7.59\%}} & \red{\textbf{+28.78\%}} \\

\bottomrule \toprule 

HAFusion
& 2497$\,\pm\,$50 & 3277$\,\pm\,$82 & 0.616$\,\pm\,$0.019
& 714$\,\pm\,$10 & 1184$\,\pm\,$13 & 0.189$\,\pm\,$0.019
& 55.0$\,\pm\,$0.9 & 76.3$\,\pm\,$1.5 & 0.035$\,\pm\,$0.013 \\

\textbf{HAFusion-DiffCT}
& \textbf{2297$\,\pm\,$92} & \textbf{2931$\,\pm\,$90} & \textbf{0.692$\,\pm\,$0.019}
& \textbf{557$\,\pm\,$22} & \textbf{1053$\,\pm\,$13} & \textbf{0.358$\,\pm\,$0.017}
& \textbf{50.1$\,\pm\,$1.1} & \textbf{70.7$\,\pm\,$2.1} & \textbf{0.171$\,\pm\,$0.026} \\

\midrule
\textbf{Improvement}
& \red{\textbf{+8.01\%}} & \red{\textbf{+10.56\%}} & \red{\textbf{+12.34\%}}
& \red{\textbf{+21.93\%}} & \red{\textbf{+11.05\%}} & \red{\textbf{+89.42\%}}
& \red{\textbf{+8.91\%}} & \red{\textbf{+7.34\%}} & \red{\textbf{+388.57\%}} \\

\bottomrule \toprule 

GURPP
& 3316$\,\pm\,$75 & 4159$\,\pm\,$87 & 0.381$\,\pm\,$0.026
& 782$\,\pm\,$1 & 1289$\,\pm\,$3 & 0.021$\,\pm\,$0.001
& 59.1$\,\pm\,$0.2 & 78.1$\,\pm\,$0.3 & -0.012$\,\pm\,$0.005 \\

\textbf{GURPP-DiffCT}
& \textbf{3210$\,\pm\,$59} & \textbf{4027$\,\pm\,$74} & \textbf{0.420$\,\pm\,$0.021}
& \textbf{678$\,\pm\,$11} & \textbf{1174$\,\pm\,$14} & \textbf{0.204$\,\pm\,$0.019}
& \textbf{51.0$\,\pm\,$1.1} & \textbf{71.1$\,\pm\,$1.4} & \textbf{0.160$\,\pm\,$0.009} \\

\midrule
\textbf{Improvement}
& \red{\textbf{+3.20\%}} & \red{\textbf{+3.17\%}} & \red{\textbf{+10.24\%}}
& \red{\textbf{+13.26\%}} & \red{\textbf{+8.94\%}} & \red{\textbf{+871.43\%}}
& \red{\textbf{+13.71\%}} & \red{\textbf{+8.96\%}} & \red{\textbf{+1433.33\%}} \\

\bottomrule
\end{tabular}
}
\end{table*}

\end{document}